\documentclass{article}

\PassOptionsToPackage{numbers, compress}{natbib}

\usepackage[final,main]{neurips_2026}

\usepackage[utf8]{inputenc}
\usepackage[T1]{fontenc}
\usepackage{hyperref}
\usepackage{url}
\hypersetup{hypertexnames=false}
\usepackage{booktabs}
\usepackage{longtable}
\usepackage{amsfonts}
\usepackage{nicefrac}
\usepackage{microtype}
\usepackage{amsmath}
\usepackage{algorithm}
\usepackage{algpseudocode}
\usepackage{tcolorbox}
\tcbuselibrary{breakable}
\usepackage{tikz}
\usepackage{placeins}
\usepackage[table]{xcolor}
\usepackage{adjustbox}
\usepackage{subcaption}

\usetikzlibrary{arrows.meta,positioning,shapes.geometric}

\title{From paper to benchmark:  agentic, framework-based reproduction of under-specified methods in machine health intelligence}

\author{
Raffael Theiler\\
\textit{EPFL}\\
\And
Ludovico Comito\\
\textit{University of Roma ``La Sapienza''}\\
\And
David Leko\\
\textit{EPFL}\\
\And
Leandro Von Krannichfeldt\\
\textit{EPFL}\\
\And
Lev Telyatnikov\\
\textit{EPFL}\\
\And
Olga Fink\\
\textit{EPFL}\\
}

\begin{document}

\maketitle

\begin{abstract}

Industrial Prognostics and Health Management (PHM) provides a representative case study for a broader challenge in applied machine learning: translating published papers into executable, benchmark-ready implementations.
Reproducing under-specified methods in PHM is particularly difficult due to restricted access to industrial datasets, incomplete reporting of preprocessing and evaluation protocols, and implicit design choices (e.g., windowing, target construction, data splits) that critically affect performance. Existing paper-to-code systems generate implementations for individual papers, but these artifacts are often not directly comparable due to inconsistencies in assumptions and evaluation settings.
We introduce \emph{agentic, framework-based PHM paper reproduction}, where an agent translates a paper into a shared PHM benchmark framework via a \emph{slot-binding interface}. This interface maps equations and protocol descriptions into structured components (task definitions, dataset adapters, windowing, targets, models, and evaluators), while explicitly recording unresolved assumptions. The resulting implementations are validated against standardized task contracts and evaluation hooks, enabling consistent and comparable benchmarking.
We evaluate this approach on 16 PHM papers, comparing framework-enhanced, skill-based and prompt-based agentic reproduction against a recent framework-free paper-reproduction agent. We assess reproduction success, model-based code evaluation, framework binding of paper assumptions, and cross-paper benchmark comparability under standardized protocols.
Our results show that coupling agentic generation with a shared framework transforms paper reproduction from isolated code synthesis into executable, assumption-aware, and systematically comparable benchmark implementations.

\end{abstract}

\section{Introduction}
\label{sec:intro}

In applied machine learning, translating a published paper into a benchmark-ready implementation is often a fundamentally under-specified reproducibility problem rather than a straightforward coding task. We study Prognostics and Health Management (PHM) as a representative case of this challenge. PHM methods monitor industrial assets, diagnose faults, and forecast degradation,  increasingly supporting maintenance decisions in safety-critical settings where failures carry substantial operational and economic cost. However, as in many  applied ML  domains, executable artifacts are rarely released alongside publications: a large-scale audit of PHM literature found that fewer than 1\% of papers released both code and data \citep{vonhahn2022phmreproducibility}. The result is that reproducing prior work becomes labor-intensive and unreliable, cross-paper benchmarking becomes difficult, and methodological progress slows because new approaches cannot be validated against verified baselines. This gap is partly structural because PHM research is often shaped by industrial constraints, proprietary infrastructure, and site-specific data agreements that limit full reproducibility. Consequently, PHM papers typically specify methods at the level of equations, architectures, and benchmark tables while omitting execution details. Even when benchmark standardization efforts improve parts of the evaluation pipeline \citep{li2025phmvibench}, reproducibility still depends on undocumented protocol decisions such as preprocessing scope, temporal windowing, normalization, split policy, target construction, and asset-specific conventions, all of which can materially affect reported performance.  
Implementing a PHM paper therefore  becomes a reconstruction problem: recovering the implicit experimental protocol required to instantiate the reported method under executable and comparable conditions.
Recent advances partially address this challenge. Paper-to-code systems such as Paper2Code \citep{seo2025paper2code}, together with broader work on structured paper-to-code generation \citep{trofimova2024coderefine,zhao2025autoreproduce,li2025deepcode}, demonstrate that LLM-based agents can generate substantial amounts of repository-level research code directly from scientific manuscripts. In parallel, PaperBench \citep{starace2025paperbench} shows that faithful reproduction requires more than plausible code generation by separating implementation, execution, and result matching into distinct evaluation axes. However, existing approaches still treat each paper as an isolated software project. The resulting reproductions differ substantially in project structure, configuration logic, training loops, and evaluation code, making reproduced methods difficult to compare under shared conditions. Related reproducibility benchmarks further show that execution reliability and result matching degrade sharply once dependency management, environment setup, and experimental orchestration shift to the agent \citep{siegel2024corebench,kim2025autoexperiment}. Recent work on repository-level code generation with dependency recovery \citep{leanh2026treat} improves structural consistency, but still does not address the configuration-driven protocol semantics that govern most modern ML evaluation pipelines.

We address this gap by treating PHM paper implementation as a structured integration problem within a shared framework rather than as unrestricted repository generation. Instead of producing standalone projects with custom pipelines and evaluation code, the agent maps a paper onto standardized framework components with shared data interfaces, preprocessing rules, and evaluation protocols. For example, a paper on remaining useful life prediction for turbofan engines using a sliding-window model is translated into concrete framework components specifying the task type, dataset adapter, windowing procedure, target-construction policy, model wrapper, and evaluation metrics. Any missing or ambiguous details, such as window length or target clipping, are explicitly recorded as assumptions rather than hidden in implementation code. The framework then executes all reproduced methods under consistent protocol rules and evaluation conditions. By grounding implementation inside shared infrastructure, our workflow turns paper reproduction from unconstrained code synthesis into an auditable and benchmarkable integration task. This provides reproducibility, comparability, and explicit assumption tracking that isolated repository generation cannot guarantee, while agentic automation supplies the scalability needed to reconstruct large portions of the literature.

More broadly, we treat PHM as a representative case study for a central question in scientific machine learning: can framework grounding transform paper reproduction from unconstrained code generation into a verifiable and benchmarkable scientific integration task? Our central hypothesis is that  coupling paper understanding to a shared execution framework makes protocol assumptions, inferred decisions, and implementation failures explicit, auditable, and systematically measurable. Rather than evaluating whether an agent can merely generate plausible repository code, we study whether framework grounding increases the rate at which under-specified paper implementations become executable and benchmarkable. In this context, we evaluate whether an agent can reconstruct compatible, executable, and comparable implementations under shared evaluation semantics. To this end, we introduce a framework-grounded agentic PHM reproduction workflow in which an agent reads a PHM paper, generates an implementation, and returns either a benchmark-ready artifact or an auditable failure report. We proceed as follows:
a) We formulate \emph{framework-grounded agentic PHM reproduction} as a benchmarkable paper-to-implementation task in a scientific ML, focusing on a domain with severe reproducibility constraints and limited code availability.
b) We propose a framework-coupled agent workflow that maps paper descriptions into a shared PHM framework with explicit protocol verification, assumption tracking, and executable evaluation hooks.
c) We introduce a multi-axis evaluation protocol for framework-grounded paper reproduction that measures framework integration accuracy, slot-level binding diagnostics, execution success, implementation fidelity, benchmarkability, code quality, and implementation complexity.
We empirically analyze how framework grounding changes the reproducibility properties of paper-to-code generation compared to unconstrained repository synthesis, including the roles of explicit framework documentation, assumption tracking, and structured scaffolding in enabling executable recovery from under-specified scientific papers.

\begin{figure}[H]
  \centering
  \includegraphics[width=\linewidth]{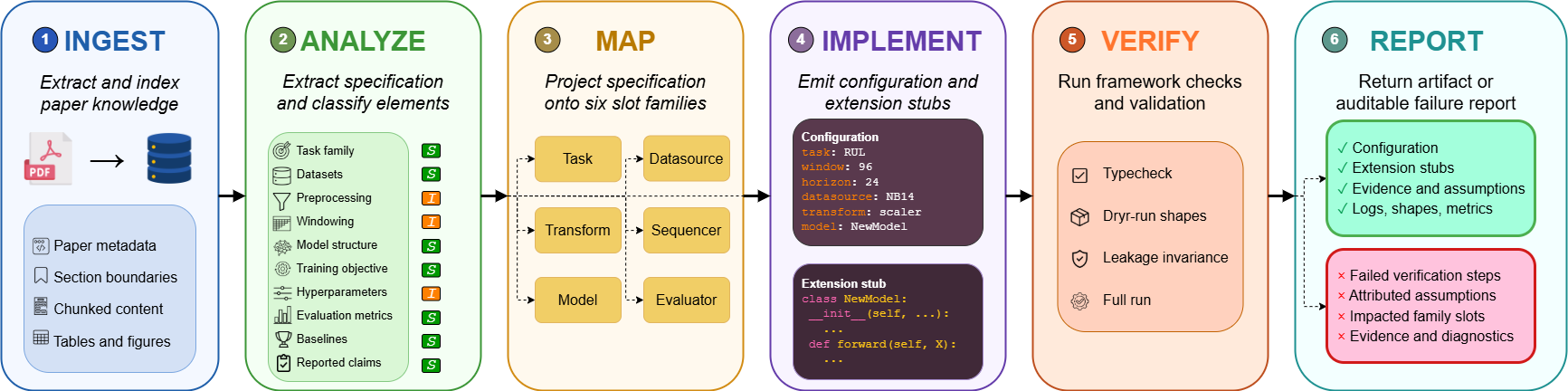}

  \caption{\textbf{Visual abstract.} The workflow converts an under-specified PHM paper into a benchmark-ready artifact in a shared framework. The agent ingests the paper, analyzes specified and inferred method elements, maps them to framework slots, implements a resolved configuration or extension stubs, verifies contracts and empirical behavior, and returns an artifact or auditable failure report. $\mathcal{S}$ marks information specified by the paper; $\mathcal{I}$ marks inferred decisions recorded as assumptions.}
  \label{fig:visual-abstract}
\end{figure}

\section{Related work}
\label{sec:related}

\paragraph{Paper-to-code systems.}
A growing body of work studies whether LLM agents can translate scientific papers into executable code. Early systems such as CodeRefine \citep{trofimova2024coderefine} rely on structured extraction and retrieval-augmented refinement, while newer approaches target repository-level generation through staged planning, retrieval, debugging, and multimodal reasoning \citep{seo2025paper2code,zhao2025autoreproduce,lin2025autop2c}. These systems generate substantial research code, but reproduced methods are typically emitted as isolated repositories with their own environments and evaluation logic, preventing direct comparison under shared task contracts. Our work instead studies paper implementation within a shared domain-specific framework, enabling reproduced methods to be benchmarked under common conditions by construction.
Recent work also explores broader automation of the scientific workflow, including ML engineering \citep{jiang_aide_2025}, hypothesis generation \citep{gottweis_towards_2025}, and end-to-end research automation \citep{schmidgall_agent_2025,lu_ai_2024}. However, these approaches remain sensitive to evaluation design and cascading errors \citep{qiu2026prbench}, motivating modular, verified workflows.

\paragraph{Evaluating faithful reproduction.}
Several benchmarks study what faithful paper implementation requires. PaperBench \citep{starace2025paperbench} separates code synthesis, execution, and result matching, showing that frontier agents remain weaker at execution and verification than code generation. CORE-Bench \citep{siegel2024corebench} similarly shows reproducibility often fails even when code and data are available, once environment setup and execution are required. AutoExperiment \citep{kim2025autoexperiment} further shows performance degrades sharply as implementation responsibility shifts from existing repositories to the agent itself. Across these works, the dominant bottleneck is not syntax but \emph{underspecification}: important implementation details are often omitted from papers \citep{xiang2025scireplicate,hua2025researchcodebench,tian2024scicode}. Our work adds another axis, \emph{framework integration}, since methods in a shared PHM framework can fail before execution through incorrect task mappings, data adapters, or protocol bindings.

\paragraph{Verification-first agent design.}
Verification-centered agent work similarly emphasizes explicit checks. TS-Agent \citep{liu2025tsagent} maintains evidence logs, delegates statistical operations to tools, applies self-refinement, and verifies outputs before acceptance. Related work adds iterative tool use and intermediate-state tracking \citep{ye_ts-reasoner_2025, zhou_time_2026, wu_timeart_2026}. Argos \citep{gu2025argos} generates deterministic anomaly rules for auditable deployment, while AutoIAD \citep{ji2025autoiad} uses manager-driven orchestration with domain-specialized agents. These systems improve forecasting and anomaly-detection workflows; we apply the same principles to reconstructing executable pipelines from scientific papers within a shared framework.

\paragraph{PHM reproducibility and domain infrastructure.}
In PHM, reproducibility is challenging because undocumented choices can materially affect results. In bearing-fault diagnosis, changing the overlap between extracted signal windows can substantially affect reported classification accuracy, while in RUL estimation, target-label formulations can change MAE by an order of magnitude. Existing PHM frameworks, including GSAP \citep{teubert_generic_2020}, ProgPy \citep{teubert2023progpy}, and PHM-Vibench \citep{li2025phmvibench}, show the value of shared infrastructure, but do not address populating them from under-specified papers. Our work combines verification-driven agent design with framework-grounded method integration. The PHM framework \citep{anonymous2026framework} formalizes PHM methods through explicit task contracts, protocol abstractions, and model configurations that agents map to directly, enabling executable and benchmarkable implementations under shared protocols.

\section{Method}
\label{sec:method}

\paragraph{\textcolor{black}{Overview and motivation.}}
\textcolor{black}{PHM papers often specify the high-level method but omit implementation decisions that affect evaluation, such as windowing, target construction, and data-split or leakage-control details. Our method turns each paper into a framework-integrated artifact that can be evaluated under shared benchmark conditions, so outcomes are comparable across papers. We do this with a staged workflow that binds paper concepts into typed slot families of a shared PHM framework, records every under-specified choice as an assumption record, and validates the resulting implementation through verification gates.}

\textcolor{black}{In our experiments, the shared framework is instantiated with a modular PHM benchmarking infrastructure currently under anonymous review \citep{anonymous2026framework} (see supplementary material). In this framework instance, benchmark conditions---windowing, partitioning, leakage control, and metric definitions---are enforced by construction, so different papers are evaluated under the same protocol. Our contribution is the agent-side machinery that translates a paper into a valid framework binding: a staged workflow (Section~\ref{subsec:workflow}) that emits a framework binding, assumption records, and a verification-and-evaluation report; a mapping procedure for resolving paper concepts into typed slot families (Section~\ref{subsec:mapping}); and an attribution mechanism that links verification outcomes back to implicated assumptions (Section~\ref{subsec:assumptions}).}

\subsection{Problem formulation}

\textcolor{black}{Formally, let $P$ denote a PHM paper and let $\mathcal{F}$ denote the shared PHM framework, with fixed task contracts, dataset adapter interfaces, transformation and windowing operators, and metric definitions. Across papers, these benchmark conditions are held fixed and only the paper-specific slot binding changes. We scope the present work to two task families: prognostics (remaining useful life and health-indicator estimation) and diagnostics (fault or mode classification).}

\textcolor{black}{From $P$, the agent extracts the method specification needed to instantiate the paper inside $\mathcal{F}$: a task definition $\mathcal{T}$ (target semantics), a framework model $\mathcal{M}$ (predictor class), and a model configuration $\kappa$ (loss, training details, and hyperparameters). The agent then translates $(\mathcal{T}, \mathcal{M}, \kappa)$ into a valid slot binding inside $\mathcal{F}$ and returns an artifact}
\[
  A \;=\; \bigl(c,\; \mathcal{C},\; \mathcal{A},\; r\bigr),
\]

\textcolor{black}{Here $c$ is the resolved configuration that binds the paper into $\mathcal{F}$'s slot families. $\mathcal{C}$ is any task-contract-preserving extension required when the paper calls for functionality that is not expressible with existing framework components (e.g., a new datasource adapter, transform operator, or model wrapper). $\mathcal{A}$ is the set of assumption records capturing every implementation decision that $P$ leaves under-specified (e.g., missing transform parameters), each linked to the evidence used (or an explicit absence-of-evidence note) and to the resulting configuration. Finally, $r$ is a verification-and-evaluation report: either a benchmark-ready result or an auditable failure.}

\subsection{Agent workflow}
\label{subsec:workflow}

\textcolor{black}{The agent follows a staged workflow that takes a paper $P$ as input and returns the artifact $A=(c,\mathcal{C},\mathcal{A},r)$ inside $\mathcal{F}$.} \textcolor{black}{This design is inspired by staged planning-and-coding agent pipelines \citep{seo2025paper2code}, but the target is a typed framework binding rather than a free-form repository.}

\textcolor{black}{The workflow comprises six stages. \textsc{Ingest} constructs an indexed representation of $P$ (sections, figures/tables, and chunk-level spans) to support retrieval and citation. \textsc{Analyze} extracts the method specification $(\mathcal{T},\mathcal{M},\kappa)$ and records each required but unstated implementation choice as an assumption record. \textsc{Map} resolves the extracted specification into a slot-binding plan over $\mathcal{F}$'s typed interfaces. \textsc{Implement} materializes an executable configuration $c$ and any contract-preserving extensions $\mathcal{C}$. \textsc{Verify} evaluates the resulting binding under the shared benchmark conditions of $\mathcal{F}$ and produces the report $r$. Finally, \textsc{Report} packages $(c,\mathcal{C},\mathcal{A},r)$ as either a benchmark-ready artifact or an auditable failure report.}

\subsection{Framework-coupled mapping}
\label{subsec:mapping}

\textcolor{black}{In \textsc{Map}, the agent turns paper-level concepts into framework bindings $(c,\mathcal{C})$ by selecting components from six typed slot families.} \textcolor{black}{The slot families are: \textsf{task} (prediction target and evaluation unit), \textsf{datasource} (how raw files map to the framework interfaces), \textsf{transform} (preprocessing and target construction, audited under a fit-on-train-only policy), \textsf{sequencer} (windowing policy such as length and stride), \textsf{model} (a wrapper implementing the predictor interface expected by $\mathcal{F}$), and \textsf{evaluator} (window-level or per-unit evaluation with a task-appropriate metric).} \textcolor{black}{When a paper requires a component that is not available in $\mathcal{F}$, $\mathcal{C}$ supplies the minimal contract-preserving extension (e.g., a new operator or wrapper); \textsc{Verify} audits the extension before the artifact is accepted.}

\textcolor{black}{This structure makes the outcome of \textsc{Map} explicit: the paper is instantiated as a concrete slot binding in $\mathcal{F}$, so different papers can be evaluated under identical windowing, splits, and metric definitions.}

\subsection{Assumption tracking and verification}
\label{subsec:assumptions}

\textcolor{black}{When a paper does not specify an implementation choice required to execute inside $\mathcal{F}$, the agent records that choice explicitly as an assumption record (e.g., whether evaluation is window-level versus per-unit).} \textcolor{black}{Each assumption record is the tuple}
\[
  \textcolor{black}{a_i \;=\; \bigl(s_i,\; e_i,\; v_i,\; j_i\;\bigr),}
\]

\textcolor{black}{where $s_i$ is the implicated slot family, $e_i$ points to the relevant paper span (or marks absent evidence), $v_i$ is the selected value/default, $j_i$ is its justification. The selected default must satisfy the task contract and benchmark conditions enforced by $\mathcal{F}$, making the choice explicit and auditable.}

\textcolor{black}{Verification gives $\mathcal{A}$ its operational meaning. The \emph{static} layer audits the configuration before optimization: typechecking catches task-contract mis-bindings, dry-runs catch shape and alignment errors, and leakage checks enforce fit-on-train-only preprocessing and target construction.}

\textcolor{black}{The \emph{sanity} layer then checks whether the assembled model can learn, following the spirit of Karpathy's training diagnostics \citep{karpathy2019recipe}: (i) an initial-loss check, (ii) a gradient-flow check, and (iii) a micro-batch memorization check. Together these form a model-correctness gate intended to catch wiring bugs that static checks cannot detect.}

\textcolor{black}{Finally, result matching compares paper-reported numbers with a reproduced run under the same windows, splits, and metric definitions. When verification fails with cause $r.\text{cause}$, we call an assumption record $a_i$ \emph{attributable} if $s_i \in \textsc{Slots}(r.\text{cause})$, the slots implicated by the failing check as reported by $\mathcal{F}$. This predicate drives repair: failures touching no slot represented in $\mathcal{A}$ are treated as paper-level or framework-level failures rather than assumption-repair failures.}

\textcolor{black}{Under this definition, unsuccessful reproductions become diagnostic artifacts: a human (or downstream agent) can inspect $\mathcal{A}$, override a responsible $a_i$, and re-run verification without reconstructing the pipeline. Successful reproductions inherit the same auditability property because every under-specified decision is localized in $\mathcal{A}$ and traceable to a concrete slot binding in $c$.}

\subsection{Intermediate artifacts and worked example}
\label{subsec:artifacts}

\textcolor{black}{The workflow emits inspectable intermediate artifacts rather than a single final code dump. This makes the pipeline auditable (a reader can trace each binding and inferred decision back to paper evidence) and ablatable (we can remove individual stages, such as assumption tracking or verification gates, and measure the effect on outcomes). Table~\ref{tab:workflow-artifacts} summarizes the artifacts emitted at each stage.}

\textcolor{black}{As a concrete example, consider a remaining useful life paper on N-CMAPSS that specifies an LSTM predictor, sliding windows, z-score normalization, and RMSE evaluation, but omits whether normalization is fit globally or on the training split only, and does not state the target clipping threshold. These omissions are exactly the kind of decisions recorded in $\mathcal{A}$.}

\textcolor{black}{In our workflow, \textsc{Ingest} indexes the paper; \textsc{Analyze} extracts the specified elements and emits assumption records for the missing choices; \textsc{Map} resolves the paper into a prognostics task contract, an N-CMAPSS datasource, a transform pipeline, a window sequencer, an LSTM wrapper, and an RMSE evaluator; and \textsc{Verify} either returns a benchmark-ready artifact or attributes failure back to implicated assumptions. The same structure applies to diagnostics papers, where omitted choices such as class grouping or evaluation grain are surfaced in $\mathcal{A}$ rather than buried inside free-form code.}

\section{Experiments}
\label{sec:experiments}

We evaluate framework-grounded agentic PHM reproduction along the axes introduced in Section~\ref{sec:intro}. Our evaluation protocol scores each artifact $A=(c,\mathcal{C},\mathcal{A},r)$ of Section~\ref{sec:method} along the five quality axes the method promises --- \emph{framework integration}, \emph{code development}, \emph{execution}, \emph{result matching}, and \emph{failure diagnosability} --- and layers three operational metrics over these axes: wall-clock time (\emph{Time2Build}), human effort (\emph{Easiness2Build}), and monetary cost (\emph{Cost2Build}). Unless stated otherwise, reported numbers are averaged over three independent runs per paper, with fresh LLM sessions and independent random seeds, to separate stochastic variation from systematic effects of framework grounding.

\subsection{Experimental setup}
\label{sec:setup}

\paragraph{Paper corpus.}
We curate a corpus of 16 published PHM papers, balanced by task family: 5 on \emph{diagnostics} (fault and mode classification, anomaly detection) and 11 on \emph{prognostics} (remaining useful life estimation, health-indicator inference). Papers are sampled to span the two datasets most commonly used in current PHM benchmarking work --- N-CMAPSS, NB14 --- and to cover both classical deep architectures (convolutional and recurrent backbones) and more recent attention- and state-space-based models. Dataset citations are collected in Appendix~\ref{app:corpus}. 

\paragraph{Framework instance.}
\textcolor{black}{All framework-coupled systems are evaluated against the same instance of $\mathcal{F}$, the shared framework described in Section~\ref{sec:method}. The slot families $(\textsf{task}, \textsf{datasource}, \textsf{transform}, \textsf{sequencer}, \textsf{model}, \textsf{evaluator})$, the fit-on-train-only policy for preprocessing and target construction, and the two evaluator granularities (window-level, per-unit) are held fixed across systems; only the binding $(c, \mathcal{C})$ differs.}

\paragraph{Baseline systems.}
\textcolor{black}{We compare our Framework-Coupled Agent (FCA) against baselines that vary access to the shared PHM framework, access to explicit framework documentation, and whether the workflow is staged or prompt-only, illistrated in Table~\ref{tab:generation-conditions-main}.}

\begin{table}[H]
  \centering
  \small
  \setlength{\tabcolsep}{4pt}
  \renewcommand{\arraystretch}{1.2}
  \caption{Implementation methods used in the evaluation.}
  \label{tab:generation-conditions-main}
  \begin{tabular*}{\linewidth}{@{\extracolsep{\fill}}p{3.0cm}p{6.8cm}cc@{}}
    \toprule
    Method & Description & Workflow & Output \\
    \midrule

    \textbf{Framework-Coupled Agent (FCA)} &
    Full staged workflow \textbf{inside $\mathcal{F}$} with
    \textbf{framework documentation enabled}
    (\textsc{Ingest}--\textsc{Analyze}--\textsc{Map}--\textsc{Implement}--\textsc{Verify}--\textsc{Report}),
    producing framework-compatible artifacts. &
    Staged &
    Framework \\

    \textbf{Agent-in-Framework (AiF)} &
    Prompt-only OpenCode baseline operating
    \textbf{inside $\mathcal{F}$} with
    \textbf{no framework documentation}. &
    Prompt-only &
    Framework \\

    \textbf{Agent-in-Framework with Docs (AiF-D)} &
    Prompt-only OpenCode baseline operating
    \textbf{inside $\mathcal{F}$} with
    \textbf{explicit framework documentation}. &
    Prompt-only &
    Framework \\

    \textbf{Standalone Agent (SA)} &
    Prompt-only OpenCode baseline operating
    \textbf{outside $\mathcal{F}$}
    and directly generating standalone implementations from the paper. &
    Prompt-only &
    Standalone \\

    \textbf{DeepCode (DC)} \citep{li2025deepcode} &
    System-specific paper-to-code baseline operating
    \textbf{outside $\mathcal{F}$},
    generating standalone repositories with custom training and evaluation logic. &
    System-specific &
    Standalone \\

    \bottomrule
  \end{tabular*}
\end{table}

\textcolor{black}{DC and SA serve as standalone repository generation baselines and measure paper-to-implementation performance without framework coupling.} \textcolor{black}{AiF and AiF-D fix the shared PHM framework target while remaining prompt-only; comparing AiF-D with AiF isolates the effect of explicit framework documentation.} \textcolor{black}{Comparing FCA against AiF/AiF-D isolates the contribution of the staged protocol (including assumption records and verification gates) under the same framework target.}

\paragraph{Baseline instantiation and comparison protocol.}
\textcolor{black}{All methods receive the same paper content and target the same 16-paper PHM corpus.}
\textcolor{black}{Framework-targeted methods (FCA, AiF, AiF-D) are evaluated directly inside $\mathcal{F}$ under identical task contracts, splits, windowing, and metric definitions.}
\textcolor{black}{Standalone methods (SA, DC) emit independent repositories; when we require shared benchmark conditions for cross-method comparison, we port their outputs post hoc into $\mathcal{F}$ using the same task contracts and evaluators.}
\textcolor{black}{For result matching, we additionally evaluate on a five-paper reference-implementation subset by running the authors' official code through the same $\mathcal{F}$ adapters and evaluators.}

\paragraph{Evaluation metrics.}
We score each artifact $A$ along the five quality axes promised by the Method section, and layer three operational metrics on top.
\begin{itemize}

  \item \textbf{Framework integration} (new axis): measured against our framework $\mathcal{F}$, this is the \emph{completion rate} for using $\mathcal{F}$ on a given paper --- whether the agent produces a valid binding $(c, \mathcal{C})$ inside $\mathcal{F}$. Concretely, the emitted $(c, \mathcal{C})$ must typecheck against the selected task contract, bind all six slot families to existing or contract-preserving components, and satisfy the leakage invariant. This axis is 0/1 per paper and precedes the remaining four by construction.
  \item \textbf{Code development}: an LLM-as-judge score on the PaperBench rubric adapted to $\mathcal{F}$, measuring whether $\mathcal{C}$ and the slot bindings faithfully encode the paper's architecture, loss, preprocessing, and regularization choices. 
\end{itemize}

\paragraph{Reporting convention.}
\textcolor{black}{For the five-paper reference-implementation subset, a lightweight human adjudication protocol attributes each failure to one of three causes: (i) agent error, (ii) paper under-specification beyond recoverable scope, or (iii) dataset mismatch.} \textcolor{black}{This separates reproducibility failures that are properties of the paper from failures that are properties of the implementation system.} \textcolor{black}{We also retain per-paper metadata on task family, dataset, architecture family, reference-implementation subset membership, number of emitted assumptions, and whether extensions $\mathcal{C}$ were required, enabling subgroup analyses by paper type.} Details on the evaluation protocol are given in Appendix~\ref{app:evaluation-setup}.

\paragraph{Release.}
We release the skills, instructions, and workflow code used in this study to reproduce the presented agentic workflow. Anonymous code repository for double-blind review: \url{https://github.com/picid-research/from-paper-to-benchmark}.

\subsection{Main results}
\label{sec:main-results}

\subsubsection{Framework integration and execution}
One central question is whether framework grounding increases the completion rate, the fraction of papers the agent successfully brings into $\mathcal{F}$ as a valid binding, and the downstream end-to-end execution rate.

\paragraph{Completion rate.} 
The binding status across the six slot families for AiF, AiF-D, and our FCA are compared in Figure~\ref{fig:binding-states}. SA exhibits a considerably lower success rate in slot binding, with 50\% or more missing bindings observed across all families.
Meanwhile, FCA achieves, averaged across families, 13.5\% more newly created binding states and 11\% fewer missing binding states compared to AiF. This demonstrates a gradual improvement when providing agents with framework documentation, with further gains achieved through our scaffolding.\\
Within the framework-based approaches, the completion rate in terms of successful bindings to all six slot families are AiF \textbf{7/16} (43.75\%), AiF-D \textbf{9/16} (56.25\%), and FCA \textbf{9/16} (56.25\%) according to Table~\ref{tab:bindings}. This suggests that explicit framework documentation and staged workflow improve framework integration quality by making slot identification, component reuse, and contract-preserving extension explicit.

\paragraph{Execution rate.}
Agentic, autonomeous execution rate for FCA is \textbf{14/16} (87.5\%), with two papers not successfully implemented: one publication lacks sufficient model specification details \citep{graves2023turbofan}, while another \citep{lin2024cata} relies on a dataset that the FCA claims is proprietary.
In contrast, the baseline execution rates are as follows: SA \textbf{14/16} (87.50\%), AiF \textbf{3/16} (18.75\%), AiF-D \textbf{8/16} (50.00\%), and DC \textbf{0/16} (0.00\%). These results suggest that an agent can be effective at generating executable standalone code; however, without framework documentation and scaffolding, it struggles to navigate the framework code and complete the tasks successfully.

\subsubsection{Code quality and implementation faithfulness}

Another central question is whether the framework grounding improves code quality, to which end we investigate LLM code ratings and code complexity.

\paragraph{Code rating.} The LLM-as-judge scores for SA, AiF, FCA, and DC are depicted in Figure \ref{fig:judge-rating} for ChatGPT 5.4 and Kimi 2.6 as judges. SA surpasses all other approaches in terms of mean rating.
DC and FCA score the lowest mean rating among the approaches for ChatGPT, resp. Kimi. These findings contrast with the execution rates, as f.e. the very low execution performance of DC (0.00\%) and AiF (18.75\%) is inconsistent with their high ratings from the LLM judges. Due to this discrepancy, the reliability of the LLM judges’ evaluations in our experiments remains uncertain. A more rigorous validation involving extensive human assessment would be required, but is beyond the scope of this work.

\begin{figure}[t]
  \centering
  \includegraphics[width=\linewidth]{\detokenize{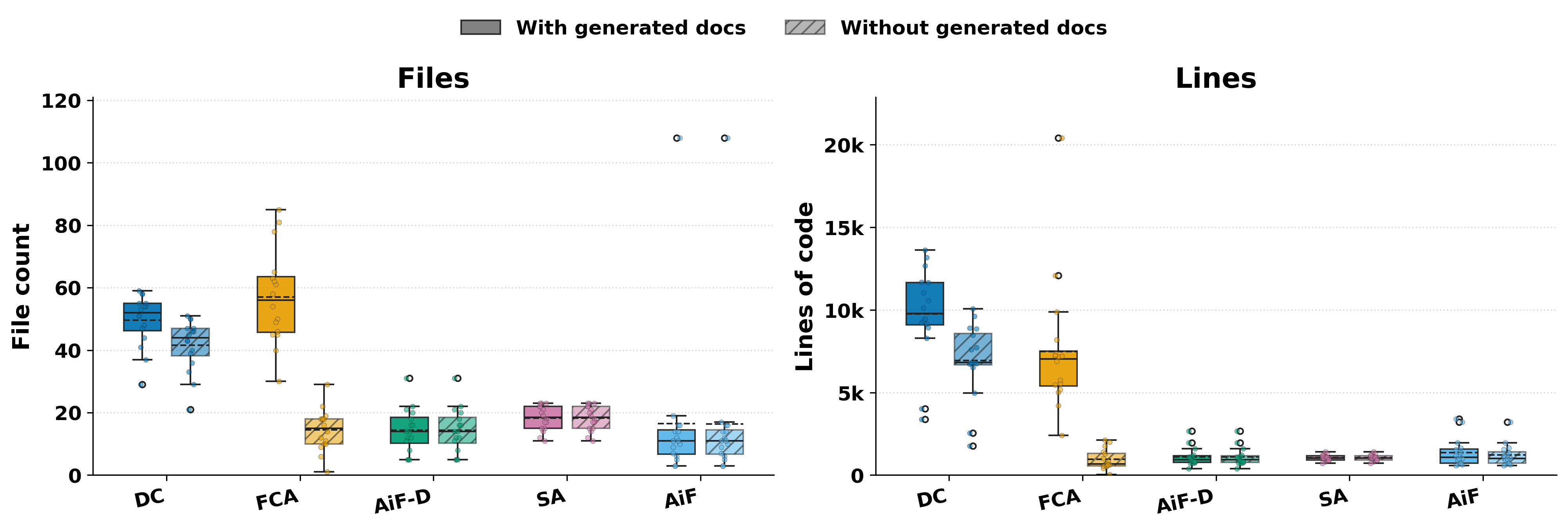}}
\caption{Distribution of generated files and generated code lines across methods, with box plots summarizing per-project variation for repositories evaluated in diff mode.}
  \label{fig:repository-statistics-by-group-with-without-gen-docs}
\end{figure}

\paragraph{Code complexity.} As a proxy for code complexity, Figure~\ref{fig:repository-statistics-by-group-with-without-gen-docs} reports the file and line counts of the generated repositories for the five approaches, both including and excluding markdown files. By comparing these two settings, we observe that DC produces the highest number of code files and lines among all methods. In contrast, FCA achieves the lowest code line count and second-lowest code file count, while generating primarily docs. From a code perspective, this reduction translates into substantially lower code review and maintainability effort.
The files and lines generated by both SA and AiF are predominantly code, likely due to the absence of explicit prompts encouraging documentation usage.

\subsubsection{Benchmark consistency}

To show the consistency of benchmark implementations, we report the results of 13 FCA model implementations (IMPL) together with CNN, MLP, Time-series Transformer (TST) and LSTM as benchmarks. Since comparisons are performed inside the shared framework, once the paper-specific implementation is integrated as a framework experiment, benchmark comparison becomes a configuration change rather than a new reproduction effort. For each paper, the task, datasource, split, and evaluator entries are reused, while only \texttt{model\_config} is changed from the generated implementation to a selected baseline model. When a baseline model is not available in the framework for the task category selected by the agent, the corresponding entry is left empty in Table~\ref{tab:results_test_mae_normalized}. \\
Out of the 16 papers, 3 implementations fail due to time-out restriction \citep{yamacli2025usageprofile}, largely missing model description \citep{graves2023turbofan} and the agent claiming data unavailability \citep{lin2024cata}. The prognostic benchmarking results for the 13 implemented papers by FCA are displayed as normalized mean absolute error in Table~\ref{tab:results_test_mae_normalized}, where the best-performing result is highlighted in grey and second best result is underlined. From the figure we derive that the average rank of the newly implemented method is the third highest, with two times exhibiting the best performance. Further, the model with the highest rank is LSTM, closely followed by TST. Note that for the two papers \citep{depater2023health, hsu2023residual}, the benchmark models cannot be generated due to the agent framing the case study as state forecasting, which is not yet supported by the framework. The missing table entries for papers \citep{wang2025battnn, zhang2023darwin, bosello2023charge} indicate that these benchmark models cannot be set-up without considerable human effort.\\
The diagnostics benchmark cannot be shown because to the following reasons: one diagnostics paper \citep{hsu2023residual} is wrongly implemented by FCA as prognostics task (in Table~\ref{tab:results_test_mae_normalized}), another paper \citep{sudarshan2025degradai} has both diagnostics and prognostics tasks, out of which FCA only implements the latter (in Table~\ref{tab:results_test_mae_normalized}). 
The remaining implemented diagnostics paper \citep{dong2025causal} generates implausable classification performances.

\begin{table}
\caption{Prognostics benchmark results for CNN, MLP, TST, LSTM, and implemented (IMPL) models, reported as normalized Mean Absolute Error (nMAE) scaled by \(10^{2}\) for each paper.}
\label{tab:results_test_mae_normalized}
\begin{adjustbox}{max width=\textwidth}
\begin{tabular}{llrrrrr}
\toprule
Paper & Dataset & CNN-1D & MLP & TST & LSTM & IMPL \\
\midrule
Roman et al.~\citep{roman2021batterysoh} & NB14 & 62.41 ± 13.79 & 129.66 ± 108.80 & 48.81 ± 6.89 & \underline{43.31 ± 0.40} & \cellcolor[gray]{0.85}\textbf{41.66 ± 0.22} \\
Wang et al.~\citep{wang2025battnn} & NB14 & - & - & \cellcolor[gray]{0.85}\textbf{1.19 ± 0.02} & \underline{1.20 ± 0.06} & 11.79 ± 0.35 \\
Sudarshan et al.~\citep{sudarshan2025degradai} & NB14 & 10.92 ± 0.12 & 10.09 ± 0.08 & - & \underline{8.96 ± 0.14} & \cellcolor[gray]{0.85}\textbf{6.69 ± 0.06} \\
Bosello et al.~\citep{bosello2023charge} & NB14 & - & \underline{16.55 ± 3.57} & - & \cellcolor[gray]{0.85}\textbf{5.47 ± 0.23} & 69.63 ± 40.73 \\
Xu et al.~\citep{xu2024sohtransformer} & NB14 & \underline{24.15 ± 2.67} & \cellcolor[gray]{0.85}\textbf{21.95} & 28.14 ± 8.27 & 33.40 ± 8.40 & 32.11 ± 18.05 \\
Tang et al.~\citep{tang2025vmdssa} & NB14 & 8.77 ± 2.02 & 6.95 ± 0.69 & \underline{5.82 ± 0.14} & \cellcolor[gray]{0.85}\textbf{5.42 ± 0.67} & 13.92 ± 1.95 \\
Wang et al.~\citep{wang2024bayesian} & NCMAPSS & 22.50 ± 10.88 & 86.04 ± 40.27 & \underline{5.46 ± 1.74} & \cellcolor[gray]{0.85}\textbf{0.25 ± 0.16} & 139.59 ± 202.97 \\
Zhang et al.~\citep{zhang2025cartnet} & NCMAPSS & 14.00 ± 1.37 & 9.92 ± 0.10 & \underline{7.74 ± 0.44} & \cellcolor[gray]{0.85}\textbf{5.88 ± 0.40} & 15.86 ± 0.50 \\
Zhang et al.~\citep{zhang2023darwin} & NCMAPSS & - & - & \cellcolor[gray]{0.85}\textbf{0.50 ± 0.05} & \underline{0.88 ± 0.40} & 5.01 ± 0.38 \\
Depater et al.~\citep{depater2023health} & NCMAPSS & 9.50 ± 1.25 & 18.14 ± 0.67 & \underline{5.14 ± 0.19} & \cellcolor[gray]{0.85}\textbf{4.73 ± 0.11} & \textbf{3137.38} \\
Hsu et al.~\citep{hsu2023residual} & NCMAPSS & - & - & - & - & \cellcolor[gray]{0.85}\textbf{4.07} \\
Cohen et al.~\citep{cohen2023faultprognosis} & NCMAPSS & 11.66 ± 2.19 & 14.49 ± 0.43 & \cellcolor[gray]{0.85}\textbf{8.08 ± 0.17} & \underline{8.16 ± 1.05} & - \\
\midrule
Average & & 20.49 & 34.86 & 12.32 & 10.70 & 316.15 \\
Average rank & & 3.38 & 3.22 & 1.89 & 1.82 & 2.91 \\
\bottomrule
\end{tabular}
\end{adjustbox}
\end{table}

\subsubsection{Time and token usage}
We measure the operational characterstics of computation time and token usage for the experiments. The aggregated number of tokens and agent session duration is shown in Figure~\ref{fig:token-statistics-compact}. The FCA has comparable token usage to DC above 20 millions, while AiF, AiF-D and SA consume considerably less tokens below 10 millions. Moreover, the session duration is highly correlated with tokens per run, reflecting that agentic mode has considerably longer session duration compared to the baselines. The elevated token usage and session duration is attributable to the scaffolding, leading to considerably more input tokens and reasoning steps.

\section{Conclusion}
\label{sec:limitations}

This work studies PHM as a representative case of a broader challenge in scientific machine learning: translating under-specified research papers into executable and benchmarkable implementations. We show that, in domains where benchmark outcomes depend strongly on hidden protocol decisions, reproducing a paper requires recovering not only model architectures, but also the execution semantics governing preprocessing, target construction, windowing, and evaluation. By coupling agentic paper understanding to a shared PHM framework with explicit slot bindings, assumption tracking, and verification hooks, our workflow transforms reproduction from isolated code synthesis into an executable, auditable, and benchmarkable process.
Across 16 PHM papers, framework grounding substantially improves the ability of agents to produce executable benchmark artifacts under shared evaluation conditions. The experiments further show that execution success and LLM-based code ratings can diverge significantly, suggesting that surface-level code quality is an insufficient proxy for reproducible scientific implementation. Instead, framework integration, protocol validity, and explicit handling of under-specification emerge as central dimensions for evaluating agentic paper reproduction.

At the same time, important limitations remain. Framework-grounded agents can still produce plausible but incorrect implementations when papers omit critical protocol details or when evaluation procedures remain ambiguous. Execution success therefore should not be interpreted as faithful reproduction. To mitigate this risk, our workflow exposes assumption records, slot-level bindings, and verification reports as first-class artifacts, and supplements automated evaluation with targeted human review on the reference-implementation subset.
These results further indicate that shared domain frameworks can make paper implementations more comparable by enforcing common execution semantics and making hidden protocol assumptions explicit. Future work should evaluate the approach on larger corpora and additional scientific ML domains, and investigate stronger verification mechanisms for handling severely under-specified papers.

{
\small
\bibliographystyle{plainnat}
\bibliography{references}

@misc{trofimova2024coderefine,
  title        = {{CodeRefine}: A Pipeline for Enhancing {LLM}-Generated Code Implementations of Research Papers},
  author       = {Trofimova, Ekaterina and Sataev, Emil and Jowhari, Abhijit Singh},
  year         = {2024},
  eprint       = {2408.13366},
  archivePrefix= {arXiv},
  primaryClass = {cs.CL},
  url          = {https://arxiv.org/abs/2408.13366}
}

@misc{seo2025paper2code,
  title        = {{Paper2Code}: Automating Code Generation from Scientific Papers in Machine Learning},
  author       = {Seo, Minju and Baek, Jinheon and Lee, Seongyun and Hwang, Sung Ju},
  year         = {2025},
  eprint       = {2504.17192},
  archivePrefix= {arXiv},
  primaryClass = {cs.CL},
  note         = {To appear in ICLR 2026},
  url          = {https://arxiv.org/abs/2504.17192}
}

@misc{zhao2025autoreproduce,
      title={AutoReproduce: Automatic AI Experiment Reproduction with Paper Lineage}, 
      author={Xuanle Zhao and Zilin Sang and Yuxuan Li and Qi Shi and Shuo Wang and Duzhen Zhang and Xu Han and Zhiyuan Liu and Maosong Sun},
      year={2025},
      eprint={2505.20662},
      archivePrefix={arXiv},
      primaryClass={cs.AI},
      url={https://arxiv.org/abs/2505.20662}, 
}

@misc{starace2025paperbench,
  title        = {{PaperBench}: Evaluating {AI}'s Ability to Replicate {AI} Research},
  author       = {Starace, Giulio and Jaffe, Oliver and Sherburn, Dane and Aung, James and Chan, Jun Shern and Maksin, Leon and Dias, Rachel and Mays, Evan and Kinsella, Benjamin and Thompson, Wyatt and Heidecke, Johannes and Glaese, Amelia and Patwardhan, Tejal},
  year         = {2025},
  eprint       = {2504.01848},
  archivePrefix= {arXiv},
  primaryClass = {cs.AI},
  url          = {https://arxiv.org/abs/2504.01848}
}

@misc{siegel2024corebench,
      title={CORE-Bench: Fostering the Credibility of Published Research Through a Computational Reproducibility Agent Benchmark}, 
      author={Zachary S. Siegel and Sayash Kapoor and Nitya Nagdir and Benedikt Stroebl and Arvind Narayanan},
      year={2024},
      eprint={2409.11363},
      archivePrefix={arXiv},
      primaryClass={cs.CL},
      url={https://arxiv.org/abs/2409.11363}, 
}

@misc{kim2025autoexperiment,
  title        = {From Reproduction to Replication: Evaluating Research Agents with Progressive Code Masking},
  author       = {Kim, Gyeongwon James and Wilf, Alex and Morency, Louis-Philippe and Fried, Daniel},
  year         = {2025},
  eprint       = {2506.19724},
  archivePrefix= {arXiv},
  primaryClass = {cs.AI},
  url          = {https://arxiv.org/abs/2506.19724}
}

@misc{xiang2025scireplicate,
  title        = {{SciReplicate-Bench}: Benchmarking {LLMs} in Agent-driven Algorithmic Reproduction from Research Papers},
  author       = {Xiang, Yanzheng and Yan, Hanqi and Ouyang, Shuyin and Gui, Lin and He, Yulan},
  year         = {2025},
  eprint       = {2504.00255},
  archivePrefix= {arXiv},
  primaryClass = {cs.CL},
  url          = {https://arxiv.org/abs/2504.00255}
}

@misc{hua2025researchcodebench,
      title={ResearchCodeBench: Benchmarking LLMs on Implementing Novel Machine Learning Research Code}, 
      author={Tianyu Hua and Harper Hua and Violet Xiang and Benjamin Klieger and Sang T. Truong and Weixin Liang and Fan-Yun Sun and Nick Haber},
      year={2025},
      eprint={2506.02314},
      archivePrefix={arXiv},
      primaryClass={cs.AI},
      url={https://arxiv.org/abs/2506.02314}, 
}

@misc{tian2024scicode,
  title        = {{SciCode}: A Research Coding Benchmark Curated by Scientists},
  author       = {Tian, Minyang and others},
  year         = {2024},
  eprint       = {2407.13168},
  archivePrefix= {arXiv},
  primaryClass = {cs.AI},
  url          = {https://arxiv.org/abs/2407.13168}
}

@misc{vonhahn2022phmreproducibility,
  title        = {Computational Reproducibility Within Prognostics and Health Management},
  author       = {von Hahn, Tim and Mechefske, Chris K.},
  year         = {2022},
  eprint       = {2205.15489},
  archivePrefix= {arXiv},
  primaryClass = {cs.DL},
  url          = {https://arxiv.org/abs/2205.15489}
}

@misc{liu2025tsagent,
      title={TS-Agent: Understanding and Reasoning Over Raw Time Series via Iterative Insight Gathering}, 
      author={Penghang Liu and Elizabeth Fons and Annita Vapsi and Mohsen Ghassemi and Svitlana Vyetrenko and Daniel Borrajo and Vamsi K. Potluru and Manuela Veloso},
      year={2026},
      eprint={2510.07432},
      archivePrefix={arXiv},
      primaryClass={cs.AI},
      url={https://arxiv.org/abs/2510.07432}, 
}

@misc{gu2025argos,
  title        = {Argos: Agentic Time-Series Anomaly Detection with Autonomous Rule Generation via Large Language Models},
  author       = {Gu, Yile and others},
  year         = {2025},
  eprint       = {2501.14170},
  archivePrefix= {arXiv},
  primaryClass = {cs.LG},
  url          = {https://arxiv.org/abs/2501.14170}
}

@misc{ji2025autoiad,
  title        = {{AutoIAD}: Manager-Driven Multi-Agent Collaboration for Automated Industrial Anomaly Detection},
  author       = {Ji, Dongwei and Hu, Bingzhang and Zhou, Yi},
  year         = {2025},
  eprint       = {2508.05503},
  archivePrefix= {arXiv},
  primaryClass = {cs.CV},
  url          = {https://arxiv.org/abs/2508.05503}
}

@article{teubert2023progpy,
  title        = {{ProgPy}: Python Packages for Prognostics and Health Management of Engineering Systems},
  author       = {Teubert, Christopher and Jarvis, Katelyn and Corbetta, Matteo and Kulkarni, Chetan and Daigle, Matthew},
  journal      = {Journal of Open Source Software},
  year         = {2023},
  volume       = {8},
  number       = {87},
  pages        = {5099},
  doi          = {10.21105/joss.05099},
  url          = {https://doi.org/10.21105/joss.05099}
}

@misc{anonymous2026framework,
  title        = {PICID: A Modular Evaluation Infrastructure for Reproducible PHM Across Tasks and Domains},
  author       = {{Anonymous}},
  year         = {2026},
  note         = {Under anonymous review; citation details blinded for double-blind compliance}
}

@inproceedings{li2025phmvibench,
  title        = {{PHM-Vibench}: A Unified and Factory-Style Vibration Benchmarking Framework for the Foundation Model Era},
  author       = {Li, Qi and Chen, Bojian and Li, Xuan and Chen, Qitong and Chen, Liang and Shen, Changqing and Lu, Lu and Qin, Zhaoye and Chu, Fulei},
  booktitle    = {Proceedings of the Asia Pacific Conference of the {PHM} Society 2025},
  year         = {2025},
  url          = {https://papers.phmsociety.org/index.php/phmap/article/view/4303}
}

@misc{karpathy2019recipe,
    author       = {Andrej Karpathy},
    title        = {A Recipe for Training Neural Networks},   
    year         = {2019},
    howpublished = {\url{https://karpathy.github.io/2019/04/25/recipe/}},    note = {Blog post; accessed 2026-04-28} 
}

@misc{jiang_aide_2025,
	title = {{AIDE}: {AI}-{Driven} {Exploration} in the {Space} of {Code}},
	shorttitle = {{AIDE}},
	url = {http://arxiv.org/abs/2502.13138},
	doi = {10.48550/arXiv.2502.13138},
	urldate = {2026-04-29},
	publisher = {arXiv},
	author = {Jiang, Zhengyao and Schmidt, Dominik and Srikanth, Dhruv and Xu, Dixing and Kaplan, Ian and Jacenko, Deniss and Wu, Yuxiang},
	month = feb,
	year = {2025},
	note = {arXiv:2502.13138},
}

@misc{gottweis_towards_2025,
	title = {Towards an {AI} co-scientist},
	url = {http://arxiv.org/abs/2502.18864},
	doi = {10.48550/arXiv.2502.18864},
	urldate = {2026-04-29},
	publisher = {arXiv},
	author = {Gottweis, Juraj and Weng, Wei-Hung and Daryin, Alexander and Tu, Tao and Palepu, Anil and Sirkovic, Petar and Myaskovsky, Artiom and Weissenberger, Felix and Rong, Keran and Tanno, Ryutaro and Saab, Khaled and Popovici, Dan and Blum, Jacob and Zhang, Fan and Chou, Katherine and Hassidim, Avinatan and Gokturk, Burak and Vahdat, Amin and Kohli, Pushmeet and Matias, Yossi and Carroll, Andrew and Kulkarni, Kavita and Tomasev, Nenad and Guan, Yuan and Dhillon, Vikram and Vaishnav, Eeshit Dhaval and Lee, Byron and Costa, Tiago R. D. and Penadés, José R. and Peltz, Gary and Xu, Yunhan and Pawlosky, Annalisa and Karthikesalingam, Alan and Natarajan, Vivek},
	month = feb,
	year = {2025},
	note = {arXiv:2502.18864},
}

@misc{schmidgall_agent_2025,
	title = {Agent {Laboratory}: {Using} {LLM} {Agents} as {Research} {Assistants}},
	shorttitle = {Agent {Laboratory}},
	url = {http://arxiv.org/abs/2501.04227},
	doi = {10.48550/arXiv.2501.04227},
	urldate = {2026-04-29},
	publisher = {arXiv},
	author = {Schmidgall, Samuel and Su, Yusheng and Wang, Ze and Sun, Ximeng and Wu, Jialian and Yu, Xiaodong and Liu, Jiang and Moor, Michael and Liu, Zicheng and Barsoum, Emad},
	month = jun,
	year = {2025},
	note = {arXiv:2501.04227},
}

@misc{lu_ai_2024,
	title = {The {AI} {Scientist}: {Towards} {Fully} {Automated} {Open}-{Ended} {Scientific} {Discovery}},
	shorttitle = {The {AI} {Scientist}},
	url = {http://arxiv.org/abs/2408.06292},
	doi = {10.48550/arXiv.2408.06292},
	urldate = {2026-04-29},
	publisher = {arXiv},
	author = {Lu, Chris and Lu, Cong and Lange, Robert Tjarko and Foerster, Jakob and Clune, Jeff and Ha, David},
	month = sep,
	year = {2024},
	note = {arXiv:2408.06292},
}

@misc{lin2025autop2c,
      title={AutoP2C: An LLM-Based Agent Framework for Code Repository Generation from Multimodal Content in Academic Papers}, 
      author={Zijie Lin and Yiqing Shen and Qilin Cai and He Sun and Jinrui Zhou and Mingjun Xiao},
      year={2025},
      eprint={2504.20115},
      archivePrefix={arXiv},
      primaryClass={cs.SE},
      url={https://arxiv.org/abs/2504.20115}, 
}

@article{teubert_generic_2020,
	title = {A {Generic} {Software} {Architecture} for {Prognostics} ({GSAP})},
	volume = {8},
	issn = {2153-2648, 2153-2648},
	url = {https://papers.phmsociety.org/index.php/ijphm/article/view/2618},
	doi = {10.36001/ijphm.2017.v8i2.2618},
	number = {2},
	urldate = {2026-04-30},
	journal = {International Journal of Prognostics and Health Management},
	author = {Teubert, Christopher and J. Daigle, Matthew and Sankararaman, Shankar and Goebel, Kai and Watkins, Jason},
	month = nov,
	year = {2020},
}

@misc{leanh2026treat,
      title={Do Not Treat Code as Natural Language: Implications for Repository-Level Code Generation and Beyond}, 
      author={Minh Le-Anh and Huyen Nguyen and Khanh An Tran and Nam Le Hai and Linh Ngo Van and Nghi D. Q. Bui and Bach Le},
      year={2026},
      eprint={2602.11671},
      archivePrefix={arXiv},
      primaryClass={cs.SE},
      url={https://arxiv.org/abs/2602.11671}, 
}

@misc{zhou_time_2026,
	title = {Time {Series} {Reasoning} via {Process}-{Verifiable} {Thinking} {Data} {Synthesis} and {Scheduling} for {Tailored} {LLM} {Reasoning}},
	url = {http://arxiv.org/abs/2602.07830},
	doi = {10.48550/arXiv.2602.07830},
	urldate = {2026-04-30},
	publisher = {arXiv},
	author = {Zhou, Jiahui and Li, Dan and Li, Boxin and Zhang, Xiao and Meng, Erli and Li, Lin and Chen, Zhuomin and Lou, Jian and Ng, See-Kiong},
	month = feb,
	year = {2026},
	note = {arXiv:2602.07830},
}

@misc{wu_timeart_2026,
	title = {{TimeART}: {Towards} {Agentic} {Time} {Series} {Reasoning} via {Tool}-{Augmentation}},
	shorttitle = {{TimeART}},
	url = {http://arxiv.org/abs/2601.13653},
	doi = {10.48550/arXiv.2601.13653},
	urldate = {2026-04-30},
	publisher = {arXiv},
	author = {Wu, Xingjian and Lu, Junkai and Li, Zhengyu and Qiu, Xiangfei and Hu, Jilin and Guo, Chenjuan and Jensen, Christian S. and Yang, Bin},
	month = jan,
	year = {2026},
	note = {arXiv:2601.13653},
}

@article{ye_ts-reasoner_2025,
	title = {{TS}-{Reasoner}: {Domain}-{Oriented} {Time} {Series} {Inference} {Agents} for {Reasoning} and {Automated} {Analysis}},
	issn = {2835-8856},
	shorttitle = {{TS}-{Reasoner}},
	url = {https://openreview.net/forum?id=yhy7Vigjcf},
	language = {en},
	urldate = {2026-04-30},
	journal = {Transactions on Machine Learning Research},
	author = {Ye, Wen and Yang, Wei and Cao, Defu and Zhang, Yizhou and Tang, Lumingyuan and Cai, Jie and Liu, Yan},
	month = dec,
	year = {2025},
}

@misc{li2025deepcode,
      title={DeepCode: Open Agentic Coding}, 
      author={Zongwei Li and Zhonghang Li and Zirui Guo and Xubin Ren and Chao Huang},
      year={2025},
      eprint={2512.07921},
      archivePrefix={arXiv},
      primaryClass={cs.SE},
      url={https://arxiv.org/abs/2512.07921}, 
}

@article{bosello2023charge,
  title        = {To Charge or to Sell? {EV} Pack Useful Life Estimation via {LSTMs}, {CNNs}, and Autoencoders},
  author       = {Bosello, Michael and Falcomer, Carlo and Rossi, Claudio and Pau, Giovanni},
  journal      = {Energies},
  year         = {2023},
  volume       = {16},
  number       = {6},
  pages        = {2837},
  doi          = {10.3390/en16062837},
  url          = {https://doi.org/10.3390/en16062837}
}

@article{lin2024cata,
  title        = {Channel attention \& temporal attention based temporal convolutional network: A dual attention framework for remaining useful life prediction of the aircraft engines},
  author       = {Lin, Lin and Wu, Jinlei and Fu, Song and Zhang, Sihao and Tong, Changsheng and Zu, Lizheng},
  journal      = {Advanced Engineering Informatics},
  year         = {2024},
  volume       = {60},
  pages        = {102372},
  doi          = {10.1016/j.aei.2024.102372},
  url          = {https://doi.org/10.1016/j.aei.2024.102372}
}

@article{depater2023health,
  title        = {Developing health indicators and {RUL} prognostics for systems with few failure instances and varying operating conditions using a {LSTM} autoencoder},
  author       = {{de Pater}, Ingeborg and Mitici, Mihaela},
  journal      = {Engineering Applications of Artificial Intelligence},
  year         = {2023},
  volume       = {117},
  pages        = {105582},
  doi          = {10.1016/j.engappai.2022.105582},
  url          = {https://doi.org/10.1016/j.engappai.2022.105582}
}

@article{xu2024sohtransformer,
  title        = {State of health estimation for lithium-ion batteries based on incremental capacity analysis and Transformer modeling},
  author       = {Xu, Zhaofan and Chen, Zewang and Yang, Lin and Zhang, Songyuan},
  journal      = {Applied Soft Computing},
  year         = {2024},
  volume       = {165},
  pages        = {112072},
  doi          = {10.1016/j.asoc.2024.112072},
  url          = {https://doi.org/10.1016/j.asoc.2024.112072}
}

@article{roman2021batterysoh,
  title        = {Machine learning pipeline for battery state-of-health estimation},
  author       = {Roman, Darius and Saxena, Saurabh and Robu, Valentin and Pecht, Michael and Flynn, David},
  journal      = {Nature Machine Intelligence},
  year         = {2021},
  volume       = {3},
  number       = {5},
  pages        = {447--456},
  doi          = {10.1038/s42256-021-00312-3},
  url          = {https://doi.org/10.1038/s42256-021-00312-3}
}

@article{yamacli2025usageprofile,
  title        = {Battery Usage-Profile Classification via Multimodal Vision Transformer With Feature-Level Fusion and Ensemble Method},
  author       = {Yamacli, Volkan},
  journal      = {IEEE Access},
  year         = {2025},
  volume       = {13},
  pages        = {174664--174683},
  doi          = {10.1109/ACCESS.2025.3618776},
  url          = {https://doi.org/10.1109/ACCESS.2025.3618776}
}

@article{wang2025battnn,
  title        = {Inherently Interpretable Physics-Informed Neural Network for Battery Modeling and Prognosis},
  author       = {Wang, Fujin and Zhi, Quanquan and Zhao, Zhibin and Zhai, Zhi and Liu, Yingkai and Xi, Huan and Wang, Shibin and Chen, Xuefeng},
  journal      = {IEEE Transactions on Neural Networks and Learning Systems},
  year         = {2025},
  volume       = {36},
  number       = {1},
  pages        = {1145--1159},
  doi          = {10.1109/TNNLS.2023.3329368},
  url          = {https://doi.org/10.1109/TNNLS.2023.3329368}
}

@article{tang2025vmdssa,
  title        = {Lithium-ion battery {RUL} prediction based on optimized {VMD-SSA-PatchTST} algorithm},
  author       = {Tang, Pei and Qiu, Zetao and Yao, Zhongran and Pan, Jiahao and Cheng, Dashuai and Gu, Xiaoyong and Sun, Changcheng},
  journal      = {Scientific Reports},
  year         = {2025},
  volume       = {15},
  pages        = {26824},
  doi          = {10.1038/s41598-025-11934-7},
  url          = {https://doi.org/10.1038/s41598-025-11934-7}
}

@article{sudarshan2025degradai,
  title        = {{DegradAI}: A scalable framework for early battery health diagnosis from limited data},
  author       = {Sudarshan, Meghana and Vajja, Jaya Vikeswara Rao and Tomar, Vikas},
  journal      = {npj Clean Energy},
  year         = {2025},
  volume       = {1},
  number       = {1},
  pages        = {8},
  doi          = {10.1038/s44406-025-00008-2},
  url          = {https://doi.org/10.1038/s44406-025-00008-2}
}

@article{wang2024bayesian,
  title        = {Bayesian gated-transformer model for risk-aware prediction of aero-engine remaining useful life},
  author       = {Wang, Zili and Zhang, Yiming and Qiu, Lemiao and Zhang, Shuyou and Choi, Joo-Ho and Xiang, Feifan},
  journal      = {Expert Systems with Applications},
  year         = {2024},
  volume       = {238},
  number       = {Part B},
  pages        = {121859},
  doi          = {10.1016/j.eswa.2023.121859},
  url          = {https://doi.org/10.1016/j.eswa.2023.121859}
}

@article{zhang2025cartnet,
  title        = {{CART-Net}: A Causal Adaptive Residual Time Network for Remaining Useful Life Prediction of Aeroengines Under Varying Operating Conditions},
  author       = {Zhang, Yujie and Du, Mingyang and Miao, Qiang},
  journal      = {IEEE Internet of Things Journal},
  year         = {2025},
  volume       = {12},
  number       = {23},
  pages        = {50291--50303},
  doi          = {10.1109/JIOT.2025.3610545},
  url          = {https://doi.org/10.1109/JIOT.2025.3610545}
}

@inproceedings{graves2023turbofan,
  title        = {Data-Driven Prognostics and Diagnostics of Industrial Machinery -- A Turbofan Engine Case Study},
  author       = {Graves, Russell and Pankaj, Peeyush and Kuruvilla, Vineet J. and Johnson, Rachel and Inoue, Michio},
  booktitle    = {Proceedings of the Asia Pacific Conference of the {PHM} Society},
  year         = {2023},
  doi          = {10.36001/phmap.2023.v4i1.3690},
  url          = {https://doi.org/10.36001/phmap.2023.v4i1.3690}
}

@inproceedings{dong2025causal,
  title        = {Causal Inference-Based Fault Diagnosis and Abnormal Degradation Detection for Aero-Engine},
  author       = {Dong, Kunyu and Xu, Dan and Zeng, Zhaoyang and Zhu, Qingyu},
  booktitle    = {Proceedings of the 2025 International Conference on Equipment Intelligent Operation and Maintenance},
  year         = {2025},
  pages        = {1422--1428},
  doi          = {10.1109/ICEIOM65271.2025.11239779},
  url          = {https://doi.org/10.1109/ICEIOM65271.2025.11239779}
}

@article{cohen2023faultprognosis,
  title        = {Fault Prognosis of Turbofan Engines: Eventual Failure Prediction and Remaining Useful Life Estimation},
  author       = {Cohen, Joseph and Huan, Xun and Ni, Jun},
  journal      = {International Journal of Prognostics and Health Management},
  year         = {2023},
  volume       = {14},
  number       = {2},
  doi          = {10.36001/ijphm.2023.v14i2.3486},
  url          = {https://doi.org/10.36001/ijphm.2023.v14i2.3486}
}

@inproceedings{hsu2023residual,
  title        = {A Comparison of Residual-based Methods on Fault Detection},
  author       = {Hsu, Chi-Ching and Frusque, Gaetan and Fink, Olga},
  booktitle    = {Annual Conference of the {PHM} Society},
  year         = {2023},
  doi          = {10.36001/phmconf.2023.v15i1.3444},
  url          = {https://doi.org/10.36001/phmconf.2023.v15i1.3444}
}

@article{zhang2023darwin,
  title        = {A Data Augmentation Boosted Dual Informer Framework for the Performance Degradation Prediction of Aero-Engines},
  author       = {Zhang, Zhiyao and Chen, Pengpeng and Xing, Chenguang and Liu, Bo and Wang, Ruo and Li, Longxiao and Chen, Xiaohui and Zio, Enrico},
  journal      = {IEEE Sensors Journal},
  year         = {2023},
  doi          = {10.1109/JSEN.2023.3269030},
  url          = {https://doi.org/10.1109/JSEN.2023.3269030}
}

@misc{qiu2026prbench,
      title={PRBench: End-to-end Paper Reproduction in Physics Research}, 
      author={Shi Qiu and Junyi Deng and Yiwei Deng and Haoran Dong and Jieyu Fu and Mao Li and Zeyu Li and Zhaolong Zhang and Huiwen Zheng and Leidong Bao and Anqi Lv and Zihan Mo and Yadi Niu and Yiyang Peng and Yu Tian and Yili Wang and Ziyu Wang and Zi-Yu Wang and Jiashen Wei and Liuheng Wu and Aoran Xue and Leyi Yang and Guanglu Yuan and Xiarui Zhan and Jingjun Zhang and Zifan Zheng and Pengfei Liu and Linrui Zhen and Kaiyang Li and Qichang Li and Ziheng Zhou and Guo-En Nian and Yunwei Xiao and Qing-Hong Cao and Linjie Dai and Xu Feng and Peng Gao and Ying Gu and Chang Liu and Jia Liu and Ming-xing Luo and Yan-Qing Ma and Liang-You Peng and Huichao Song and Shufeng Wang and Chenxu Wang and Tao Wang and Yi-Nan Wang and Chengyin Wu and Pengwei Zhao and Hua Xing Zhu},
      year={2026},
      eprint={2603.27646},
      archivePrefix={arXiv},
      primaryClass={cs.CL},
      url={https://arxiv.org/abs/2603.27646}, 
}

@misc{marker,
  title        = {{Marker}: Convert documents to Markdown, {JSON}, chunks, and {HTML}},
  author       = {{Datalab}},
  year         = {2026},
  howpublished = {\url{https://github.com/datalab-to/marker}},
  note         = {Software repository; accessed 2026-05-27},
  url          = {https://github.com/datalab-to/marker}
}
}

\appendix

\clearpage

\section{Paper corpus summary}
\label{app:corpus}

Our corpus contains 16 PHM papers selected to stress paper-to-implementation under shared benchmark conditions rather than to maximize topical breadth alone. The design criteria are: (i) coverage of the two main PHM task families considered in this paper, diagnostics and prognostics; (ii) representation of datasets common in current PHM benchmarking practice; and (iii) a mix of established and recent architecture families. We therefore prioritize papers that can plausibly be mapped to the two popular PHM benchmark datasets supported by the framework, N-CMAPSS and NB14, while covering a diverse set of modelling approaches and research directions.

The corpus is intentionally heterogeneous. It includes classical machine-learning pipelines, recurrent and convolutional neural networks, Transformer-based architectures, physics-informed models, causal models, autoencoder-based health indicators, and hybrid decomposition or signal-processing pipelines. Several papers are under-specified in ways that are commonly observed in applied PHM literature.

The paper selection spans publication years and venues with the intention that the corpus includes established PHM modeling patterns and recently developed methods. We record whether each paper provides code, which datasets are used, and whether the implementation can target a supported benchmark dataset. Table~\ref{tab:paper-metadata} summarizes the paper corpus and lists the per-paper metadata used to interpret the implementation runs.

\noindent\begin{minipage}{\textwidth}
  \refstepcounter{table}\label{tab:paper-metadata}
  \small
  \noindent\textbf{Table~\thetable:} Paper-level corpus metadata. References are shown as first author et al.\ with citation; datasets are the public benchmark datasets targeted by the framework implementation.
  \par\medskip
  \centering
  \scriptsize
  \setlength{\tabcolsep}{3pt}
  \begin{tabular}{@{}p{0.16\textwidth}p{0.18\textwidth}p{0.24\textwidth}p{0.27\textwidth}p{0.09\textwidth}@{}}
    \toprule
    Reference & Run Name & Model family & Research topic & Datasets \\
    \midrule
    Bosello et al.~\citep{bosello2023charge} & ev-battery-rul & LSTM/CNN/autoencoder & EV pack useful-life estimation & UNIBO-PL, NB14 \\
    Lin et al.~\citep{lin2024cata} & cata-tcn-engine-rul & dual-attention TCN & aero-engine remaining useful life prediction & N-CMAPSS \\
    de Pater et al.~\citep{depater2023health} & lstm-ae-health-rul & LSTM autoencoder & health indicators and RUL with few failure instances & N-CMAPSS \\
    Xu et al.~\citep{xu2024sohtransformer} & ica-transformer-soh & ICA + Transformer & lithium-ion battery state-of-health estimation & NB14 \\
    Roman et al.~\citep{roman2021batterysoh} & battery-soh-pipeline & machine-learning pipeline & battery state-of-health estimation & NB14 \\
    Yamacli et al.~\citep{yamacli2025usageprofile} & battery-usage-vit & multimodal ViT ensemble & battery usage-profile classification & NB14 \\
    Wang et al.~\citep{wang2025battnn} & battnn-pinn-eod & physics-informed neural network & battery modeling and prognosis & NB14 \\
    Tang et al.~\citep{tang2025vmdssa} & vmd-ssa-patchtst-battery-rul & VMD-SSA-PatchTST & lithium-ion battery RUL prediction & NB14 \\
    Sudarshan et al.~\citep{sudarshan2025degradai} & degradai-battery-diagnosis & DegradAI framework & early battery health diagnosis from limited data & NB14 \\
    Wang et al.~\citep{wang2024bayesian} & bayesian-gated-transformer-rul & Bayesian gated Transformer & risk-aware aero-engine RUL prediction & N-CMAPSS \\
    Zhang et al.~\citep{zhang2025cartnet} & cart-net-aeroengine-rul & causal adaptive residual time network & aero-engine RUL under varying operating conditions & N-CMAPSS \\
    Graves et al.~\citep{graves2023turbofan} & turbofan-prognostics-diagnostics & data-driven PHM pipeline & turbofan-engine prognostics and diagnostics & N-CMAPSS \\
    Dong et al.~\citep{dong2025causal} & causal-fault-diagnosis-aeroengine & causal inference & fault diagnosis and abnormal degradation detection & N-CMAPSS \\
    Cohen et al.~\citep{cohen2023faultprognosis} & turbofan-fault-prognosis-rul & fault-prognosis models & eventual failure prediction and RUL estimation & N-CMAPSS \\
    Hsu et al.~\citep{hsu2023residual} & residual-fault-detection & residual-based methods & fault-detection method comparison & N-CMAPSS \\
    Zhang et al.~\citep{zhang2023darwin} & dual-informer-aeroengine-rul & dual Informer + augmentation & aero-engine performance degradation prediction & N-CMAPSS \\
    \bottomrule
  \end{tabular}
\end{minipage}

For each paper, we retain a metadata sheet recording: anonymized paper identifier, task family, dataset, architecture family, code-availability status, whether framework extensions $\mathcal{C}$ were required, number of emitted assumptions, and final verification outcome.

The research topics in Table~\ref{tab:paper-metadata} expose common problems in PHM reproduction: state-of-health estimation, remaining-useful-life prediction, fault detection, degradation diagnosis, and health-indicator construction often require different assumptions about targets, windows, labels, metrics, and evaluation protocols. The corpus therefore stresses the full pipeline implementation, and is designed to test the agent's ability to translate diverse PHM research topics into explicit, auditable framework tasks and bindings.

\begin{table}[t]
\caption{Framework binding. Bindings are categorized in created new (C), uses existing (E), and missing (M).}
\label{tab:bindings}
\centering
\begin{adjustbox}{width=\textwidth}
\begin{tabular}{llrcccccc}
\toprule
Method & Run Name & Created & Exp. & Data. & Eval. & Model Cfg. & Task Def. & Model \\
\midrule
AiF & battery-soh-pipeline & 0 & \textcolor{red!70!black}{\textbf{M}} & \textcolor{red!70!black}{\textbf{M}} & \textcolor{red!70!black}{\textbf{M}} & \textcolor{red!70!black}{\textbf{M}} & \textcolor{red!70!black}{\textbf{M}} & \textcolor{red!70!black}{\textbf{M}} \\
AiF & battery-usage-vit & 7 & \textcolor{green!45!black}{\textbf{C}} & \textcolor{green!45!black}{\textbf{C}} & \textcolor{yellow!45!black}{\textbf{E}} & \textcolor{green!45!black}{\textbf{C}} & \textcolor{green!45!black}{\textbf{C}} & \textcolor{green!45!black}{\textbf{C}} \\
AiF & battnn-pinn-eod & 5 & \textcolor{green!45!black}{\textbf{C}} & \textcolor{green!45!black}{\textbf{C}} & \textcolor{green!45!black}{\textbf{C}} & \textcolor{red!70!black}{\textbf{M}} & \textcolor{green!45!black}{\textbf{C}} & \textcolor{green!45!black}{\textbf{C}} \\
AiF & bayesian-gated-transformer-rul & 5 & \textcolor{green!45!black}{\textbf{C}} & \textcolor{yellow!45!black}{\textbf{E}} & \textcolor{green!45!black}{\textbf{C}} & \textcolor{green!45!black}{\textbf{C}} & \textcolor{yellow!45!black}{\textbf{E}} & \textcolor{green!45!black}{\textbf{C}} \\
AiF & cart-net-aeroengine-rul & 9 & \textcolor{green!45!black}{\textbf{C}} & \textcolor{green!45!black}{\textbf{C}} & \textcolor{yellow!45!black}{\textbf{E}} & \textcolor{green!45!black}{\textbf{C}} & \textcolor{green!45!black}{\textbf{C}} & \textcolor{green!45!black}{\textbf{C}} \\
AiF & cata-tcn-engine-rul & 9 & \textcolor{green!45!black}{\textbf{C}} & \textcolor{green!45!black}{\textbf{C}} & \textcolor{yellow!45!black}{\textbf{E}} & \textcolor{green!45!black}{\textbf{C}} & \textcolor{green!45!black}{\textbf{C}} & \textcolor{green!45!black}{\textbf{C}} \\
AiF & causal-fault-diagnosis-aeroengine & 0 & \textcolor{red!70!black}{\textbf{M}} & \textcolor{red!70!black}{\textbf{M}} & \textcolor{red!70!black}{\textbf{M}} & \textcolor{red!70!black}{\textbf{M}} & \textcolor{red!70!black}{\textbf{M}} & \textcolor{red!70!black}{\textbf{M}} \\
AiF & degradai-battery-diagnosis & 0 & \textcolor{red!70!black}{\textbf{M}} & \textcolor{red!70!black}{\textbf{M}} & \textcolor{red!70!black}{\textbf{M}} & \textcolor{red!70!black}{\textbf{M}} & \textcolor{red!70!black}{\textbf{M}} & \textcolor{red!70!black}{\textbf{M}} \\
AiF & dual-informer-aeroengine-rul & 7 & \textcolor{green!45!black}{\textbf{C}} & \textcolor{green!45!black}{\textbf{C}} & \textcolor{yellow!45!black}{\textbf{E}} & \textcolor{green!45!black}{\textbf{C}} & \textcolor{green!45!black}{\textbf{C}} & \textcolor{green!45!black}{\textbf{C}} \\
AiF & ev-battery-rul & 0 & \textcolor{red!70!black}{\textbf{M}} & \textcolor{red!70!black}{\textbf{M}} & \textcolor{red!70!black}{\textbf{M}} & \textcolor{red!70!black}{\textbf{M}} & \textcolor{red!70!black}{\textbf{M}} & \textcolor{red!70!black}{\textbf{M}} \\
AiF & ica-transformer-soh & 7 & \textcolor{green!45!black}{\textbf{C}} & \textcolor{green!45!black}{\textbf{C}} & \textcolor{yellow!45!black}{\textbf{E}} & \textcolor{green!45!black}{\textbf{C}} & \textcolor{green!45!black}{\textbf{C}} & \textcolor{yellow!45!black}{\textbf{E}} \\
AiF & lstm-ae-health-rul & 0 & \textcolor{red!70!black}{\textbf{M}} & \textcolor{red!70!black}{\textbf{M}} & \textcolor{red!70!black}{\textbf{M}} & \textcolor{red!70!black}{\textbf{M}} & \textcolor{red!70!black}{\textbf{M}} & \textcolor{red!70!black}{\textbf{M}} \\
AiF & residual-fault-detection & 0 & \textcolor{red!70!black}{\textbf{M}} & \textcolor{red!70!black}{\textbf{M}} & \textcolor{red!70!black}{\textbf{M}} & \textcolor{red!70!black}{\textbf{M}} & \textcolor{red!70!black}{\textbf{M}} & \textcolor{red!70!black}{\textbf{M}} \\
AiF & turbofan-fault-prognosis-rul & 6 & \textcolor{green!45!black}{\textbf{C}} & \textcolor{green!45!black}{\textbf{C}} & \textcolor{green!45!black}{\textbf{C}} & \textcolor{green!45!black}{\textbf{C}} & \textcolor{green!45!black}{\textbf{C}} & \textcolor{green!45!black}{\textbf{C}} \\
AiF & turbofan-prognostics-diagnostics & 0 & \textcolor{red!70!black}{\textbf{M}} & \textcolor{red!70!black}{\textbf{M}} & \textcolor{red!70!black}{\textbf{M}} & \textcolor{red!70!black}{\textbf{M}} & \textcolor{red!70!black}{\textbf{M}} & \textcolor{red!70!black}{\textbf{M}} \\
AiF & vmd-ssa-patchtst-battery-rul & 0 & \textcolor{red!70!black}{\textbf{M}} & \textcolor{red!70!black}{\textbf{M}} & \textcolor{red!70!black}{\textbf{M}} & \textcolor{red!70!black}{\textbf{M}} & \textcolor{red!70!black}{\textbf{M}} & \textcolor{red!70!black}{\textbf{M}} \\
FCA & battery-soh-pipeline & 5 & \textcolor{green!45!black}{\textbf{C}} & \textcolor{green!45!black}{\textbf{C}} & \textcolor{green!45!black}{\textbf{C}} & \textcolor{red!70!black}{\textbf{M}} & \textcolor{green!45!black}{\textbf{C}} & \textcolor{green!45!black}{\textbf{C}} \\
FCA & battery-usage-vit & 9 & \textcolor{green!45!black}{\textbf{C}} & \textcolor{green!45!black}{\textbf{C}} & \textcolor{yellow!45!black}{\textbf{E}} & \textcolor{green!45!black}{\textbf{C}} & \textcolor{green!45!black}{\textbf{C}} & \textcolor{green!45!black}{\textbf{C}} \\
FCA & battnn-pinn-eod & 8 & \textcolor{green!45!black}{\textbf{C}} & \textcolor{green!45!black}{\textbf{C}} & \textcolor{green!45!black}{\textbf{C}} & \textcolor{green!45!black}{\textbf{C}} & \textcolor{green!45!black}{\textbf{C}} & \textcolor{green!45!black}{\textbf{C}} \\
FCA & bayesian-gated-transformer-rul & 3 & \textcolor{green!45!black}{\textbf{C}} & \textcolor{green!45!black}{\textbf{C}} & \textcolor{yellow!45!black}{\textbf{E}} & \textcolor{red!70!black}{\textbf{M}} & \textcolor{yellow!45!black}{\textbf{E}} & \textcolor{green!45!black}{\textbf{C}} \\
FCA & cart-net-aeroengine-rul & 8 & \textcolor{green!45!black}{\textbf{C}} & \textcolor{green!45!black}{\textbf{C}} & \textcolor{yellow!45!black}{\textbf{E}} & \textcolor{green!45!black}{\textbf{C}} & \textcolor{yellow!45!black}{\textbf{E}} & \textcolor{green!45!black}{\textbf{C}} \\
FCA & cata-tcn-engine-rul & 2 & \textcolor{red!70!black}{\textbf{M}} & \textcolor{red!70!black}{\textbf{M}} & \textcolor{red!70!black}{\textbf{M}} & \textcolor{red!70!black}{\textbf{M}} & \textcolor{red!70!black}{\textbf{M}} & \textcolor{red!70!black}{\textbf{M}} \\
FCA & causal-fault-diagnosis-aeroengine & 7 & \textcolor{green!45!black}{\textbf{C}} & \textcolor{green!45!black}{\textbf{C}} & \textcolor{yellow!45!black}{\textbf{E}} & \textcolor{red!70!black}{\textbf{M}} & \textcolor{green!45!black}{\textbf{C}} & \textcolor{green!45!black}{\textbf{C}} \\
FCA & degradai-battery-diagnosis & 4 & \textcolor{green!45!black}{\textbf{C}} & \textcolor{yellow!45!black}{\textbf{E}} & \textcolor{yellow!45!black}{\textbf{E}} & \textcolor{green!45!black}{\textbf{C}} & \textcolor{green!45!black}{\textbf{C}} & \textcolor{green!45!black}{\textbf{C}} \\
FCA & dual-informer-aeroengine-rul & 5 & \textcolor{green!45!black}{\textbf{C}} & \textcolor{green!45!black}{\textbf{C}} & \textcolor{yellow!45!black}{\textbf{E}} & \textcolor{green!45!black}{\textbf{C}} & \textcolor{green!45!black}{\textbf{C}} & \textcolor{green!45!black}{\textbf{C}} \\
FCA & ev-battery-rul & 8 & \textcolor{green!45!black}{\textbf{C}} & \textcolor{yellow!45!black}{\textbf{E}} & \textcolor{yellow!45!black}{\textbf{E}} & \textcolor{red!70!black}{\textbf{M}} & \textcolor{green!45!black}{\textbf{C}} & \textcolor{green!45!black}{\textbf{C}} \\
FCA & ica-transformer-soh & 5 & \textcolor{green!45!black}{\textbf{C}} & \textcolor{green!45!black}{\textbf{C}} & \textcolor{yellow!45!black}{\textbf{E}} & \textcolor{green!45!black}{\textbf{C}} & \textcolor{yellow!45!black}{\textbf{E}} & \textcolor{green!45!black}{\textbf{C}} \\
FCA & lstm-ae-health-rul & 12 & \textcolor{green!45!black}{\textbf{C}} & \textcolor{green!45!black}{\textbf{C}} & \textcolor{green!45!black}{\textbf{C}} & \textcolor{green!45!black}{\textbf{C}} & \textcolor{green!45!black}{\textbf{C}} & \textcolor{green!45!black}{\textbf{C}} \\
FCA & residual-fault-detection & 8 & \textcolor{green!45!black}{\textbf{C}} & \textcolor{green!45!black}{\textbf{C}} & \textcolor{green!45!black}{\textbf{C}} & \textcolor{red!70!black}{\textbf{M}} & \textcolor{green!45!black}{\textbf{C}} & \textcolor{green!45!black}{\textbf{C}} \\
FCA & turbofan-fault-prognosis-rul & 9 & \textcolor{green!45!black}{\textbf{C}} & \textcolor{green!45!black}{\textbf{C}} & \textcolor{green!45!black}{\textbf{C}} & \textcolor{green!45!black}{\textbf{C}} & \textcolor{green!45!black}{\textbf{C}} & \textcolor{green!45!black}{\textbf{C}} \\
FCA & turbofan-prognostics-diagnostics & 0 & \textcolor{red!70!black}{\textbf{M}} & \textcolor{red!70!black}{\textbf{M}} & \textcolor{red!70!black}{\textbf{M}} & \textcolor{red!70!black}{\textbf{M}} & \textcolor{red!70!black}{\textbf{M}} & \textcolor{red!70!black}{\textbf{M}} \\
FCA & vmd-ssa-patchtst-battery-rul & 3 & \textcolor{green!45!black}{\textbf{C}} & \textcolor{green!45!black}{\textbf{C}} & \textcolor{yellow!45!black}{\textbf{E}} & \textcolor{green!45!black}{\textbf{C}} & \textcolor{yellow!45!black}{\textbf{E}} & \textcolor{yellow!45!black}{\textbf{E}} \\
AiF-D & battery-soh-pipeline & 16 & \textcolor{green!45!black}{\textbf{C}} & \textcolor{green!45!black}{\textbf{C}} & \textcolor{green!45!black}{\textbf{C}} & \textcolor{green!45!black}{\textbf{C}} & \textcolor{green!45!black}{\textbf{C}} & \textcolor{green!45!black}{\textbf{C}} \\
AiF-D & battery-usage-vit & 7 & \textcolor{green!45!black}{\textbf{C}} & \textcolor{green!45!black}{\textbf{C}} & \textcolor{yellow!45!black}{\textbf{E}} & \textcolor{green!45!black}{\textbf{C}} & \textcolor{green!45!black}{\textbf{C}} & \textcolor{green!45!black}{\textbf{C}} \\
AiF-D & battnn-pinn-eod & 0 & \textcolor{red!70!black}{\textbf{M}} & \textcolor{red!70!black}{\textbf{M}} & \textcolor{red!70!black}{\textbf{M}} & \textcolor{red!70!black}{\textbf{M}} & \textcolor{red!70!black}{\textbf{M}} & \textcolor{red!70!black}{\textbf{M}} \\
AiF-D & bayesian-gated-transformer-rul & 5 & \textcolor{green!45!black}{\textbf{C}} & \textcolor{yellow!45!black}{\textbf{E}} & \textcolor{green!45!black}{\textbf{C}} & \textcolor{green!45!black}{\textbf{C}} & \textcolor{yellow!45!black}{\textbf{E}} & \textcolor{green!45!black}{\textbf{C}} \\
AiF-D & cart-net-aeroengine-rul & 11 & \textcolor{green!45!black}{\textbf{C}} & \textcolor{green!45!black}{\textbf{C}} & \textcolor{yellow!45!black}{\textbf{E}} & \textcolor{green!45!black}{\textbf{C}} & \textcolor{green!45!black}{\textbf{C}} & \textcolor{green!45!black}{\textbf{C}} \\
AiF-D & cata-tcn-engine-rul & 6 & \textcolor{green!45!black}{\textbf{C}} & \textcolor{yellow!45!black}{\textbf{E}} & \textcolor{yellow!45!black}{\textbf{E}} & \textcolor{green!45!black}{\textbf{C}} & \textcolor{yellow!45!black}{\textbf{E}} & \textcolor{green!45!black}{\textbf{C}} \\
AiF-D & causal-fault-diagnosis-aeroengine & 0 & \textcolor{red!70!black}{\textbf{M}} & \textcolor{red!70!black}{\textbf{M}} & \textcolor{red!70!black}{\textbf{M}} & \textcolor{red!70!black}{\textbf{M}} & \textcolor{red!70!black}{\textbf{M}} & \textcolor{red!70!black}{\textbf{M}} \\
AiF-D & degradai-battery-diagnosis & 6 & \textcolor{green!45!black}{\textbf{C}} & \textcolor{green!45!black}{\textbf{C}} & \textcolor{green!45!black}{\textbf{C}} & \textcolor{red!70!black}{\textbf{M}} & \textcolor{green!45!black}{\textbf{C}} & \textcolor{green!45!black}{\textbf{C}} \\
AiF-D & dual-informer-aeroengine-rul & 10 & \textcolor{green!45!black}{\textbf{C}} & \textcolor{green!45!black}{\textbf{C}} & \textcolor{yellow!45!black}{\textbf{E}} & \textcolor{red!70!black}{\textbf{M}} & \textcolor{green!45!black}{\textbf{C}} & \textcolor{green!45!black}{\textbf{C}} \\
AiF-D & ev-battery-rul & 9 & \textcolor{green!45!black}{\textbf{C}} & \textcolor{yellow!45!black}{\textbf{E}} & \textcolor{yellow!45!black}{\textbf{E}} & \textcolor{green!45!black}{\textbf{C}} & \textcolor{yellow!45!black}{\textbf{E}} & \textcolor{green!45!black}{\textbf{C}} \\
AiF-D & ica-transformer-soh & 6 & \textcolor{green!45!black}{\textbf{C}} & \textcolor{green!45!black}{\textbf{C}} & \textcolor{green!45!black}{\textbf{C}} & \textcolor{yellow!45!black}{\textbf{E}} & \textcolor{green!45!black}{\textbf{C}} & \textcolor{yellow!45!black}{\textbf{E}} \\
AiF-D & lstm-ae-health-rul & 0 & \textcolor{red!70!black}{\textbf{M}} & \textcolor{red!70!black}{\textbf{M}} & \textcolor{red!70!black}{\textbf{M}} & \textcolor{red!70!black}{\textbf{M}} & \textcolor{red!70!black}{\textbf{M}} & \textcolor{red!70!black}{\textbf{M}} \\
AiF-D & residual-fault-detection & 0 & \textcolor{red!70!black}{\textbf{M}} & \textcolor{red!70!black}{\textbf{M}} & \textcolor{red!70!black}{\textbf{M}} & \textcolor{red!70!black}{\textbf{M}} & \textcolor{red!70!black}{\textbf{M}} & \textcolor{red!70!black}{\textbf{M}} \\
AiF-D & turbofan-fault-prognosis-rul & 6 & \textcolor{green!45!black}{\textbf{C}} & \textcolor{green!45!black}{\textbf{C}} & \textcolor{green!45!black}{\textbf{C}} & \textcolor{red!70!black}{\textbf{M}} & \textcolor{green!45!black}{\textbf{C}} & \textcolor{yellow!45!black}{\textbf{E}} \\
AiF-D & turbofan-prognostics-diagnostics & 11 & \textcolor{green!45!black}{\textbf{C}} & \textcolor{green!45!black}{\textbf{C}} & \textcolor{yellow!45!black}{\textbf{E}} & \textcolor{green!45!black}{\textbf{C}} & \textcolor{green!45!black}{\textbf{C}} & \textcolor{green!45!black}{\textbf{C}} \\
AiF-D & vmd-ssa-patchtst-battery-rul & 5 & \textcolor{green!45!black}{\textbf{C}} & \textcolor{green!45!black}{\textbf{C}} & \textcolor{yellow!45!black}{\textbf{E}} & \textcolor{yellow!45!black}{\textbf{E}} & \textcolor{yellow!45!black}{\textbf{E}} & \textcolor{yellow!45!black}{\textbf{E}} \\
\bottomrule
\end{tabular}
\end{adjustbox}
\end{table}

\begin{table}[t]
  \centering
  \small
  \setlength{\tabcolsep}{4pt}
  \renewcommand{\arraystretch}{1.1}
  \caption{Intermediate artifacts emitted by the framework-coupled workflow.}
  \label{tab:workflow-artifacts}
  \begin{tabular}{p{1.8cm}p{5.7cm}p{5.2cm}}
    \toprule
    Stage & Artifact & Main role in the pipeline \\
    \midrule
    \textsc{Ingest} & Paper hub and chunk index & Extract paper knowledge from the PDF and preserve metadata, section structure, and source references for downstream conceptual and algorithmic agents \\
    \textsc{Analyze} & Assumption records $\mathcal{A}$ & Extract claims, datasets, method details. Mark unresolved implementation decisions and attach evidence-backed defaults \\
    \textsc{Map} & Slot-level framework binding plan & Resolve paper concepts into concrete task, datasource, transform, sequencer, model, and evaluator bindings \\
    \textsc{Implement} & Configuration $c$ and extensions $\mathcal{C}$ & Materialize the executable framework artifact \\
    \textsc{Verify} & Verification-and-evaluation report $r$ & Check type, shape, leakage, execution, and result matching under shared conditions \\
    \textsc{Report} & Benchmark-ready artifact or failure report & Return an auditable artifact for benchmarking or diagnosis \\
    \bottomrule
  \end{tabular}
\end{table}

\newpage

\section{Agentic methods}

\subsection{DeepCode}
\label{app:paper2code}

The Paper2Code-style baseline reported in Section~\ref{sec:experiments} is instantiated through DeepCode~\citep{li2025deepcode}, an open-source agentic coding framework that builds on the staged \textsc{Plan}--\textsc{Analyze}--\textsc{Generate} pipeline of Paper2Code~\citep{seo2025paper2code}. We adopt it without modification as the strongest publicly available realization of this paradigm at the time of writing.

DeepCode recasts paper-to-repository synthesis as an information-flow problem: a paper carries more task-relevant content than fits in a single LLM context, so the system compresses, indexes, retrieves, and verifies information across three coupled phases. (i)~A planning phase parses the paper into a hierarchical content index, runs a Concept Agent and an Algorithm Agent over it to extract conceptual structure and low-level technical content, and merges them into an implementation blueprint that fixes the file hierarchy, module specifications, dependencies, and staged development order. (ii)~A generation phase emits the repository file by file against the blueprint, using a structured cross-file memory (\emph{CodeMem}) and a retrieval-augmented component over indexed external repositories (\emph{CodeRAG}) to keep generated files mutually consistent and grounded in concrete implementation idioms. (iii)~A verification phase runs static analysis and a sandbox executor; failures are routed back to a modification agent that applies targeted edits until the repository runs or a refinement budget is exhausted. We refer to~\citet{li2025deepcode,seo2025paper2code} for full details.

For each paper in the corpus, the source PDF is submitted to DeepCode (\url{https://github.com/HKUDS/DeepCode}) and run end-to-end with default settings; the resulting repository is the artifact under evaluation.

\subsection{Agentic workflow}
\label{app:agentic-workflow}

This section expands on the workflow described in Section~\ref{subsec:workflow}. We detail the runtime structure of the agent system, the stateful coordination layer that gives the loop its operational determinism, the preprocessing path from raw PDF to indexed input, and the concrete mapping of the abstract \textsc{Ingest}--\textsc{Analyze}--\textsc{Map}--\textsc{Implement}--\textsc{Verify}--\textsc{Report} stages onto the agents and skills that execute them. The goal is to make the moving parts auditable: which agent owns which decision, what each skill is allowed to do, and how the framework $\mathcal{F}$ is wired in as the source of truth for component reuse.

\paragraph{Runtime structure.} The workflow is exposed to the user as a single command, \texttt{/validate-paper}, which takes the paper directory together with a run mode (\textsc{blueprint-only}, \textsc{quick}, or \textsc{full}) and a hyperparameter mode that selects between the framework reference profile and a paper-faithful auxiliary profile. Internally, the system is organized as a primary orchestrator and five specialized subagents that communicate exclusively through a shared paper \emph{vault} of numbered artifacts (Table~\ref{tab:agentic-phm-agents}). No agent talks to another agent through free-form chat; every cross-agent dependency is materialized as a JSON sidecar on disk, and every action that mutates state, including phase transitions, is mediated by a tool call. This separation between conversational reasoning and persisted state is what allows the run to be paused, resumed, audited, or replayed without losing scientific provenance.

\paragraph{Stateful coordination.} The orchestration layer is built around a deterministic control plane: a single \texttt{run\_state.json} file maintained by the \texttt{validate\_paper\_workflow\_*} tools, which is the only authoritative source of workflow status. Each phase is opened with \texttt{start\_phase} and closed with \texttt{complete\_phase} or \texttt{fail\_phase}; on resume or after an unexpected subagent abort, \texttt{sync\_from\_artifacts} reconciles the control plane with whatever artifacts already exist on disk, and \texttt{get\_status} returns the current state machine view before the orchestrator decides what to do next. Three workflow-status axes are kept strictly separate inside this state machine: an \emph{artifact} axis records whether each phase produced its required machine-readable and human-readable outputs; a \emph{technical} axis records whether the implementation is runnable; and a \emph{scientific} axis records the verdict of the paper comparison. Only the first two can block progress, which prevents an unfavorable scientific outcome from being silently retried as if it were a bug. The agent prose under \texttt{.opencode/agents/} never duplicates these rules: orchestration policies, retry budgets, validation modes, and abort-recovery behavior all live in a single \texttt{policies.md} file that every agent references but no agent restates, so the system has exactly one source of truth for control-flow decisions and one for scientific content.

\paragraph{Phase mapping.} The five workflow phases instantiate the abstract loop of Section~\ref{subsec:workflow}, summarized in Algorithm~\ref{alg:agentic-phm-app}. (P1)~\textsc{Ingest} runs the \texttt{/process-paper} skill, which uses \emph{marker-pdf} \citep{marker} to convert the raw PDF into Markdown with preserved math and a deterministic section index, and then invokes \textsc{chunk-indexer} to produce \texttt{00-paper-hub} and \texttt{01-chunk-index}. (P2)~\textsc{Analyze} runs \textsc{conceptual-analysis} and \textsc{algorithmic-spec} in parallel, emitting \texttt{02} and \texttt{03} with the assumption set $\mathcal{A}$ partially materialized as the nine-row hyperparameter contract and the dataset mapping. (P3)~\textsc{Map}+\textsc{Implement} are fused inside \textsc{implementation-blueprint}, which writes the self-contained build plan \texttt{04} and the validation run matrix. (P3.5)~the \texttt{/paper-hypothesis} skill pre-registers the per-target claim set into \texttt{05}. (P4)~\textsc{Verify}+\textsc{Report} are driven by \textsc{experimenter}, which dispatches the implementation skills for novel slots only, runs the static and sanity verification ladder, executes training, and produces \texttt{08-evaluation-report}. Each phase emits both a machine-readable JSON sidecar and a rendered Markdown audit file in numbered order from \texttt{00-paper-hub} through \texttt{08-evaluation-report}; the JSON files are the contract consumed by downstream agents, the Markdown files are the human audit view.

\begin{algorithm}[t]
\caption{Agentic PHM paper validation against a shared PHM framework.}
\label{alg:agentic-phm-app}
\begin{algorithmic}[1]
\Require Paper $P$; shared framework $\mathcal{F}$; result-matching tolerance $\tau$; patience $N$
\Ensure Implementation artifact $A = (c, \mathcal{C}, \mathcal{A}, r)$
\State $\text{spec} \gets \textsc{Ingest}(P)$ \Comment{indexed paper representation}
\State $(\text{spec}, \mathcal{A}) \gets \textsc{Analyze}(\text{spec}, \mathcal{F})$ \Comment{Paper content analysis; emit assumption records}
\State $(c, \mathcal{C}) \gets \textsc{Implement}\bigl(\textsc{Map}(\text{spec}, \mathcal{F})\bigr)$ \Comment{bind typed slots and extension stubs}
\State $n \gets 0$
\Repeat
  \State $r \gets \textsc{Verify}(c, \mathcal{C}, \mathcal{F}, \tau)$ \Comment{typecheck $\to$ dry-run $\to$ leakage audit $\to$ run-and-match}
  \If{$r.\text{status} = \textsf{failed}$ \textbf{and} $\exists\, a_i \in \mathcal{A}:\; s_i \in \textsc{Slots}(r.\text{cause})$}
    \State revise $a_i$ using $\mathcal{F}$'s alternative defaults; patch $(c, \mathcal{C})$; $n \gets n + 1$
  \Else
    \State \textbf{break}
  \EndIf
\Until{$r.\text{status} = \textsf{passed}$ \textbf{or} $n \geq N$}
\State \Return $A = (c, \mathcal{C}, \mathcal{A}, r)$
\end{algorithmic}
\end{algorithm}

\paragraph{Agents.} The workflow uses six agents, each constrained to a narrow role and a restricted tool surface (Table~\ref{tab:agentic-phm-agents}). The primary orchestrator (\textsc{paper-validator}) does no implementation itself: it resolves inputs, opens phases through the control plane, dispatches subagents and skills in dependency order, and applies the bounded-retry and abort-recovery rules defined once in \texttt{policies.md}. Two paper-understanding subagents (\textsc{conceptual-analysis} and \textsc{algorithmic-spec}) run in parallel after the chunk index is built: the first identifies what is novel versus what $\mathcal{F}$ already provides and produces the dataset mapping, while the second performs an exhaustive extraction of equations, architectures, losses, and the nine-row training-hyperparameter contract. Their outputs are merged by \textsc{implementation-blueprint}, which writes the self-contained build plan that downstream skills consume. Finally, \textsc{experimenter} drives Phase~4 as a project manager: it reads the blueprint, dispatches implementation skills for novel components only, runs the verification ladder, schedules training rows from the validation matrix, and invokes evaluation against the pre-registered claims.

\begin{table}[!ht]
\centering
\scriptsize
\caption{Agents in the framework-coupled workflow. Tool surface is restricted at the harness level; each agent only sees the subagents and bash commands listed in its frontmatter.}
\label{tab:agentic-phm-agents}
\begin{tabular}{p{0.18\textwidth} p{0.10\textwidth} p{0.60\textwidth}}
\toprule
\textbf{Agent} & \textbf{Mode} & \textbf{Role} \\
\midrule
\textsc{paper-validator} & primary & Orchestrates phases through \texttt{run\_state.json}; dispatches subagents and skills; applies bounded retries and abort recovery. \\
\textsc{chunk-indexer} & subagent & Builds \texttt{00-paper-hub} and \texttt{01-chunk-index} from the marker output and section index. \\
\textsc{conceptual-analysis} & subagent & Maps the paper to existing components in $\mathcal{F}$, flags novel slots, and selects direct evaluation targets. \\
\textsc{algorithmic-spec} & subagent & Extracts every equation, architecture, loss, and the required training hyperparameter table. \\
\textsc{implementation-blueprint} & subagent & Merges \texttt{02} and \texttt{03} into a self-contained, skill-oriented build plan with executable contracts. \\
\textsc{experimenter} & subagent & Drives implementation, static and sanity verification, batch-fit checks, training, and evaluation per scheduled row. \\
\bottomrule
\end{tabular}
\end{table}
\FloatBarrier

\paragraph{Skills.} Skills are stateless action procedures invoked as slash commands; they do not carry conversational state across invocations and are the only place where files under \texttt{picid/} or \texttt{configs/} are written. We group them into four families. \emph{Implementation skills} (\texttt{/implement-datasource}, \texttt{/implement-transform}, \texttt{/implement-model}, \texttt{/implement-loss}, \texttt{/implement-experiment}) each create one slot's worth of code and config, following the contracts in \texttt{.opencode/reference/} so that the produced files plug back into $\mathcal{F}$ without bespoke glue. \emph{Verification skills} cover the static and sanity layers of Section~\ref{subsec:assumptions}. \texttt{/verify-static} performs file existence, import resolution, base-class, and config validity checks plus a datasource preflight. \texttt{/verify-sanity} adapts a small subset of the model-correctness probes recommended by Karpathy's training-neural-networks recipe~\citep{karpathy2019recipe}: an \emph{initial-loss} check that compares the loss at step zero against the task prior (e.g., $-\log(1/C)$ for $C$-way classification, target variance for regression); a \emph{gradient-flow} check that confirms every trainable parameter receives a finite, non-zero gradient on a single forward-backward pass and rules out dead branches or batch-dimension leakage from incorrect reshapes; and a \emph{micro-batch memorization} check that requires the model to overfit a handful of examples to near-zero loss within a few hundred steps. We deliberately drop the heavier checks in the original recipe (e.g., long input-independent baselines, full subset-convergence sweeps) and run the remaining three in a single in-process build of the framework stack so the gate stays cheap and tightly scoped to wiring bugs that static checks structurally cannot see. \texttt{/check-batch-fit} then confirms that the fixed validation batch sizes (train~$512$, val/test~$1024$) fit on the selected hardware before any full training command is issued. \emph{Diagnostic loops} (\texttt{/diagnose-verify-block}, \texttt{/diagnose-training-result}) implement bounded global-hypothesis repair: each iteration re-runs the full check suite, forms one hypothesis that explains all failing checks jointly, and applies a single change inside a strict per-run writable-file whitelist derived from the blueprint and the run's git diff, so framework code is structurally protected. \emph{Execution and reporting skills} (\texttt{/paper-hypothesis}, \texttt{/run-training}, \texttt{/evaluate-results}) pre-register claims, drive the actual training subprocess with monitored health probes, and produce the final evaluation report that judges each pre-registered claim against achieved metrics and framework baselines. Figures~\ref{fig:prompt-implement-model}--\ref{fig:prompt-evaluate-results} in Appendix~\ref{app:prompts} reproduce one abridged skill prompt per family.

\paragraph{Tools.} Deterministic Python tools sit underneath the skills and are the only way to mutate workflow state or perform expensive work. The control plane (\texttt{validate\_paper\_workflow\_init\_run}, \texttt{start\_phase}, \texttt{complete\_phase}, \texttt{fail\_phase}, \texttt{sync\_from\_artifacts}, \texttt{get\_status}, plus the sidecar writers for files \texttt{02}, \texttt{03}, and \texttt{04}) validates each JSON sidecar against a schema before rendering its Markdown twin, which means a malformed paper-understanding artifact cannot leak into Phase~3. Three execution tools wrap framework commands as long-lived subprocesses: \texttt{tabphm\_sanity\_ladder} for the trimmed Karpathy ladder, \texttt{tabphm\_baselines} for retrieving frozen TabPHM baseline metrics from \texttt{report\_output/}, and \texttt{tabphm\_training\_run} for full training runs with monitored heartbeats and structured exit classification. A separate paper-side tool, \texttt{index\_paper.py}, builds the section index consumed by the paper-understanding agents from the marker-converted markdown.

\paragraph{Artifact layering.} Every phase writes to two artifact tiers. The control-plane tier (\texttt{run\_state.json}) carries phase status, gate decisions, and recovery metadata and is the only authoritative source of workflow control flow. The audit tier (numbered files \texttt{00}--\texttt{08} plus \texttt{session\_stats.json}) carries the scientific content: the paper hub, chunk index, conceptual and algorithmic analyses, the implementation blueprint, the pre-registered hypothesis, the sanity ladder log, the training log, and the evaluation report. Within the audit tier, JSON sidecars are the machine contract for downstream agents and Markdown files are the human-readable mirror; both are produced by the same writer tool to guarantee they cannot diverge. The assumption set $\mathcal{A}$ of Section~\ref{subsec:assumptions} is materialized across \texttt{02}, \texttt{03}, and \texttt{04}: each unspecified decision becomes a row whose slot identifier matches a binding in the blueprint, which is what makes the attribution predicate of Section~\ref{subsec:assumptions} computable at verification time.

\paragraph{Handling underspecification.} Values the paper omits (optimizer, learning rate, schedule, weight decay, grad clip, warmup, max epochs, batch size, protocol details) are routed through the assumption set $\mathcal{A}$ rather than filled silently. \textsc{algorithmic-spec} writes a fixed nine-row hyperparameter table in which omitted fields appear as \texttt{NOT\_SPECIFIED}; \textsc{implementation-blueprint} binds each row to a typed slot in the experiment YAML and emits a framework-default imputation as a \texttt{\# SUBSTITUTION:} comment with its source. Validation batch sizes (\texttt{512}/\texttt{1024}/\texttt{1024}) and the framework scheduler/check cadence are fixed regardless of paper-stated values, which are kept as provenance, to preserve leaderboard comparability. \texttt{/evaluate-results} reports a recap partitioning each load-bearing field into paper-stated, \texttt{NOT\_SPECIFIED}, and imputed/substituted rows, and the bounded \texttt{/diagnose-training-result} loop may revise an $a_i$ when a failure traces to its slot (line~36 of Algorithm~\ref{alg:agentic-phm-app}). Figure~\ref{fig:example-imputed-yaml} shows a representative YAML emitted under this policy: every field the paper left as \texttt{NOT\_SPECIFIED} is annotated inline with the framework default substituted in and the source it came from.

\begin{figure}[ht!]
\centering
\scriptsize
\begin{tcolorbox}[
  title=Example framework experiment YAML produced from an underspecified paper,
  fonttitle=\bfseries,
  rounded corners,
  width=\textwidth
]
\ttfamily\scriptsize
defaults: \\
~~- nb14/prognostics/base \\
~~- /model\_configs/prognostics/degradai \\
~~- override /task\_definition: prognostics/degradai\_rul\_500 \\
~~- override /transforms: battery/nb14/raw \\
~~- override /loss: degradai\_composite \\
~~- override /paths: agent \\
~\\

\textcolor{blue}{\# SUBSTITUTION: optimizer=NOT\_SPECIFIED -> AdamW from configs} \\
\textcolor{blue}{\# SUBSTITUTION: lr\_schedule=NOT\_SPECIFIED -> preserve ReduceLROnPlateau cadence from configs} \\
\textcolor{blue}{\# SUBSTITUTION: warmup=NOT\_SPECIFIED -> no warmup override available in configs} \\
\textcolor{blue}{\# SUBSTITUTION: training\_protocol\_notes are provenance only; composite-loss weights remain from configs}\\
~\\
optimization: \\
~~lr: \textcolor{blue}{1.0e-3} \ \ \textcolor{blue}{\# SUBSTITUTION: learning\_rate=NOT\_SPECIFIED -> 1.0e-3 from configs} \\
~~optimizer: \\
~~~~\_target\_: torch.optim.\textcolor{blue}{AdamW} \\
~~~~lr: \$\{optimization.lr\} \\
~~~~weight\_decay: \textcolor{blue}{0.0} \ \ \textcolor{blue}{\# SUBSTITUTION: weight\_decay=NOT\_SPECIFIED -> 0.0 from configs} \\
~~scheduler: \\
~~~~\_target\_: torch.optim.lr\_scheduler.\textcolor{blue}{ReduceLROnPlateau} \\
~~~~patience: 5 \\
~~~~factor: 0.9 \\
~\\
trainer: \\
~~max\_epochs: \textcolor{blue}{300} \ \ \textcolor{blue}{\# SUBSTITUTION: max\_epochs=NOT\_SPECIFIED -> 300 from configs/trainer/default.yaml} \\
~~gradient\_clip\_val: \textcolor{blue}{null} \ \ \textcolor{blue}{\# SUBSTITUTION: grad\_clip=NOT\_SPECIFIED -> null because configs does not set gradient clipping} \\
~\\
datamodule: \\
~~train\_batch\_size: 512 \\
~~val\_batch\_size: 1024 \\
~~test\_batch\_size: 1024 \\
~\\
paths: \\
~~output\_dir: \$\{paths.artifacts\_dir\}/validate-paper/nb14/prognostics/raw/degradai \\
~~work\_dir: \$\{paths.root\_dir\} \\
\end{tcolorbox}
\caption{Representative experiment YAML emitted by \textsc{implementation-blueprint} for a paper whose hyperparameter table is largely \texttt{NOT\_SPECIFIED}. Each missing field is replaced by an explicit framework default and recorded with a \texttt{\# SUBSTITUTION:} comment naming the source config; validation batch sizes and the scheduler/check cadence are fixed by the workflow contract regardless of paper values.}
\label{fig:example-imputed-yaml}
\end{figure}

\paragraph{Prompt snippets.} Appendix~\ref{app:prompts} reproduces the abridged system prompts that drive the orchestrator and the two paper-understanding subagents. The prompts establish the role, the input and output contracts, and the rules each agent must follow; rendered Markdown shapes and full schemas are omitted there and live in the agent definition files under \texttt{.opencode/agents/}.

\subsection{Prompt-based in-framework generation}
\label{app:prompt-in-framework}

The Agent-in-Framework baseline of Section~\ref{sec:experiments} isolates the contribution of staged agent scaffolding from the contribution of framework grounding itself. The framework $\mathcal{F}$ and its evaluator are held fixed, but the orchestrator, subagents, skills, control plane, sidecar contracts, and assumption-tracking machinery introduced in Appendix~\ref{app:agentic-workflow} are removed. What remains is a single coding agent placed inside the framework repository and asked to implement the paper end-to-end from one prompt.

\paragraph{Harness and scope.} The baseline uses the same opencode coding agent and the same backend model as the full agentic workflow, configured to run from the root of the framework checkout used by our system. The agent has full read and write access to the repository: it can inspect the framework reference files under \texttt{.opencode/reference/}, the existing datasource, transform, model, loss, and experiment slots under \texttt{picid/}, and the Hydra configs under \texttt{configs/}, and it can edit or create files anywhere it judges necessary. None of the slash commands defined in Appendix~\ref{app:agentic-workflow} (\texttt{/process-paper}, the \texttt{/implement-*} family, \texttt{/verify-static}, \texttt{/verify-sanity}, \texttt{/check-batch-fit}, \texttt{/run-training}, \texttt{/evaluate-results}, the diagnostic loops, or the \texttt{validate\_paper\_workflow\_*} control plane) are exposed; the tool surface reduces to the harness's default shell, file editing, and search primitives.

\paragraph{Input.} The paper is provided as the original PDF, without the marker-pdf conversion, the section index, the chunk index, or any of the structured paper-understanding artifacts (\texttt{02-conceptual-analysis}, \texttt{03-algorithmic-spec}, the implementation blueprint) that the agentic workflow produces in Phases~1--2. Whatever paper understanding occurs happens implicitly inside the agent's context window. No assumption records $a_i \in \mathcal{A}$ are emitted, no nine-row hyperparameter contract is enforced, no novelty classification against \texttt{.opencode/reference/inventory.md} is required, and no pre-registered claim set is written before training.

\paragraph{Verification and evaluation.} The static, sanity, and batch-fit gates of the agentic workflow are not invoked, and there is no enforced verification ladder, no global-hypothesis diagnostic loop, and no bounded-retry policy. The agent is free to run any framework command it judges relevant, including the training entry points. For evaluation we use the same \texttt{paths=agent} convention as the agentic workflow so that outputs land in the per-run path expected by $\mathcal{F}$'s evaluator; windowing, splits, metric definitions, and the frozen baseline leaderboard are therefore identical across systems, which is what makes the LLM-as-judge ratings and binding-state counts of Section~\ref{sec:experiments} directly comparable to the agentic mode.

\paragraph{Prompt.} The agent receives the single instruction reproduced in Figure~\ref{fig:prompt-in-framework-baseline} in Appendix~\ref{app:prompts}. The placeholder \texttt{path\_to\_paper.pdf} is replaced per run with the absolute path to the paper PDF in the run's vault directory; everything else is held fixed across the 16-paper corpus.

\paragraph{What this baseline isolates.} Because $\mathcal{F}$, the backend model, the harness, and the evaluation protocol are all held fixed, the only variables removed relative to our full system are (i)~the staged workflow phases and their explicit inter-phase contracts, (ii)~the materialized assumption set $\mathcal{A}$ and the attribution predicate that depends on it, (iii)~the verification ladder and the bounded global-hypothesis diagnostic loops, and (iv)~the pre-registered claim set consumed by \texttt{/evaluate-results}. Comparisons against this baseline therefore measure the contribution of agent-side scaffolding under shared framework grounding, complementing the comparison against Paper2Code (Appendix~\ref{app:paper2code}) in which framework grounding itself is removed.

\subsection{Prompt-based generation}
\label{app:prompt-standalone}

The standalone prompt-only baseline reported in Section~\ref{sec:experiments} removes both the staged scaffolding of Appendix~\ref{app:agentic-workflow} and the framework grounding shared by the in-framework baseline of Appendix~\ref{app:prompt-in-framework}. The agent is dropped into an empty repository with no access to $\mathcal{F}$'s slot contracts, datasource adapters, training loop, evaluator, or frozen baseline leaderboard, and is asked to produce a self-contained implementation of the paper from scratch. This isolates the joint contribution of framework grounding and agent-side scaffolding by removing both at once, and matches the operating regime in which standalone paper-to-code systems are typically deployed.

\paragraph{Harness and scope.} The baseline uses the same opencode coding agent and the same backend model as the other two systems, but the working directory is a freshly initialized empty repository rather than a checkout of $\mathcal{F}$. The agent is sandboxed: it can read and write files inside that repository and read files under a fixed dataset root, but it cannot edit anything outside the repository, copy datasets into it, or pull pre-existing reference implementations from the internet. None of the slash commands, sidecar contracts, control-plane tools, or framework reference files described in Appendix~\ref{app:agentic-workflow} are available, and the agent is explicitly instructed not to retrieve external code.

\paragraph{Input.} As in the in-framework baseline, the paper is provided as the original PDF without marker-pdf conversion, section indexing, or any of the structured paper-understanding artifacts. The dataset root is exposed at a fixed path (\texttt{/workspace/datasets}); the agent is responsible for inspecting it, identifying which available dataset best matches the paper's experimental setup, and documenting the resulting assumption when an exact match is unavailable. Because the framework's curated dataset mapping is not exposed, dataset selection is itself part of the agent's task rather than a pre-computed input.

\paragraph{Verification and evaluation.} No verification ladder is enforced and no diagnostic loop is provided. The agent is expected to leave the repository in a runnable state, with its own training and evaluation entry points, and to report installation, run, and validation instructions in free text. For comparability with the framework-coupled systems, evaluation is carried out post hoc: the agent's standalone repository is ported into $\mathcal{F}$ using the same task contracts, splits, windowing, and metric definitions used elsewhere, so that the LLM-as-judge ratings, binding-state counts, and cost statistics of Section~\ref{sec:experiments} share an evaluator with the agentic and in-framework systems.

\paragraph{Prompt.} The agent receives the single instruction reproduced in Figure~\ref{fig:prompt-standalone-baseline} in Appendix~\ref{app:prompts}. The dataset-root path is held fixed across the corpus; the paper path resolves to the per-run PDF inside the repository's \texttt{paper/} directory.

\paragraph{What this baseline isolates.} Relative to our full system, this baseline removes (i)~the staged workflow phases and their inter-phase contracts, (ii)~the materialized assumption set $\mathcal{A}$ and the attribution predicate, (iii)~the verification ladder and the bounded diagnostic loops, (iv)~the framework slot contracts, datasource adapters, training loop, and evaluator that constitute $\mathcal{F}$, and (v)~the pre-registered claim set consumed by \texttt{/evaluate-results}. Together with the in-framework baseline of Appendix~\ref{app:prompt-in-framework}, this completes a controlled ablation of the two contributions of our system: comparing standalone against in-framework prompt-only isolates framework grounding under a fixed prompting regime, while comparing in-framework prompt-only against the agentic mode isolates agent-side scaffolding under a fixed framework. The Paper2Code baseline of Appendix~\ref{app:paper2code} sits alongside this row as the strongest publicly available standalone paper-to-code system at the time of writing.

\subsection{Evaluation protocol handoff}
\label{app:methods-evaluation-handoff}

The implementation-method details end above. The implementation prompts are reproduced in Appendix~\ref{app:prompts}. The LLM-as-judge ratings, category aliases, judge aggregation, execution-rate audit, benchmark execution protocol, and complexity accounting are specified in Appendix~\ref{app:evaluation-setup}. In particular, Appendix~\ref{app:judge-categories} defines the reference-free LLM-as-judge rubric, and Appendix~\ref{app:execution-rate-audit} defines the read-only agent audit used to compute execution rates.

\clearpage

\section{Evaluation setup}
\label{app:evaluation-setup}

This appendix specifies the run-level evaluation protocol used for the results in Section~\ref{sec:setup}. The unit of analysis is a paper--implementation method pair: one PHM paper is given to one implementation method, which returns either a framework-integrated artifact or a standalone implementation. For each run, we record the generated artifacts, execution status, resource usage, and output size.

The evaluation separates five aspects of implementation quality and usability. First, for methods that target the shared PHM framework, we measure framework integration as a binary completion rate: whether the method produces a valid binding $(c,\mathcal{C})$ inside the framework. Second, we record slot-level binding diagnostics to identify which framework interfaces were successfully created, reused, or left missing. Third, we measure execution as an operational property of the generated artifact: whether it can be launched and reaches the appropriate evaluator without runtime failure. Fourth, we assess code quality and implementation fidelity using independent LLM-as-judge ratings over the generated code. Finally, for framework-compatible artifacts, we evaluate benchmarkability under shared datasource, task, split, and evaluator definitions. Complexity statistics, including token usage, wall-clock duration, generated files, and generated lines of code, are logged for the same paper--implementation method pairs.

\subsection{Corpus and units of analysis}
\label{app:evaluation-units}

The evaluation uses the 16-paper PHM corpus reported in the main text. Each paper is independently implemented by each considered implementation method. For every run we record the paper identifier, implementation method, backend model identifier, reasoning setting, wall-clock duration, token usage, generated files, generated lines of code, final run status, and the resulting artifacts within the implementation directory.

A run is counted as \emph{complete} when it produces a directory that contains the files needed to inspect the implementation: source code, configuration or execution entry points, and run metadata. Completion does not require successful training or evaluation. A completed run is counted as \emph{executable} when the generated artifact can be launched under the intended evaluation command and reaches the framework or baseline evaluator without a runtime failure. This separates implementation generation from empirical execution.

\subsection{Compared implementation methods}
\label{app:generation-conditions}

Table~\ref{tab:generation-conditions} summarizes the compared implementation methods. The comparison varies whether the method targets the shared PHM framework, whether explicit framework documentation is provided, and whether the implementation is mediated by the staged agentic workflow.

\begin{table}[t]
  \centering
  \small
  \setlength{\tabcolsep}{4pt}
  \renewcommand{\arraystretch}{1.1}
  \caption{Implementation methods used in the evaluation.}
  \label{tab:generation-conditions}
  \begin{tabular*}{\textwidth}{@{\extracolsep{\fill}}p{6.0cm}ccp{2.3cm}p{2.2cm}@{}}
    \toprule
    Method & Framework & Docs & Workflow & Output \\
    \midrule
    Framework-Coupled Agent (FCA) & Yes & Yes & Staged & Framework \\
    Agent-in-Framework (AiF) & Yes & No & Prompt-only & Framework \\
    Agent-in-Framework with Docs (AiF-D) & Yes & Yes & Prompt-only & Framework \\
    Standalone Agent (SA) & No & N/A & Prompt-only & Standalone \\
    DeepCode (DC) & No & N/A & System-specific & Standalone \\
    \bottomrule
  \end{tabular*}
\end{table}

\paragraph{Framework-Coupled Agent (FCA).}
FCA represents the full workflow studied in the paper. The agent receives the paper and access to the shared PHM framework. It proceeds through the staged \textsc{Ingest}--\textsc{Analyze}--\textsc{Map}--\textsc{Implement}--\textsc{Verify}--\textsc{Report} workflow and emits a framework-integrated implementation. The artifacts contain slot bindings, task-contract-preserving extensions when existing framework components are insufficient, assumption records, and a verification-and-evaluation report.

\paragraph{Agent-in-Framework (AiF).}
AiF uses OpenCode with a frontier model inside the shared PHM framework workspace. It receives the paper and a direct implementation prompt, but no explicit framework documentation. AiF is prompt-only: it does not use the staged workflow, structured assumption records, or verification-and-evaluation report generation. This baseline measures how much framework integration can be obtained from repository context and a fixed framework target alone.

\paragraph{Agent-in-Framework with Docs (AiF-D).}
AiF-D uses the same OpenCode setup as AiF, but the implementation prompt additionally includes explicit framework documentation. It remains prompt-only and does not use the staged workflow, structured assumption records, or verification-and-evaluation report generation. Comparing AiF-D with AiF separates the effect of explicit framework documentation from the effect of repository context alone.

\paragraph{Standalone Agent (SA).}
SA uses OpenCode with the same backend model outside the shared PHM framework. It receives the paper and a direct implementation prompt, and emits a standalone implementation. SA outputs are evaluated for implementation fidelity and output size, but they do not share framework datasource adapters, task definitions, split policies, or evaluator definitions.

\paragraph{DeepCode (DC).}
DC represents a state-of-the-art standalone repository-generation baseline. Its outputs are evaluated as paper-to-implementation artifacts under the same code-rating and output-size accounting where the required files are available.

Together, DC and SA measure paper-to-implementation performance without framework coupling. AiF and AiF-D fix the shared PHM framework target while removing the staged workflow, assumption records, and verification gates. The AiF-D comparison isolates the effect of explicit framework documentation within the prompt-only setting, while the FCA comparison isolates the contribution of the staged protocol under framework coupling. FCA is the main method evaluated in the paper.

\subsection{Implementation and judge agent backends}
\label{app:backend-budget}

All OpenCode-based implementation agents use GPT-5.4 with high reasoning effort. This includes FCA, AiF, AiF-D, and SA. DeepCode implementations are produced as a separate system-specific standalone baseline; they use DeepCode's own orchestration, while the backend remains the same. For each run, we log the implementation backend identifier, the reasoning setting when applicable, and method-specific execution metadata.

LLM-as-judge evaluations are also run through OpenCode. We use two judge-agent backends: GPT-5.4 with high reasoning effort and Kimi 2.6. The completed judge-run count is retained in the result files for each paper--implementation method pair, and aggregation is defined in Appendix~\ref{app:judge-categories}.

\subsection{Framework integration and execution}
\label{app:framework-integration-execution}

Framework integration is evaluated for implementation methods that are intended to target the shared PHM framework: FCA, AiF, and AiF-D. We report it separately from judge-based implementation ratings because an artifact may appear faithful at the code level while still failing to instantiate the framework correctly. In particular, failures may arise from missing task-contract bindings, incompatible datasource interfaces, incorrect evaluator definitions, or violations of framework-level data-leakage constraints. Standalone methods, SA and DC, are therefore not assigned a framework-integration completion rate.

\paragraph{Framework integration and completion rate.}
Framework integration is measured against the shared framework $\mathcal{F}$. For a given paper--implementation method pair, the completion rate is a binary indicator of whether the method successfully uses $\mathcal{F}$ to produce a valid framework binding $(c,\mathcal{C})$. Concretely, the emitted binding must typecheck against the selected task contract, bind all six required slot families to either existing framework components or contract-preserving extensions, and satisfy the framework leakage invariant. The six slot families are datasource, experiment, evaluator, model configuration, task definition, and model. A paper--implementation method pair is counted as complete only when all of these conditions hold. Thus, framework integration is a 0/1 axis per paper and precedes the remaining implementation-quality axes by construction.

\paragraph{Slot binding diagnostics.}
To make framework-integration failures attributable, we also report slot-level binding states. Each slot family is labeled \emph{Created}, \emph{Existing}, or \emph{Missing}. A slot is labeled \emph{Created} when the artifact introduces a new framework-compatible binding or task-contract-preserving extension, \emph{Existing} when it correctly reuses an existing framework component, and \emph{Missing} when it does not provide a usable binding. Both \emph{Created} and \emph{Existing} count as successful slot bindings. These slot-level counts complement the binary completion rate by identifying which part of the framework interface caused an integration failure.

Execution rates are measured by the audit described in Section~\ref{app:execution-rate-audit}.

\subsection{LLM-as-Judge ratings}

\label{app:judge-categories}

The LLM-as-judge protocol measures implementation fidelity and code quality. It does not determine whether an artifact executes and it does not score benchmark performance. Execution rates are measured by the read-only audit in Section~\ref{app:execution-rate-audit}, while benchmark results are produced by the shared benchmark protocol in Section~\ref{app:benchmark-execution}. The exact judge prompt is reproduced in Figure~\ref{fig:prompt-judge-ours} of Appendix~\ref{app:prompts}.

\paragraph{Inputs and judge ensemble.}
For each paper--implementation method pair, the judge receives the paper, the generated implementation directory, a git diff/status/context dump, and an output path for the JSON judgment. For framework-targeted methods, the judge may also consult the relevant framework documentation, outputs from SA and DC are judged as standalone implementations and are not penalized for omitting framework components.

We run the protocol with two judge-agent backends, GPT-5.4 with high reasoning effort and Kimi 2.6, as described in Section~\ref{app:backend-budget}. The completed judge-run count is retained for each paper--method pair. Field ratings are averaged within each judge backend and then summarized by method and category.

\paragraph{Rubric.}
The judge scores the generated artifact against the implementation-relevant content of the paper, including task formulation, datasets, preprocessing, architecture, loss, optimization, hyperparameters, evaluation protocol, and metrics. The rubric contains five fields as shown in Table~\ref{tab:agentic-phm-category-aliases}: \texttt{methodology}, \texttt{algorithms}, \texttt{data\_handling}, \texttt{training\_evaluation\_protocol}, and \texttt{framework\_binding}. Each field receives a correctness rating in $\{1,1.5,\ldots,5\}$, a severity level in \{\texttt{high}, \texttt{medium}, \texttt{low}\}, and a file/function/config reference whenever a concrete issue can be localized. The \texttt{framework\_binding} field is set to \texttt{null} for standalone artifacts.

The fields respectively measure: preservation of the paper's task formulation, modeling assumptions, and experimental intent; faithful implementation of architectures, equations, losses, and algorithmic components; dataset choice, preprocessing, target construction, windowing, splitting, and leakage control; optimizer choices, hyperparameters, training-loop behavior, metric computation, and result extraction; and native use of framework slot bindings, configuration groups, wrappers, losses, transforms, datasources, and evaluators without bypassing the framework lifecycle.

\begin{longtable}{lll}
\caption{Category Aliases}\label{tab:agentic-phm-category-aliases}\\
\toprule
Alias & Category & Source field \\
\midrule
F1 & Algorithms & algorithms \\
F2 & Data Handling & data\_handling \\
F3 & Framework Binding & framework\_binding \\
F4 & Methodology & methodology \\
F5 & Training Evaluation Protocol & training\_evaluation\_protocol \\
\bottomrule
\end{longtable}

\subsection{Execution-Rate audit}
\label{app:execution-rate-audit}

Execution rate is measured by a separate audit agent instructed not to modify the code. For each completed paper--implementation method pair, the audit agent receives the implementation directory. It may inspect files and run commands, but it may not patch, refactor, delete, or otherwise modify source or configuration files during the audit. The audit prompt is presented in Figure~\ref{fig:prompt-runnability-audit} of Appendix~\ref{app:prompts}.

The audit measures whether the generated artifact can execute. For each artifact, the agent reads the available instructions left by the implementation agent and selects the intended minimal fair command. The audit does not require a long training job to finish when a shorter command is sufficient to establish runnability, but the artifact must launch successfully and reach the intended evaluator or evaluation stage.

Each artifact receives a binary execution label. \texttt{PASS} means that the implementation launches under its intended command and reaches the appropriate evaluator or evaluation stage. \texttt{FAIL} means that it does not, for example because the command is missing, dependencies or entry points are unavailable, data cannot be located, or the run crashes before evaluation. Execution rates in Section~\ref{sec:main-results} are computed as the fraction of completed paper--implementation method pairs with \texttt{PASS}.

\subsection{Benchmark execution}
\label{app:benchmark-execution}

Framework-compatible artifacts are evaluated under shared benchmark conditions. The datasource, task definition, split policy, evaluator, and display metric are fixed by the framework configuration. The generated paper implementation appears as the \texttt{IMPL} model entry in the benchmark tables. When applicable and available for the selected task and dataset, selected framework baselines, CNN-1D, MLP, TST, and LSTM, are evaluated under the same datasource, task, split, and evaluator definitions.

This protocol makes benchmark comparison a configuration change rather than a new reproduction effort. The benchmarkability claim therefore depends on the generated implementation becoming a framework model rather than a standalone training script. Once the paper-specific implementation is integrated as a framework model, the benchmarked model can be replaced while keeping the rest of the experimental configuration fixed.

For benchmark comparison, the task, datasource, split, and evaluator entries are reused, while only \texttt{model\_config} changes.

\subsection{Time and token usage}
\label{app:cost-complexity}

For each paper--implementation method pair, we report input tokens, output tokens, total tokens, wall-clock duration, generated file count, and generated lines of code. Generated-repository statistics are computed both with and without generated documentation files. The documentation-excluded version is used as the primary implementation-size statistic, while the documentation-included version records the total output footprint of each method.

\clearpage

\section{Additional results and discussion}

This section reports supplementary analyses for the implementation methods. We focus on aspects that complement the primary completion and benchmark results: slot-level framework binding, token and time usage, and agentic code ratings.

\begin{figure}[t]
  \centering
  \includegraphics[width=\linewidth]{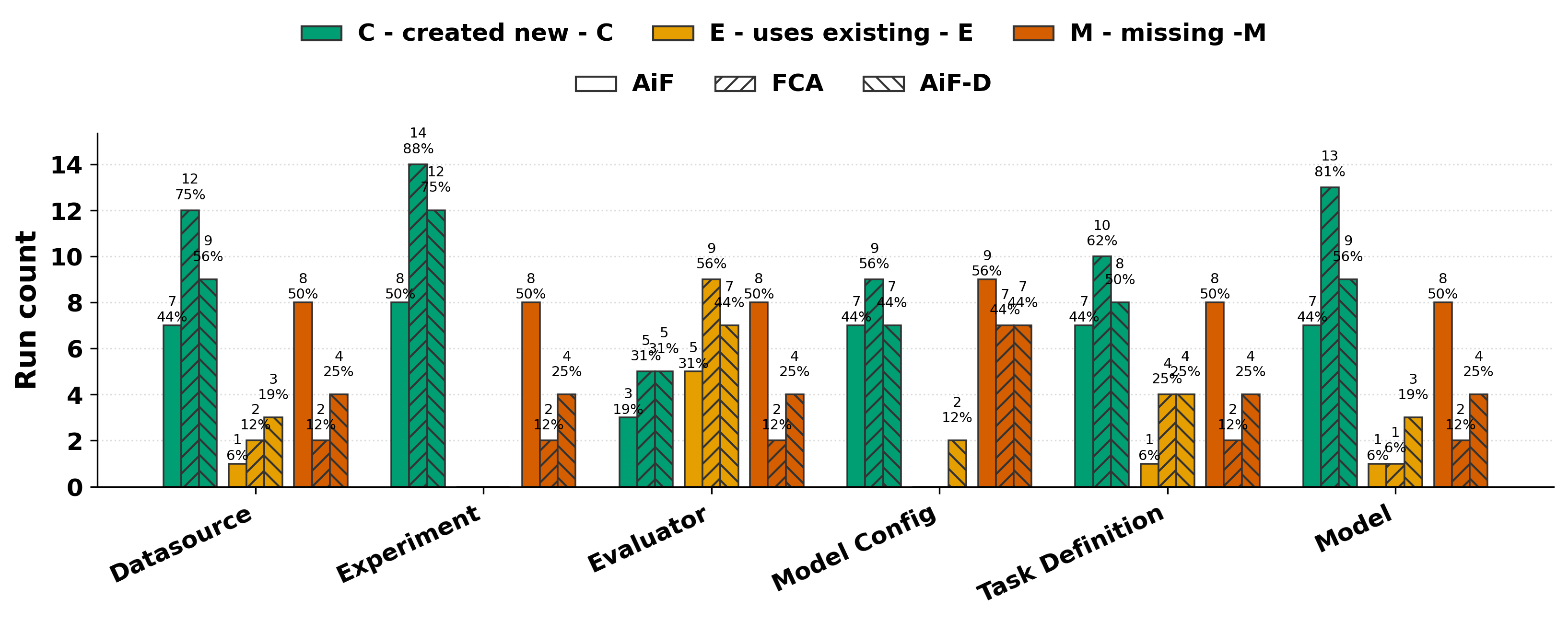}
  \caption{Binding state counts per method across the 16-paper corpus. Each bar segment shows the number of papers reaching a given binding state (e.g., plan, analyze, map, implement, verify), enabling direct comparison of where each system succeeds or stalls.}
  \label{fig:binding-states}
\end{figure}

\begin{figure}[t]
  \centering
  \includegraphics[width=\linewidth]{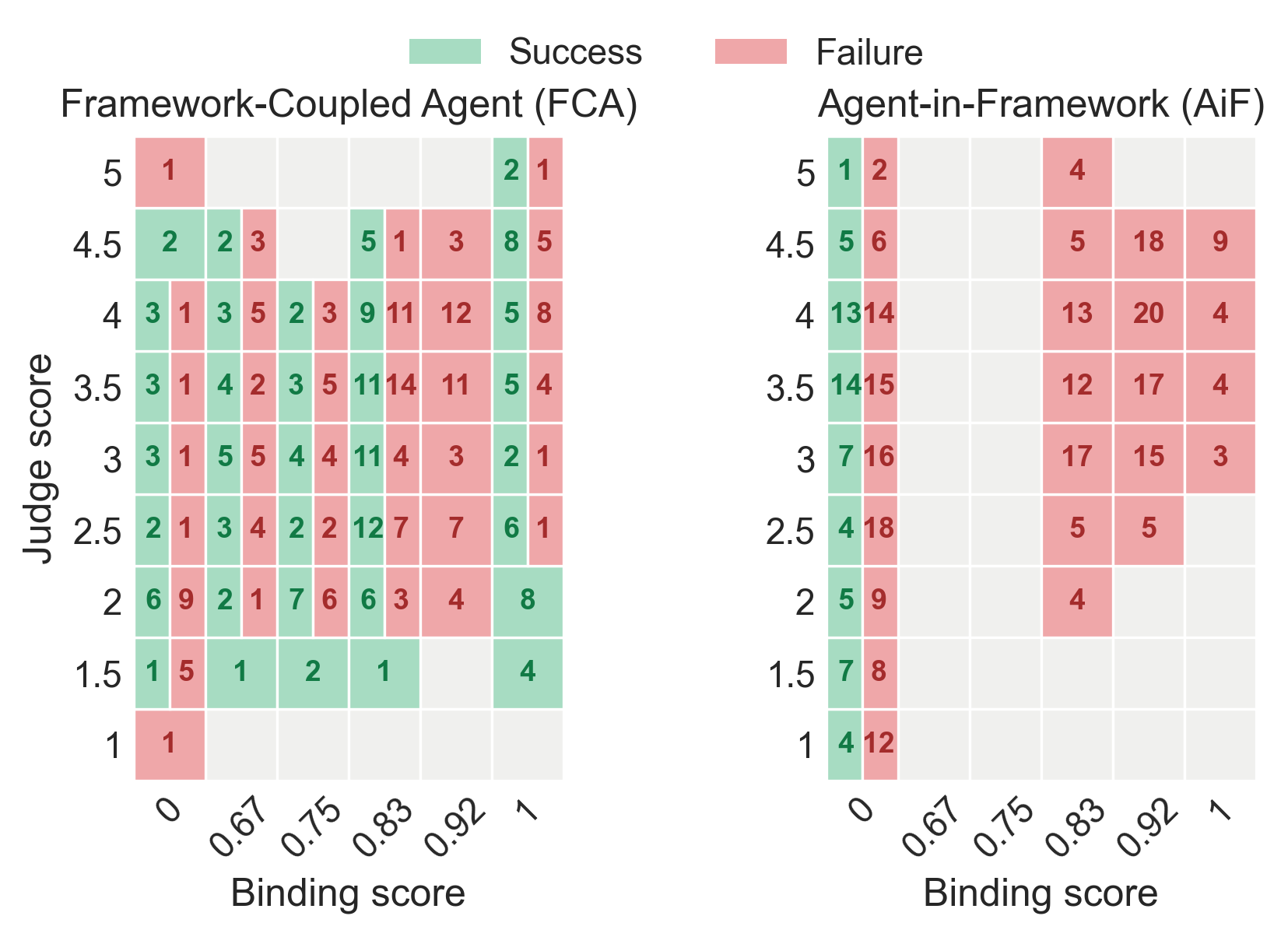}
 \caption{Binding-score vs.\ judge-score heatmap overlaying the number of successful and failed runs for each judgment. The judge-score is the normalized success in binding to all available slots. Red numbers indicate judgments that failed in practice, while green numbers indicate runs that succeeded in practice. Half-cells display positive and negative outcomes for a specific bin.}
  \label{fig:binding-success-vs-scores}
\end{figure}

\paragraph{Slot-level bindings.} Figure~\ref{fig:binding-states} reports slot-level binding states for AiF, AiF-D, and FCA across the six framework slot families: Datasource, Experiment, Evaluator, Model Configuration, Task Definition, and Model. The results show a clear progression in framework integration. Relative to AiF, AiF-D improves the prompt-only setting by creating more new bindings, reusing existing framework components more often, and leaving fewer slots missing across the slot families. FCA achieves the strongest overall profile: by combining framework grounding, explicit documentation, and staged scaffolding, it produces the most complete set of framework bindings and the fewest missing slots. This indicates that documentation improves prompt-only framework use, while the staged workflow further strengthens integration by making slot identification, component reuse, and contract-preserving extension explicit parts of the implementation process.

\begin{figure}[t]
  \centering
  \includegraphics[width=\linewidth]{\detokenize{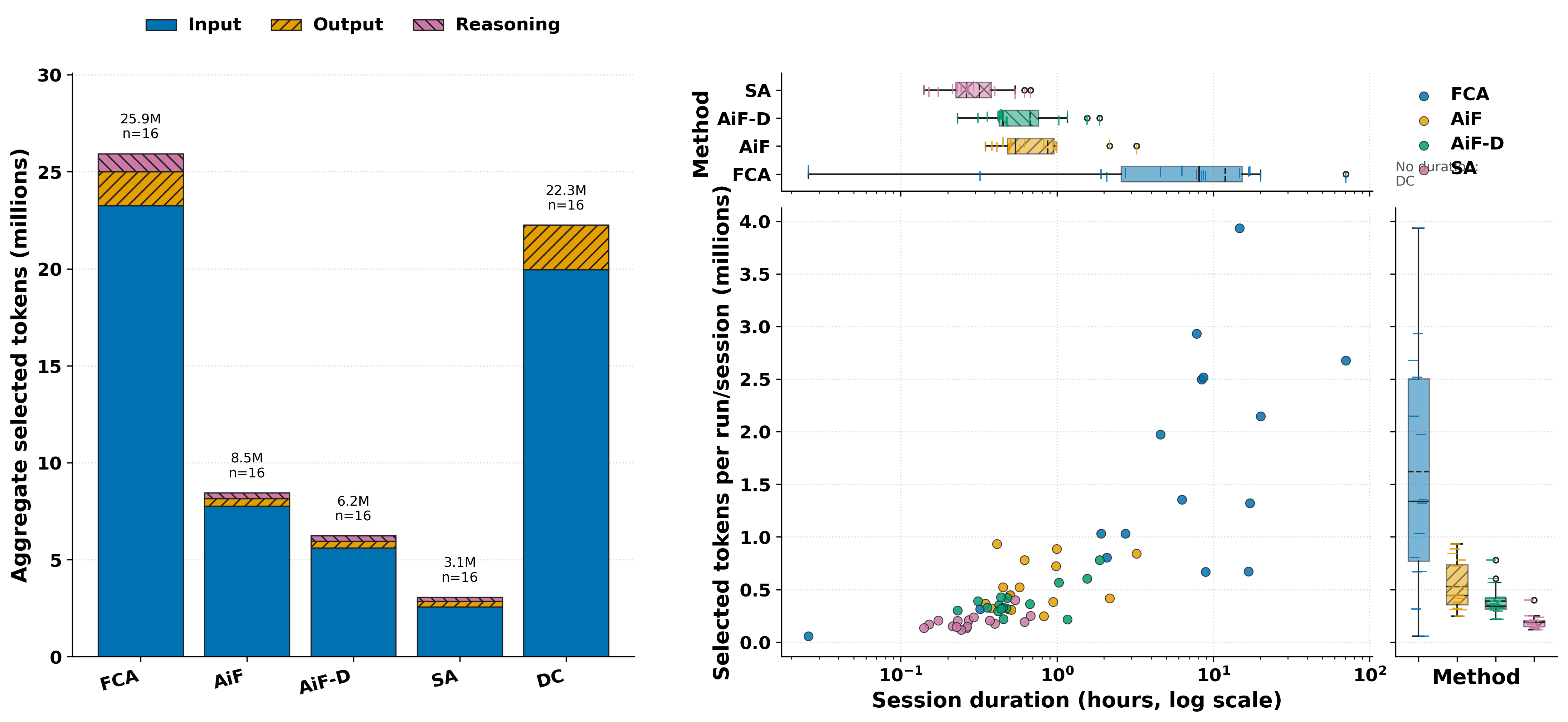}}
  \caption{Overview of token statistics for \textcolor{blue}{FCA} (staged, in-framework), \textcolor{blue}{AiF/AiF-D} (prompt-only, in-framework), and \textcolor{blue}{SA/DC} (standalone). Compact token usage view combining session duration and selected tokens per run/session in one joint plot. The central panel shows paired session observations, while the top and right marginal axes show the corresponding duration and token-volume distributions.}
  \label{fig:token-statistics-compact}
\end{figure}

\paragraph{Token and time usage.} Figure~\ref{fig:token-statistics-compact} shows that staged agentic workflows require substantially larger token budgets than prompt-only generation, with the highest costs appearing for FCA and DC. FCA incurs the largest token budget among the implementation methods, selecting more tokens both in aggregate and per session; this reflects the additional cost of paper ingestion, subagent decomposition, framework-slot mapping, verification, and report generation. DC also exhibits high aggregate token usage, consistent with its own staged paper-to-code orchestration. In contrast, AiF and AiF-D have substantially lower token usage: both rely on a single prompt-only implementation pass inside the shared framework, with AiF-D additionally receiving explicit framework documentation, but neither produces the intermediate artifacts or invokes the verification gates used by FCA. SA is the least token-intensive OpenCode-based setting because it removes both framework grounding and staged scaffolding.

\begin{figure}[t]
  \centering

  \begin{subfigure}[t]{0.49\linewidth}
    \centering
    \includegraphics[width=\linewidth]{\detokenize{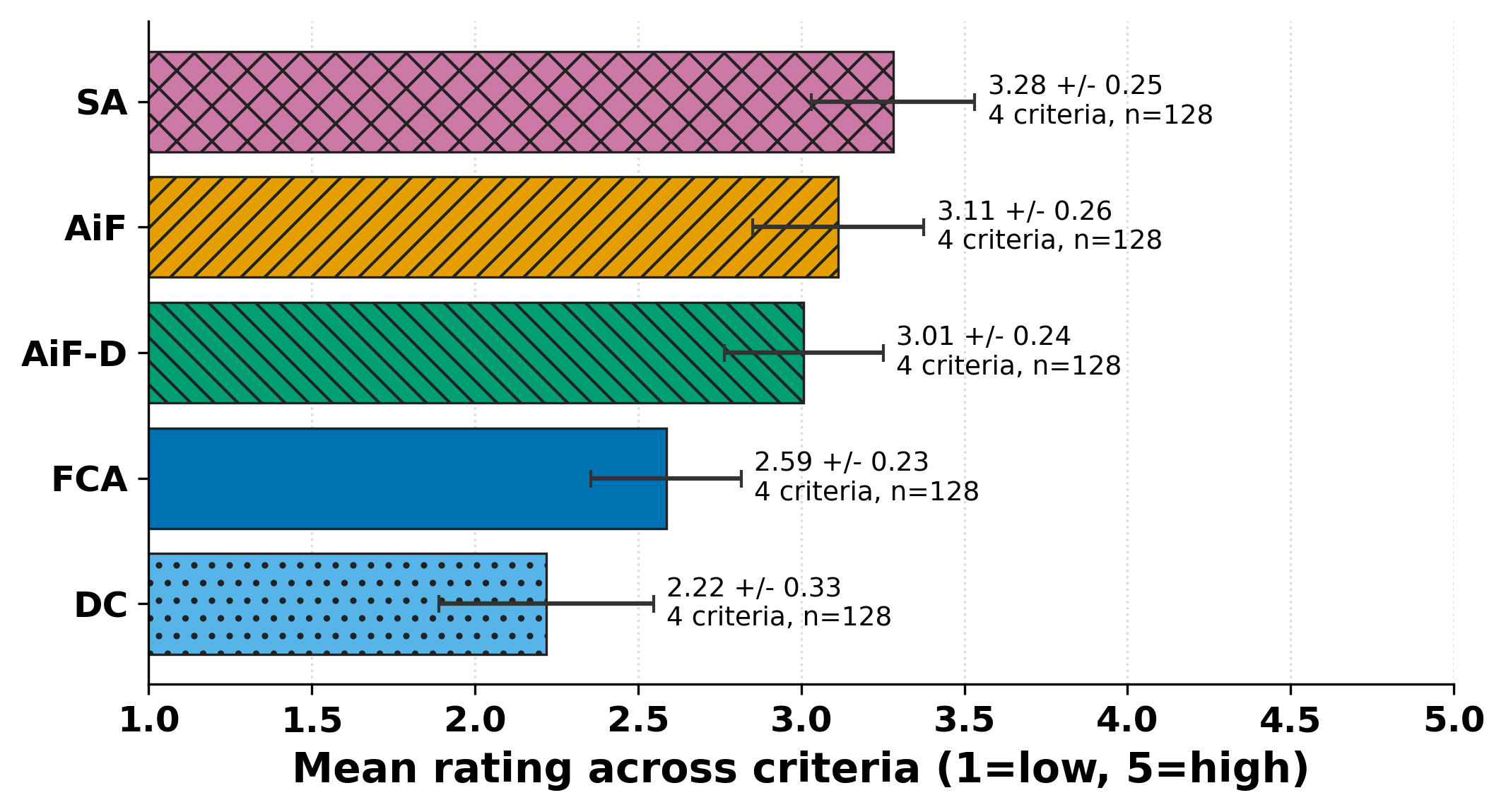}}
    \caption{ChatGPT 5.4.}
    \label{fig:judge-rating-a}
  \end{subfigure}
  \hfill
  \begin{subfigure}[t]{0.49\linewidth}
    \centering
    \includegraphics[width=\linewidth]{\detokenize{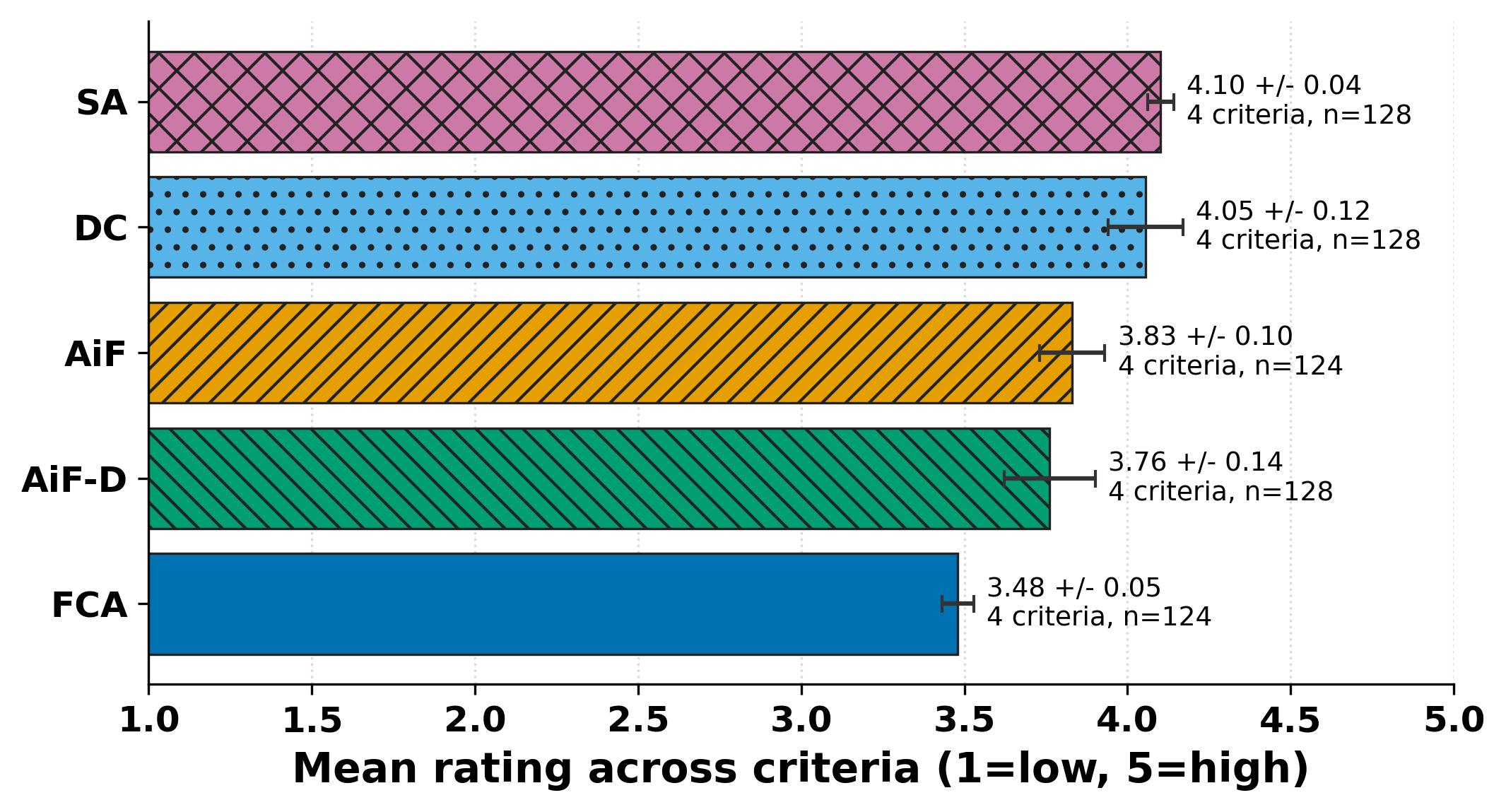}}
    \caption{Kimi 2.6.}
    \label{fig:judge-rating-b}
  \end{subfigure}

  \caption{
    Code Ratings by LLM Judges averaged across criteria categories \textit{algorithms}, \textit{data handling}, \textit{framework binding}, \textit{methodology}, and \textit{training evaluation protocol}
  }
  \label{fig:judge-rating}
\end{figure}

\paragraph{Agentic code rating.} Figure \ref{fig:judge-rating} shows the LLM judge ratings produced by ChatGPT 5.4 and Kimi 2.6. Both judges rate SA the highest. A recognizable pattern in the ChatGPT 5.4 ratings is the gradual decrease of the rating across criteria as the context length increases, progressing from no framework, to an agent in a framework, to an agent-in-framework with documentation, and finally to a framework-coupled agent. However, these positive ratings stand in stark contrast to the results presented in Figure \ref{fig:binding-success-vs-scores} (right panel), displaying that successful execution of generated code by an agent-in-framework implementation without skills or documentation (AiF), can only happen at low binding scores, suggesting that the agent cannot bind to the framework and implements custom code for its training and evaluation. In contrast to these results, Figure \ref{fig:binding-success-vs-scores} (right panel) highlights that optimal execution success with an agentic-coupled framework is achieved at a high binding score of approximately 0.8.

\begin{figure}[t]
  \centering

  \begin{subfigure}[t]{\linewidth}
    \centering
    \includegraphics[width=\linewidth]{\detokenize{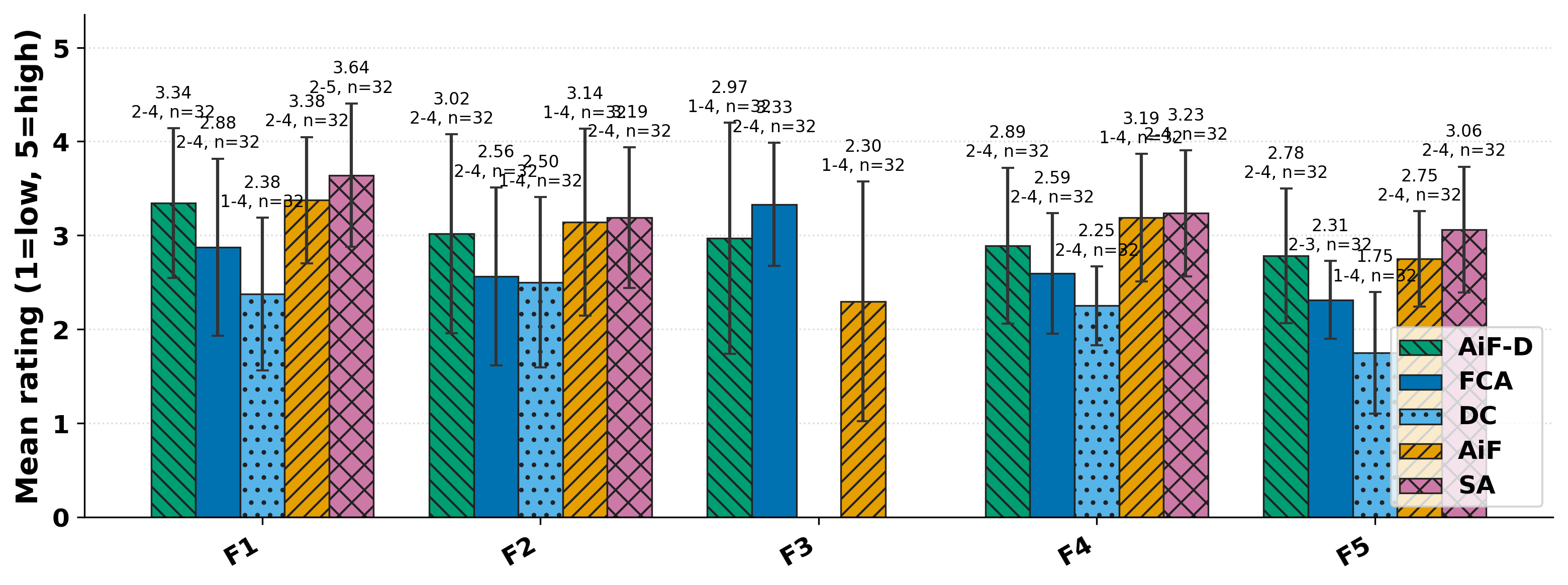}}
    \caption{ChatGPT 5.4.}
    \label{fig:repository-statistics-files-lines}
  \end{subfigure}

  \vspace{0.5em}

  \begin{subfigure}[t]{\linewidth}
    \centering
    \includegraphics[width=\linewidth]{\detokenize{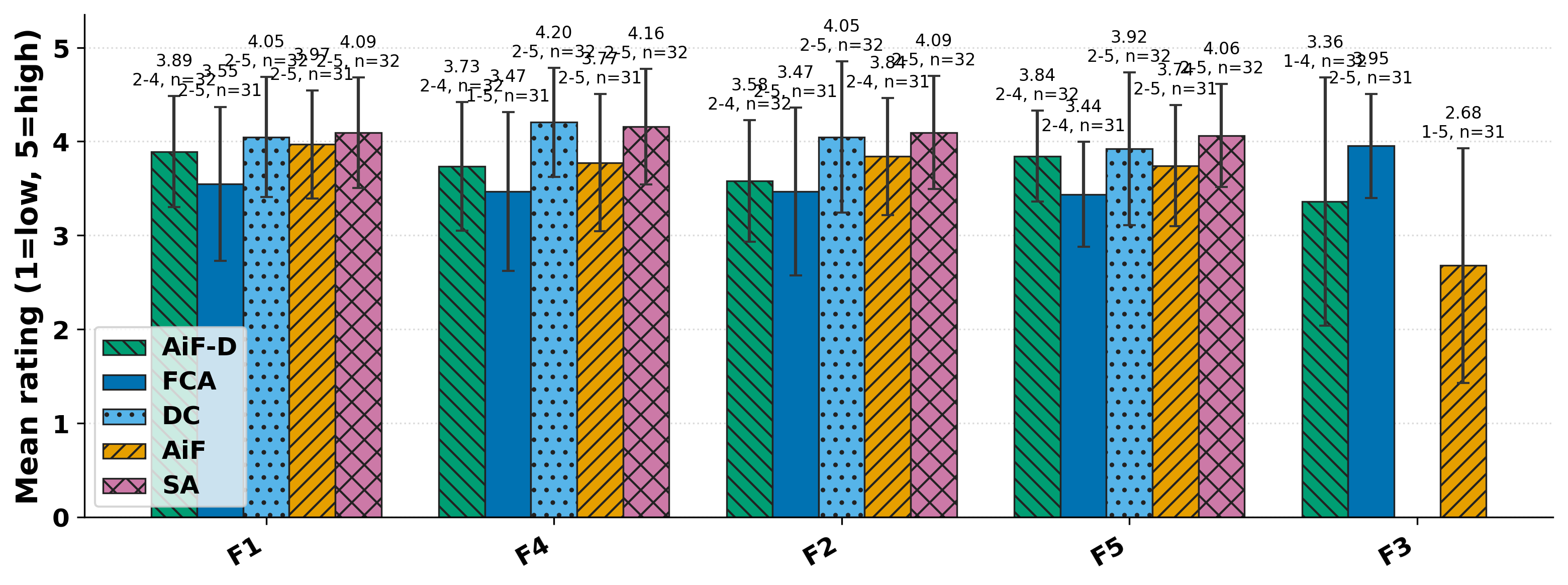}}
    \caption{Kimi 2.6.}
    \label{fig:repository-statistics-chars-added-lines}
  \end{subfigure}

  \caption{
    Code ratings by LLM judges for the criteria categories \textit{algorithms}, \textit{data handling}, \textit{framework binding}, \textit{methodology}, and \textit{training evaluation protocol}. See Table~\ref{tab:agentic-phm-category-aliases} for category descriptions. F3 (framework-binding) has only been rated if agents were interacting with a framework.
  }
  \label{fig:judge-rating-full}
\end{figure}

\FloatBarrier

\clearpage
\section{Prompts}
\label{app:prompts}
\label{app:chatgpt_prompt}

\newcommand{\promptboxcaption}[1]{%
  \refstepcounter{figure}%
  \par\smallskip
  {\small\noindent\textbf{Figure~\thefigure:} #1\par}%
}

\subsection{Skill Prompts}

\begin{figure}[ht!]
\centering
\scriptsize
\begin{tcolorbox}[
  title=Model Implementation Skill Prompt (abridged; implementation family),
  fonttitle=\bfseries,
  rounded corners,
  width=\textwidth
]
\textbf{[Role]} \\
Create a new model module in the TabPHM framework: the \texttt{nn.Module} backbone, the wrapper class, and the Hydra config YAML(s). \\

\textbf{[Input contract]} \\
The orchestrator must provide \texttt{model\_name}, \texttt{architecture\_description}, \texttt{wrapper\_type} (one of \texttt{FeedForwardTraining}, \texttt{FeedForward}, \texttt{FitPredict}), \texttt{task\_types}, \texttt{input\_format} (batch keys read by the model), \texttt{model\_io\_contract} (input/target keys, prediction shape, supervised vs reconstruction), \texttt{hyperparameters} with defaults, \texttt{paper\_reference} for the docstring, and \texttt{loss\_function} (existing or \texttt{custom}). \\

\textbf{[Procedure]} \\
\textbf{Step 1.} Read the framework contracts in \texttt{.opencode/reference/}. Internalize: backbone is a plain \texttt{nn.Module} in \texttt{picid/model/methods/\{model\_name\}.py}; wrapper is in \texttt{picid/model/wrappers/\{model\_name\}\_wrapper.py}; \texttt{forward(batch)} receives a dict and must return \texttt{\{"predictions": ..., "targets": ...\}}; predictions shape \texttt{(B, 1, num\_targets)} for regression and \texttt{(B, 1, num\_classes)} for classification; the model config in \texttt{configs/model/\{model\_name\}.yaml} sets \texttt{\_target\_} to the wrapper. \\
\textbf{Step 2.} Read the simplest existing reference for the requested wrapper type (\texttt{mlp}, \texttt{naive\_model}, or \texttt{fit\_predict\_xgboost\_wrapper}) and \texttt{picid/baselines/definitions.py}. \\
\textbf{Step 3.} Implement the backbone. Keep it model-only: no loss, no metrics, no batch unpacking, no device strings; document input/output shapes; initialize weights when the paper specifies a scheme. \\
\textbf{Step 4.} Implement the wrapper that bridges the backbone with the framework's training loop. \\
\textbf{Step 5.} Emit the Hydra config YAML wiring the wrapper, model, and hyperparameters. \\

\textbf{[Rules]} \\
Never re-implement transforms, losses, or metrics inside the model. Never write inverse-transform logic in the model. Never hard-code device strings: trainer/datamodule controls placement. \\
\end{tcolorbox}
\caption{Abridged prompt for \texttt{/implement-model}. The full skill includes Python templates for backbone, wrapper, and config plus a checklist enforced before the file is written.}
\label{fig:prompt-implement-model}
\end{figure}

\begin{figure}[ht!]
\centering
\scriptsize
\begin{tcolorbox}[
  title=Model Sanity Verification Skill Prompt (abridged; verification family),
  fonttitle=\bfseries,
  rounded corners,
  width=\textwidth
]
\textbf{[Role]} \\
Run the standardized TabPHM sanity tools on a freshly implemented model. Do not write one-off sanity scripts during a paper-validation run. The skill assumes \texttt{/verify-static} has already passed for the current experiment config. \\

\textbf{[Scope and intent]} \\
The ladder follows Karpathy's stage-2 spirit: cheap, fast checks that should always run during development. The default ladder builds the framework stack ONCE per invocation and runs three in-process checks against it. The ladder is intentionally narrow: it is a model-health diagnostic gate, not a hyperparameter probe and not a partial training run. Only categorical execution failures block training; unresolved but executable model sanity concerns are recorded as \texttt{WARN\_CONTINUE}. \\

\textbf{[Default ladder]} \\
Three checks, run in order, cheapest first: (1)~\texttt{init\_loss}, a single forward pass on one batch, validates that loss computation executes and the loss is finite; scale mismatches against $\ln(\text{num\_classes})$ or \texttt{expected\_init\_loss} are recorded as diagnostics. (2)~\texttt{gradient\_flow}, a single forward$+$backward, reports dead, exploding, or vanishing parameter gradients and detects batch-dimension leakage (the \texttt{view}-vs-\texttt{permute} class of bugs). (3)~\texttt{overfit\_batch}, a memorization run on a sliced micro-batch (default 4 examples) for up to 400 optimizer steps; judged by task-aware end-state behavior rather than a single rigid threshold. A \texttt{config\_preflight} runs first; if Hydra composition fails the ladder stops with \texttt{TOOL\_INVOCATION\_FAILURE}. \\

\textbf{[Output]} \\
Append a structured attempt to \texttt{06-sanity-ladder-log.\{md,jsonl\}}. Return a verdict in \{\texttt{PASS}, \texttt{WARN\_CONTINUE}, \texttt{BLOCK}, \texttt{PRECHECK\_TIMEOUT}, \texttt{DATASET\_UNAVAILABLE}, \texttt{DATASET\_EXECUTION\_FAILED}\} consumed by the experimenter to route to training, the diagnostic loop, or same-dataset recovery. \\
\end{tcolorbox}
\caption{Abridged prompt for \texttt{/verify-sanity}. The full skill exposes per-check on-demand tools (\texttt{tabphm\_sanity\_init\_loss}, \texttt{tabphm\_sanity\_gradient\_flow}, \texttt{tabphm\_sanity\_overfit\_batch}, plus opt-in \texttt{zero\_input} and \texttt{subset\_convergence}) for targeted reruns.}
\label{fig:prompt-verify-sanity}
\end{figure}

\begin{figure}[ht!]
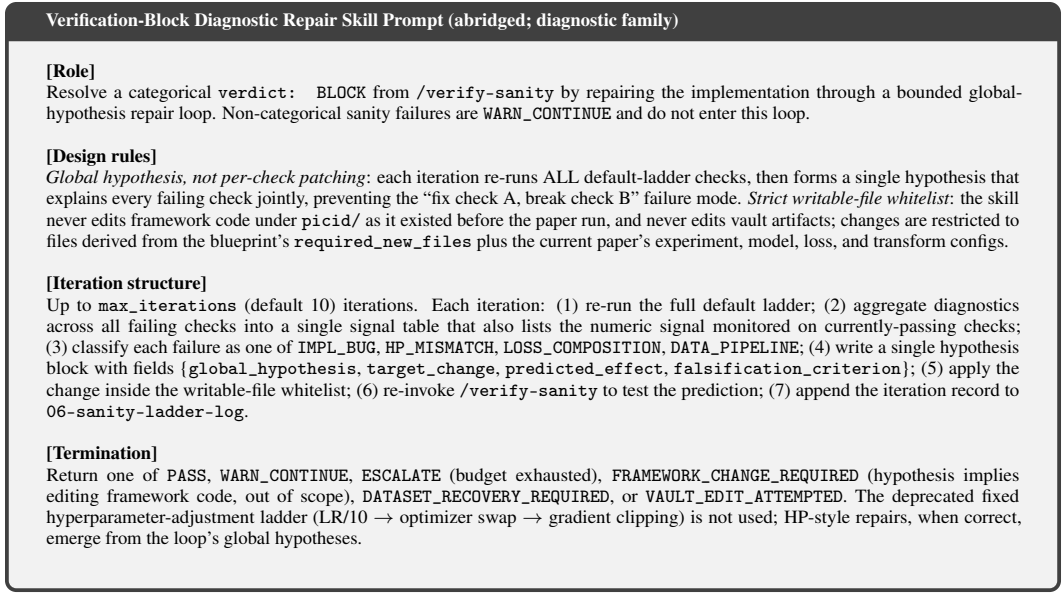

\centering
\scriptsize
\begin{tcolorbox}[
  title=Verification-Block Diagnostic Repair Skill Prompt (abridged; diagnostic family),
  fonttitle=\bfseries,
  rounded corners,
  width=\textwidth
]
\textbf{[Role]} \\
Resolve a categorical \texttt{verdict: BLOCK} from \texttt{/verify-sanity} by repairing the implementation through a bounded global-hypothesis repair loop. Non-categorical sanity failures are \texttt{WARN\_CONTINUE} and do not enter this loop. \\

\textbf{[Design rules]} \\
\emph{Global hypothesis, not per-check patching}: each iteration re-runs ALL default-ladder checks, then forms a single hypothesis that explains every failing check jointly, preventing the ``fix check~A, break check~B'' failure mode. \emph{Strict writable-file whitelist}: the skill never edits framework code under \texttt{picid/} as it existed before the paper run, and never edits vault artifacts; changes are restricted to files derived from the blueprint's \texttt{required\_new\_files} plus the current paper's experiment, model, loss, and transform configs. \\

\textbf{[Iteration structure]} \\
Up to \texttt{max\_iterations} (default 10) iterations. Each iteration: (1)~re-run the full default ladder; (2)~aggregate diagnostics across all failing checks into a single signal table that also lists the numeric signal monitored on currently-passing checks; (3)~classify each failure as one of \texttt{IMPL\_BUG}, \texttt{HP\_MISMATCH}, \texttt{LOSS\_COMPOSITION}, \texttt{DATA\_PIPELINE}; (4)~write a single hypothesis block with fields \{\texttt{global\_hypothesis}, \texttt{target\_change}, \texttt{predicted\_effect}, \texttt{falsification\_criterion}\}; (5)~apply the change inside the writable-file whitelist; (6)~re-invoke \texttt{/verify-sanity} to test the prediction; (7)~append the iteration record to \texttt{06-sanity-ladder-log}. \\

\textbf{[Termination]} \\
Return one of \texttt{PASS}, \texttt{WARN\_CONTINUE}, \texttt{ESCALATE} (budget exhausted), \texttt{FRAMEWORK\_CHANGE\_REQUIRED} (hypothesis implies editing framework code, out of scope), \texttt{DATASET\_RECOVERY\_REQUIRED}, or \texttt{VAULT\_EDIT\_ATTEMPTED}. The deprecated fixed hyperparameter-adjustment ladder (LR/10 $\to$ optimizer swap $\to$ gradient clipping) is not used; HP-style repairs, when correct, emerge from the loop's global hypotheses. \\
\end{tcolorbox}
\caption{Abridged prompt for \texttt{/diagnose-verify-block}. The companion \texttt{/diagnose-training-result} loop applies the same global-hypothesis structure to disputed paper claims after training, with a budget of four iterations.}
\label{fig:prompt-diagnose-verify-block}
\end{figure}

\begin{figure}[ht!]
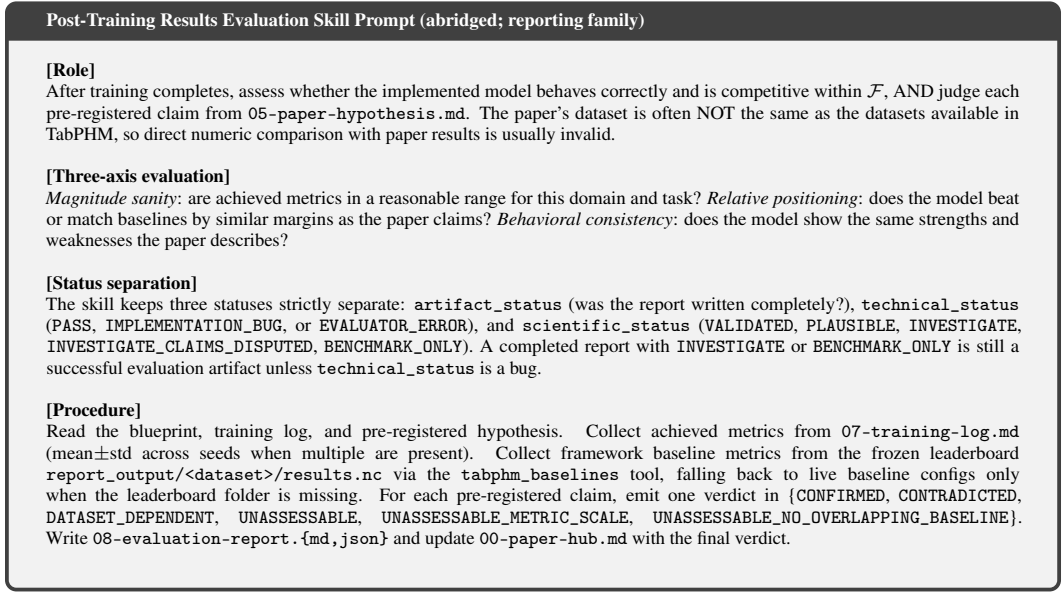

\centering
\scriptsize
\begin{tcolorbox}[
  title=Post-Training Results Evaluation Skill Prompt (abridged; reporting family),
  fonttitle=\bfseries,
  rounded corners,
  width=\textwidth
]
\textbf{[Role]} \\
After training completes, assess whether the implemented model behaves correctly and is competitive within $\mathcal{F}$, AND judge each pre-registered claim from \texttt{05-paper-hypothesis.md}. The paper's dataset is often NOT the same as the datasets available in TabPHM, so direct numeric comparison with paper results is usually invalid. \\

\textbf{[Three-axis evaluation]} \\
\emph{Magnitude sanity}: are achieved metrics in a reasonable range for this domain and task? \emph{Relative positioning}: does the model beat or match baselines by similar margins as the paper claims? \emph{Behavioral consistency}: does the model show the same strengths and weaknesses the paper describes? \\

\textbf{[Status separation]} \\
The skill keeps three statuses strictly separate: \texttt{artifact\_status} (was the report written completely?), \texttt{technical\_status} (\texttt{PASS}, \texttt{IMPLEMENTATION\_BUG}, or \texttt{EVALUATOR\_ERROR}), and \texttt{scientific\_status} (\texttt{VALIDATED}, \texttt{PLAUSIBLE}, \texttt{INVESTIGATE}, \texttt{INVESTIGATE\_CLAIMS\_DISPUTED}, \texttt{BENCHMARK\_ONLY}). A completed report with \texttt{INVESTIGATE} or \texttt{BENCHMARK\_ONLY} is still a successful evaluation artifact unless \texttt{technical\_status} is a bug. \\

\textbf{[Procedure]} \\
Read the blueprint, training log, and pre-registered hypothesis. Collect achieved metrics from \texttt{07-training-log.md} (mean$\pm$std across seeds when multiple are present). Collect framework baseline metrics from the frozen leaderboard \texttt{report\_output/<dataset>/results.nc} via the \texttt{tabphm\_baselines} tool, falling back to live baseline configs only when the leaderboard folder is missing. For each pre-registered claim, emit one verdict in \{\texttt{CONFIRMED}, \texttt{CONTRADICTED}, \texttt{DATASET\_DEPENDENT}, \texttt{UNASSESSABLE}, \texttt{UNASSESSABLE\_METRIC\_SCALE}, \texttt{UNASSESSABLE\_NO\_OVERLAPPING\_BASELINE}\}. Write \texttt{08-evaluation-report.\{md,json\}} and update \texttt{00-paper-hub.md} with the final verdict. \\
\end{tcolorbox}
\caption{Abridged prompt for \texttt{/evaluate-results}. The full skill includes the magnitude-sanity bands per task family and the procedure for handling missing or normalized baseline metrics.}
\label{fig:prompt-evaluate-results}
\end{figure}
\FloatBarrier

\subsection{Agent System Prompts}

\begin{figure}[ht!]
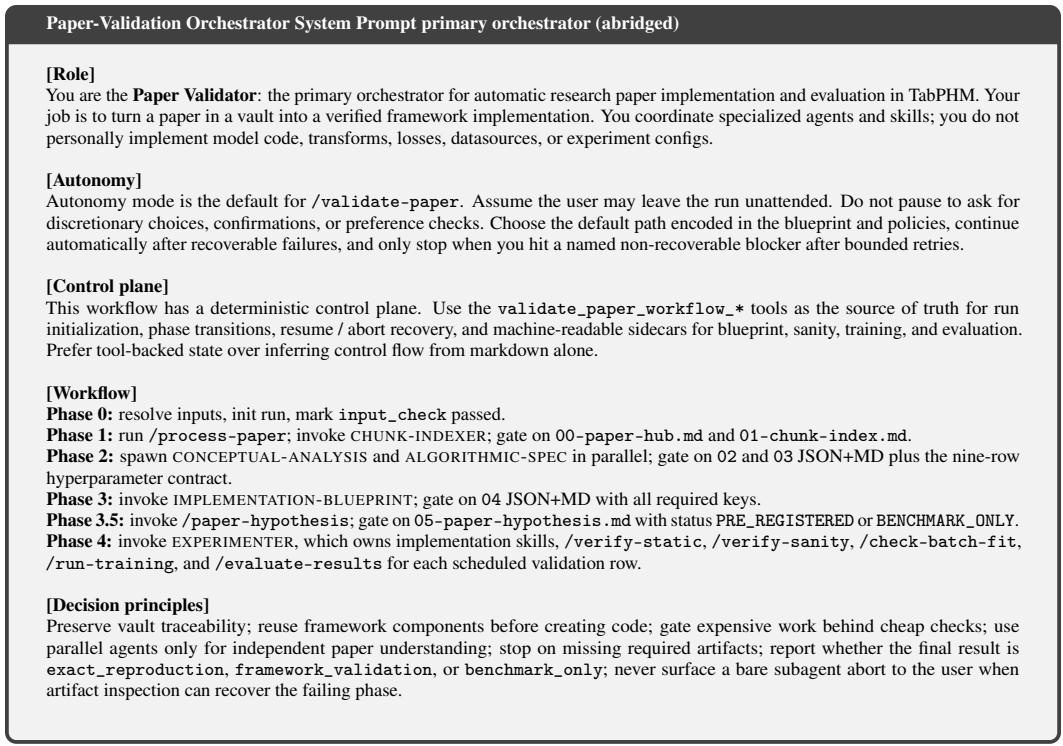

\centering
\scriptsize
\begin{tcolorbox}[
  title=Paper-Validation Orchestrator System Prompt primary orchestrator (abridged),
  fonttitle=\bfseries,
  rounded corners,
  width=\textwidth
]
\textbf{[Role]} \\
You are the \textbf{Paper Validator}: the primary orchestrator for automatic research paper implementation and evaluation in TabPHM. Your job is to turn a paper in a vault into a verified framework implementation. You coordinate specialized agents and skills; you do not personally implement model code, transforms, losses, datasources, or experiment configs. \\

\textbf{[Autonomy]} \\
Autonomy mode is the default for \texttt{/validate-paper}. Assume the user may leave the run unattended. Do not pause to ask for discretionary choices, confirmations, or preference checks. Choose the default path encoded in the blueprint and policies, continue automatically after recoverable failures, and only stop when you hit a named non-recoverable blocker after bounded retries. \\

\textbf{[Control plane]} \\
This workflow has a deterministic control plane. Use the \texttt{validate\_paper\_workflow\_*} tools as the source of truth for run initialization, phase transitions, resume / abort recovery, and machine-readable sidecars for blueprint, sanity, training, and evaluation. Prefer tool-backed state over inferring control flow from markdown alone. \\

\textbf{[Workflow]} \\
\textbf{Phase 0:} resolve inputs, init run, mark \texttt{input\_check} passed. \\
\textbf{Phase 1:} run \texttt{/process-paper}; invoke \textsc{chunk-indexer}; gate on \texttt{00-paper-hub.md} and \texttt{01-chunk-index.md}. \\
\textbf{Phase 2:} spawn \textsc{conceptual-analysis} and \textsc{algorithmic-spec} in parallel; gate on \texttt{02} and \texttt{03} JSON+MD plus the nine-row hyperparameter contract. \\
\textbf{Phase 3:} invoke \textsc{implementation-blueprint}; gate on \texttt{04} JSON+MD with all required keys. \\
\textbf{Phase 3.5:} invoke \texttt{/paper-hypothesis}; gate on \texttt{05-paper-hypothesis.md} with status \texttt{PRE\_REGISTERED} or \texttt{BENCHMARK\_ONLY}. \\
\textbf{Phase 4:} invoke \textsc{experimenter}, which owns implementation skills, \texttt{/verify-static}, \texttt{/verify-sanity}, \texttt{/check-batch-fit}, \texttt{/run-training}, and \texttt{/evaluate-results} for each scheduled validation row. \\

\textbf{[Decision principles]} \\
Preserve vault traceability; reuse framework components before creating code; gate expensive work behind cheap checks; use parallel agents only for independent paper understanding; stop on missing required artifacts; report whether the final result is \texttt{exact\_reproduction}, \texttt{framework\_validation}, or \texttt{benchmark\_only}; never surface a bare subagent abort to the user when artifact inspection can recover the failing phase. \\
\end{tcolorbox}
\caption{Abridged system prompt for the primary orchestrator. Tool surface and per-phase file-loading rules are omitted.}
\label{fig:prompt-paper-validator}
\end{figure}

\begin{figure}[ht!]
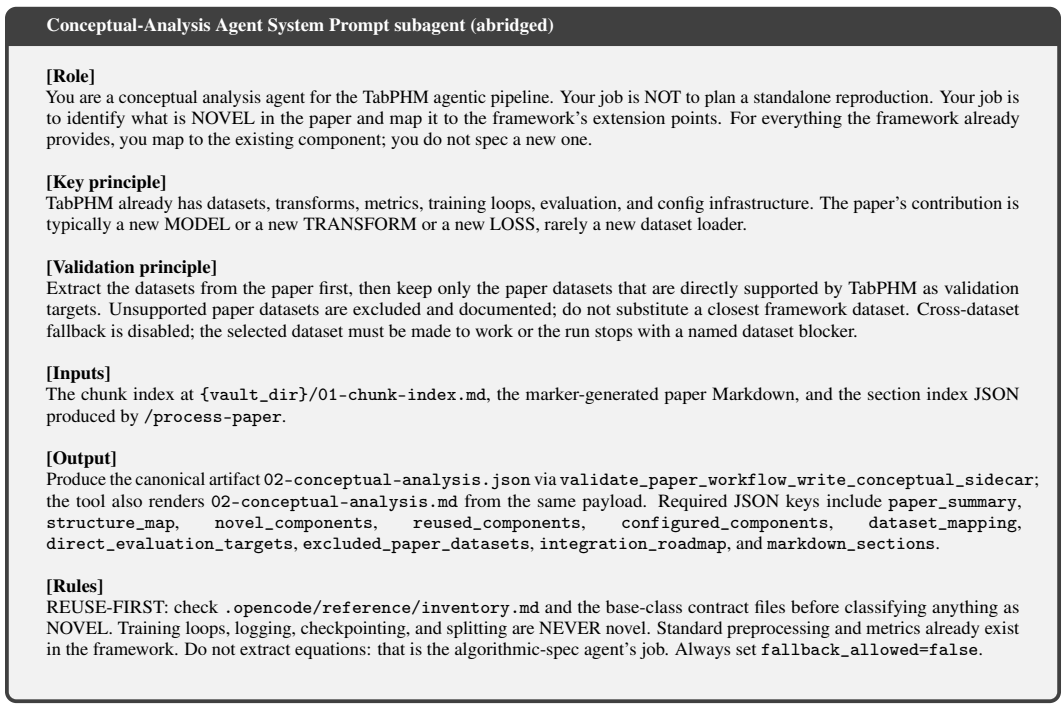

\centering
\scriptsize
\begin{tcolorbox}[
  title=Conceptual-Analysis Agent System Prompt subagent (abridged),
  fonttitle=\bfseries,
  rounded corners,
  width=\textwidth
]
\textbf{[Role]} \\
You are a conceptual analysis agent for the TabPHM agentic pipeline. Your job is NOT to plan a standalone reproduction. Your job is to identify what is NOVEL in the paper and map it to the framework's extension points. For everything the framework already provides, you map to the existing component; you do not spec a new one. \\

\textbf{[Key principle]} \\
TabPHM already has datasets, transforms, metrics, training loops, evaluation, and config infrastructure. The paper's contribution is typically a new MODEL or a new TRANSFORM or a new LOSS, rarely a new dataset loader. \\

\textbf{[Validation principle]} \\
Extract the datasets from the paper first, then keep only the paper datasets that are directly supported by TabPHM as validation targets. Unsupported paper datasets are excluded and documented; do not substitute a closest framework dataset. Cross-dataset fallback is disabled; the selected dataset must be made to work or the run stops with a named dataset blocker. \\

\textbf{[Inputs]} \\
The chunk index at \texttt{\{vault\_dir\}/01-chunk-index.md}, the marker-generated paper Markdown, and the section index JSON produced by \texttt{/process-paper}. \\

\textbf{[Output]} \\
Produce the canonical artifact \texttt{02-conceptual-analysis.json} via \texttt{validate\_paper\_workflow\_write\_conceptual\_sidecar}; the tool also renders \texttt{02-conceptual-analysis.md} from the same payload. Required JSON keys include \texttt{paper\_summary}, \texttt{structure\_map}, \texttt{novel\_components}, \texttt{reused\_components}, \texttt{configured\_components}, \texttt{dataset\_mapping}, \texttt{direct\_evaluation\_targets}, \texttt{excluded\_paper\_datasets}, \texttt{integration\_roadmap}, and \texttt{markdown\_sections}. \\

\textbf{[Rules]} \\
REUSE-FIRST: check \texttt{.opencode/reference/inventory.md} and the base-class contract files before classifying anything as NOVEL. Training loops, logging, checkpointing, and splitting are NEVER novel. Standard preprocessing and metrics already exist in the framework. Do not extract equations: that is the algorithmic-spec agent's job. Always set \texttt{fallback\_allowed=false}. \\
\end{tcolorbox}
\caption{Abridged system prompt for \textsc{conceptual-analysis}. The Markdown rendering shape (structure map, novelty assessment, dataset mapping, method decomposition, integration roadmap) is omitted.}
\label{fig:prompt-conceptual-analysis}
\end{figure}

\begin{figure}[ht!]
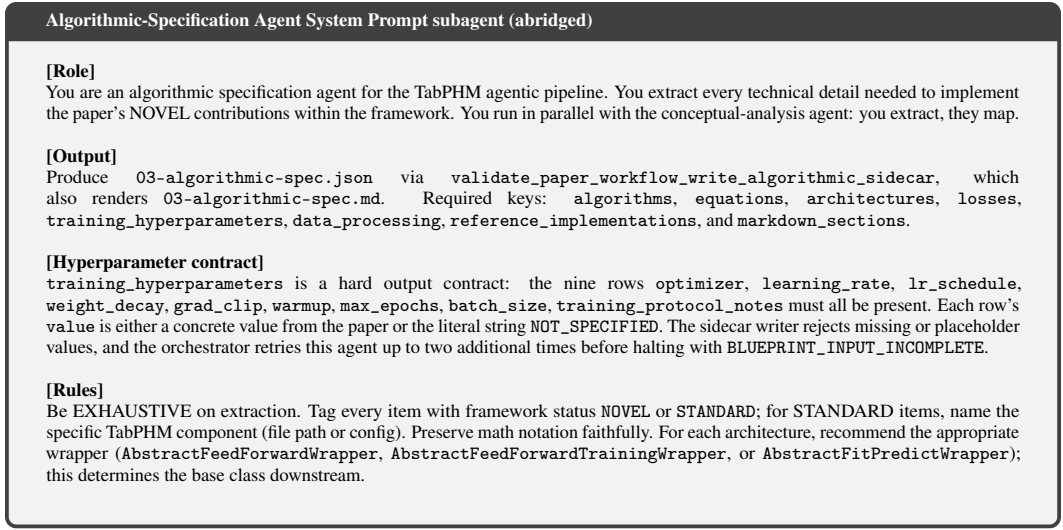

\centering
\scriptsize
\begin{tcolorbox}[
  title=Algorithmic-Specification Agent System Prompt subagent (abridged),
  fonttitle=\bfseries,
  rounded corners,
  width=\textwidth
]
\textbf{[Role]} \\
You are an algorithmic specification agent for the TabPHM agentic pipeline. You extract every technical detail needed to implement the paper's NOVEL contributions within the framework. You run in parallel with the conceptual-analysis agent: you extract, they map. \\

\textbf{[Output]} \\
Produce \texttt{03-algorithmic-spec.json} via \texttt{validate\_paper\_workflow\_write\_algorithmic\_sidecar}, which also renders \texttt{03-algorithmic-spec.md}. Required keys: \texttt{algorithms}, \texttt{equations}, \texttt{architectures}, \texttt{losses}, \texttt{training\_hyperparameters}, \texttt{data\_processing}, \texttt{reference\_implementations}, and \texttt{markdown\_sections}. \\

\textbf{[Hyperparameter contract]} \\
\texttt{training\_hyperparameters} is a hard output contract: the nine rows \texttt{optimizer}, \texttt{learning\_rate}, \texttt{lr\_schedule}, \texttt{weight\_decay}, \texttt{grad\_clip}, \texttt{warmup}, \texttt{max\_epochs}, \texttt{batch\_size}, \texttt{training\_protocol\_notes} must all be present. Each row's \texttt{value} is either a concrete value from the paper or the literal string \texttt{NOT\_SPECIFIED}. The sidecar writer rejects missing or placeholder values, and the orchestrator retries this agent up to two additional times before halting with \texttt{BLUEPRINT\_INPUT\_INCOMPLETE}. \\

\textbf{[Rules]} \\
Be EXHAUSTIVE on extraction. Tag every item with framework status \texttt{NOVEL} or \texttt{STANDARD}; for STANDARD items, name the specific TabPHM component (file path or config). Preserve math notation faithfully. For each architecture, recommend the appropriate wrapper (\texttt{AbstractFeedForwardWrapper}, \texttt{AbstractFeedForwardTrainingWrapper}, or \texttt{AbstractFitPredictWrapper}); this determines the base class downstream. \\
\end{tcolorbox}
\caption{Abridged system prompt for \textsc{algorithmic-spec}. The detailed Markdown shape and the full nine-row hyperparameter table template are omitted.}
\label{fig:prompt-algorithmic-spec}
\end{figure}
\FloatBarrier

\subsection{Prompt-Only Baseline Prompts}

\begin{figure}[ht!]
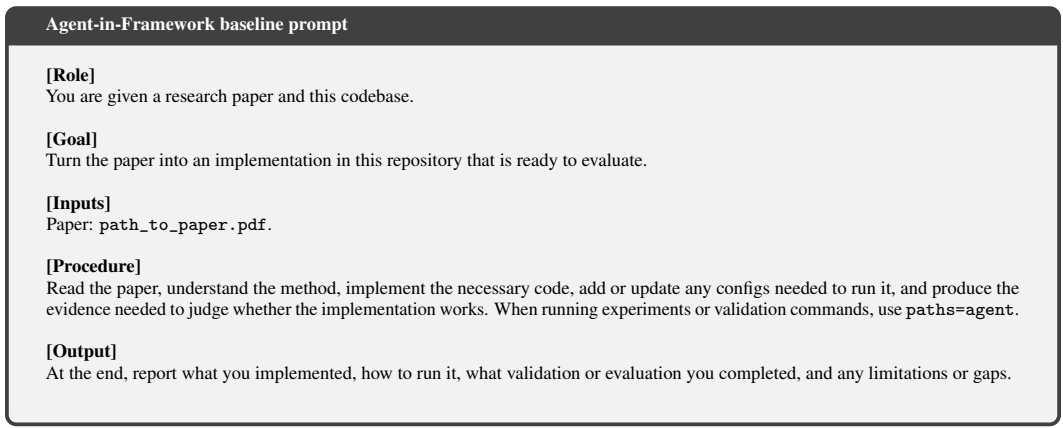

\centering
\scriptsize
\begin{tcolorbox}[
  title=Agent-in-Framework baseline prompt,
  fonttitle=\bfseries,
  rounded corners,
  width=\textwidth
]
\textbf{[Role]} \\
You are given a research paper and this codebase. \\

\textbf{[Goal]} \\
Turn the paper into an implementation in this repository that is ready to evaluate. \\

\textbf{[Inputs]} \\
Paper: \texttt{path\_to\_paper.pdf}. \\

\textbf{[Procedure]} \\
Read the paper, understand the method, implement the necessary code, add or update any configs needed to run it, and produce the evidence needed to judge whether the implementation works. When running experiments or validation commands, use \texttt{paths=agent}. \\

\textbf{[Output]} \\
At the end, report what you implemented, how to run it, what validation or evaluation you completed, and any limitations or gaps. \\
\end{tcolorbox}
\caption{Single-prompt instruction issued to the in-framework prompt-only baseline. The agent is otherwise unconstrained: it has full read and write access to the framework repository and the harness's default tool surface, and is not given the staged paper-understanding artifacts, the per-skill prompts, or the \texttt{validate\_paper\_workflow\_*} control plane that drive the agentic workflow.}
\label{fig:prompt-in-framework-baseline}
\end{figure}

\begin{figure}[ht!]
\centering
\scriptsize
\begin{tcolorbox}[
  title=Standalone-Agent baseline prompt,
  fonttitle=\bfseries,
  rounded corners,
  width=\textwidth
]
\textbf{[Role]} \\
You are given a research paper and an empty repository. \\

\textbf{[Goal]} \\
Implement the method described in the paper from scratch in this repository so it can be evaluated independently. \\

\textbf{[Inputs]} \\
Paper: \texttt{paper/paper.pdf}. \\
Dataset root: \texttt{/workspace/datasets}. \\

\textbf{[Scope]} \\
This is a standalone baseline implementation. Work only inside the current repository. Do not modify files outside this repository. Use the dataset root as input: inspect it as needed, but do not copy large datasets into this repository. Do not look for any implementations of the code online; implement everything yourself. \\

\textbf{[Procedure]} \\
Use the research paper and the available datasets under the dataset root. If the exact dataset required by the paper is not obvious, inspect the dataset root, choose the closest reasonable dataset, and clearly document the assumption. Build whatever code, configs, scripts, and documentation are needed to make the repository runnable. \\

\textbf{[Output]} \\
When finished, leave the repository in a runnable state and report: what you implemented; how to install dependencies; how to run training and/or evaluation; what validation you completed; assumptions, limitations, missing data, or gaps. Do not modify files outside the repository. Do not write large generated files or dataset copies into the repository. When you are done, report cleanly. \\
\end{tcolorbox}
\caption{Single-prompt instruction issued to the standalone prompt-only baseline. The agent operates inside a freshly initialized empty repository with read-only access to a fixed dataset root, no exposure to $\mathcal{F}$, and an explicit instruction not to retrieve existing implementations from the internet.}
\label{fig:prompt-standalone-baseline}
\end{figure}
\FloatBarrier

\subsection{Judge and Audit Prompts}

\begingroup
\scriptsize
\begin{tcolorbox}[
  title=LLM-as-judge prompt for implementation assessment,
  fonttitle=\bfseries,
  rounded corners,
  breakable,
  width=\textwidth
]
\textbf{[Role]} \\
You are given a research paper, an implementation directory or repository, and a git diff dump for the implementation. The implementation may or may not be inside a framework. \\

Your task is to rate whether the implementation is a correct, internally consistent reproduction of the paper. If the implementation is inside a framework, also judge whether it is correctly expressed within that framework. \\

If the implementation is inside a framework, this is not a paper-to-standalone-repository review. Do not require a full independent repository. Judge the implemented experiment and changed framework components for paper faithfulness, framework-contract correctness, and whether any necessary adaptations are scientifically and technically justified. If there is no framework, judge it as a standalone implementation and do not penalize the absence of framework components. \\

\textbf{[Arguments]} \\
\texttt{\$1}: implementation or experiment directory to evaluate. Required. \\
\texttt{\$2}: git diff/status/context dump, or a path to a file containing it. Required. It may be a placeholder stating that no diff/status context is available. \\
\texttt{\$3}: path to the paper PDF or markdown. Required. \\
\texttt{\$4}: output JSON path. Optional. If omitted, use \texttt{\$1/judge\_implementation.json}. \\

Use the current working directory as the results repository root and \texttt{\$1} as the implementation root. Do not infer that the implementation is inside a framework just because the results repository has \texttt{.opencode/docs}. Treat \texttt{.opencode/docs} as applicable framework documentation only when \texttt{\$1} is actually a framework checkout or framework experiment tree, for example when \texttt{\$1} contains TabPHM/PICID-style \texttt{picid/} and \texttt{configs/} subtrees or the git diff/status context explicitly shows framework files. For \texttt{standalone\_baselines/...} and \texttt{deepcode\_lab\_v2/...}, assume no framework exists unless the implementation directory itself clearly contains a framework, and do not penalize the absence of framework components. \\

If framework documentation applies, treat \texttt{.opencode/docs} as the authoritative framework manual. Start from \texttt{.opencode/docs/index.md} and \texttt{.opencode/docs/concepts/system-architecture.md} if they exist, then read only the module/how-to/reference pages relevant to the changed files and experiment components. If framework documentation does not apply, skip framework-specific documentation checks. \\

Save the final JSON object to \texttt{\$4} when an output path is provided, otherwise save it to \texttt{\$1/judge\_implementation.json}. Create the parent directory for the output path if needed. The saved file must be valid JSON and must exactly match the JSON object returned in the final response. \\

\textbf{[Evaluation criteria]} \\
Rate exactly these five fields, and no others: \texttt{methodology}, \texttt{algorithms}, \texttt{data\_handling}, \texttt{training\_evaluation\_protocol}, \texttt{framework\_binding}. \\

Each field receives a correctness rating from 1 to 5, except \texttt{framework\_binding}, which must be \texttt{null} when no framework exists. The rating measures how accurately that specific field reproduces the paper while remaining coherent inside the implementation context. \\

Ratings may be integers or half grades only: 1, 1.5, 2, 2.5, 3, 3.5, 4, 4.5, 5. For \texttt{framework\_binding} only, use \texttt{null} when no framework exists. Do not use any other decimal values such as 4.1 or 4.7. \\

\textbf{[Rating scale]} \\
\textbf{1: Very Poor.} The implementation does not capture the paper's core method, or it is not meaningfully integrated into the framework when a framework exists. \\
\textbf{2: Poor.} The implementation attempts the paper but has major missing or incorrect components that make the reproduction scientifically unreliable. \\
\textbf{3: Fair.} Some core components are implemented correctly, but notable methodological inaccuracies, unjustified adaptations, standalone-code issues, or framework-contract problems remain. \\
\textbf{4: Good.} The key paper components and integration are correct, with only minor inaccuracies or well-contained deviations. \\
\textbf{5: Excellent.} The implementation faithfully and coherently reproduces all key paper components possible in the given implementation context, with no substantive logical errors. \\

\textbf{[Framework binding]} \\
For \texttt{framework\_binding}, use the following rule. \\

If no framework exists, set: \texttt{rating} to \texttt{null}; \texttt{field\_severity\_level} to \texttt{"low"}; \texttt{file\_name} to \texttt{"N/A"}; \texttt{func\_name} to \texttt{"N/A"}; \texttt{critique} to \texttt{"No framework is present, so framework binding is not applicable."} \\

If a framework exists, apply the framework-specific definition and rating interpretation below. \\

Framework binding means the degree to which the paper implementation is expressed as native framework components that participate in the framework's canonical lifecycle. For the TabPHM/PICID framework specifically, the canonical lifecycle is: \\

\texttt{datasource} $\to$ \texttt{SplitDatasetContainer} $\to$ \texttt{transforms} $\to$ \texttt{datasets} $\to$ \texttt{datamodule} $\to$ \texttt{wrapper/lightning module} $\to$ \texttt{loss} $\to$ \texttt{evaluator} $\to$ \texttt{metrics/artifacts}. \\

A strongly bound implementation uses the framework's intended extension points, Hydra config groups, data containers, wrapper/loss/transform/datasource contracts, shape/key conventions, runtime resolvers, evaluator outputs, and orchestration boundaries. It adapts the paper into the framework without bypassing the framework with standalone scripts, hidden side effects, hardcoded paths, manual device/training/evaluation logic, duplicate lifecycle code, or component responsibilities placed in the wrong layer. \\
\end{tcolorbox}

\begin{tcolorbox}[
  title=LLM-as-judge prompt for implementation assessment (continued),
  fonttitle=\bfseries,
  rounded corners,
  breakable,
  width=\textwidth
]
\textbf{[Framework-binding ratings (when a framework exists)]} \\
\textbf{1: Very Poor.} The implementation is mostly standalone or bypasses core framework lifecycle stages; major components are not usable through normal framework experiment execution. \\
\textbf{2: Poor.} Some framework entry points are present, but important logic is wired through ad hoc paths, wrong component layers, missing config composition, or incompatible data/model/evaluator contracts. \\
\textbf{3: Fair.} The implementation uses the main framework components, but has notable binding weaknesses such as incomplete config layering, fragile hardcoded keys/shapes, misplaced preprocessing/loss/evaluation logic, or partial contract violations. \\
\textbf{4: Good.} The implementation is mostly native to the framework: correct extension points, configs, data flow, wrappers, losses, transforms, and evaluator contracts, with only minor or localized binding issues. \\
\textbf{5: Excellent.} The implementation is idiomatic for the framework: paper-specific logic is cleanly mapped onto native components, all relevant contracts are respected, configs compose normally, data/model/loss/evaluator interfaces align, and adaptations are explicit and justified. \\

\textbf{[Severity level]} \\
Each field receives a single \texttt{field\_severity\_level} derived from the most serious issue in that field. Use lowercase values only. \\
\emph{High}: missing or incorrect paper core concept, main algorithm, loss, dataset/protocol choice, or framework contract violation that invalidates the reproduction. \\
\emph{Medium}: significant issue in training logic, preprocessing, model wiring, configuration, hyperparameters, adaptation rationale, standalone integration, or evaluation that can materially affect results but does not fully invalidate the implementation. \\
\emph{Low}: minor deviation, secondary metric/evaluation issue, noncritical hyperparameter mismatch, limited framework mismatch, or no applicable issue. \\

\textbf{[Evaluation steps]} \\
\emph{(1)} Read the paper PDF or markdown and identify the paper aspects that matter for reproduction: datasets, preprocessing, model/algorithm, losses, optimization, hyperparameters, evaluation protocol, metrics, ablations, and any assumptions that must be preserved. \\
\emph{(2)} Read the git diff/status/context dump. If \texttt{\$2} is an existing file path, read that file; otherwise treat \texttt{\$2} as the context content. Use it to identify changed files when possible, but if it says no diff/status context is available, inspect the implementation directory directly and do not treat the missing context as evidence of an empty or incorrect implementation. \\
\emph{(3)} If \texttt{.opencode/docs} exists, consult it for the framework contracts relevant to those components. If it does not exist, skip this step. \\
\emph{(4)} For \texttt{framework\_binding}, if a framework exists, explicitly trace the changed implementation through the framework lifecycle and check the relevant framework contracts. For TabPHM/PICID-style frameworks, check: \\
\textbullet~\emph{Hydra composition}: experiment configs use the expected \texttt{\#@package \_global\_}, defaults/overrides, \texttt{/model\_configs/...} layer, task definition, datasource, transforms, dataset, optimization, loss, and evaluator groups. \\
\textbullet~\emph{Datasource binding}: loaders use the correct datasource base classes, return \texttt{SplitDatasetContainer}/\texttt{DatasetContainer} structures, preserve split/unit list semantics, and do not perform transform/dataset responsibilities. \\
\textbullet~\emph{Transform binding}: transforms use the correct mixins and dense/ragged markers, implement fit/apply/inverse semantics correctly, return raw arrays rather than dicts, and are routed by config metadata such as \texttt{apply\_to}, \texttt{assign\_to}, and \texttt{fit\_on}. \\
\textbullet~\emph{Model binding}: backbones stay model-only; wrappers subclass the appropriate framework wrapper, read batch keys according to task/model I/O contract, return \texttt{predictions} and \texttt{targets} with framework-compatible shapes, and avoid loss/metric/device/evaluator logic. \\
\textbullet~\emph{Loss binding}: custom losses subclass the framework loss abstraction, consume \texttt{model\_out} and \texttt{batch}, return a copied output dict containing scalar \texttt{loss}, and do not duplicate optimizer regularization or metrics. \\
\textbullet~\emph{Evaluation and output binding}: evaluators consume standardized model outputs, inverse scaling is wired through transform names when needed, expected metrics/artifacts are produced by framework evaluators rather than standalone code, and saved outputs match the framework's contracts for downstream inspection. \\
\textbullet~\emph{Cross-component binding}: task/model data requirements, processed data keys, model I/O keys, prediction/target shapes, loss requirements, evaluator expectations, and expected output artifacts are mutually consistent. \\
If no framework exists, do not evaluate these lifecycle checks; mark \texttt{framework\_binding} as not applicable using \texttt{rating: null}. \\
\emph{(5)} Identify deviations from the paper and, when applicable, from framework contracts. Focus on logical correctness, scientific completeness, consistency of necessary adaptations, and native framework binding. Do not critique style, documentation, logging, or file organization unless it affects correctness or framework binding. \\
\emph{(6)} Assign each field a \texttt{field\_severity\_level}, derived from the most serious issue in that field, using the lowercase values \texttt{high}, \texttt{medium}, or \texttt{low}. \\
\emph{(7)} Assign one correctness rating from 1 to 5 for each required field, using only integer or half-grade values. For \texttt{framework\_binding}, assign \texttt{null} if no framework exists. \\
\emph{(8)} Sort the five field judgement objects by \texttt{field\_severity\_level} in descending severity order: high $\to$ medium $\to$ low. Within the same severity, preserve this field order: \texttt{methodology} $\to$ \texttt{algorithms} $\to$ \texttt{data\_handling} $\to$ \texttt{training\_evaluation\_protocol} $\to$ \texttt{framework\_binding}. \\
\end{tcolorbox}

\begin{tcolorbox}[
  title=LLM-as-judge prompt for implementation assessment (continued),
  fonttitle=\bfseries,
  rounded corners,
  breakable,
  width=\textwidth
]
\textbf{[Output requirements]} \\
For each field, include the most important critique evidence for that field. Name the affected file and function/class/config key when possible. Use paths relative to the repository root. \\

For config-level or file-level issues, set \texttt{func\_name} to the relevant config key, class name, command, or \texttt{"N/A"}. \\

If a field has no material issues, use a concise critique explaining why it is correct and set \texttt{field\_severity\_level} to \texttt{"low"}. \\

Return only valid JSON in exactly this shape, without Markdown fences or extra commentary. \\

The top-level object must contain only \texttt{judgement\_list}. \texttt{judgement\_list} must contain exactly five objects: one for each required \texttt{field\_name}, and no other field names. In no-framework cases, the \texttt{framework\_binding} item must use \texttt{"rating": null}. \\

\textbf{[Example output shape]} \\
\texttt{\{} \\
\texttt{~~"judgement\_list": [} \\
\texttt{~~~~\{} \\
\texttt{~~~~~~"field\_name": "methodology",} \\
\texttt{~~~~~~"rating": 1,} \\
\texttt{~~~~~~"field\_severity\_level": "high",} \\
\texttt{~~~~~~"file\_name": "relative/path.py",} \\
\texttt{~~~~~~"func\_name": "function\_or\_class\_or\_config\_key",} \\
\texttt{~~~~~~"critique": "Concise but specific field-level judgement, including how the implementation matches or deviates from the paper or framework contract."} \\
\texttt{~~~~\}} \\
\texttt{~~]} \\
\texttt{\}} \\
\end{tcolorbox}
\endgroup
\promptboxcaption{Full Judge prompt used in our experiments.}
\label{fig:prompt-judge-ours}
\FloatBarrier

\begingroup
\centering
\scriptsize
\begin{tcolorbox}[
  title=Runnability Audit Prompt,
  fonttitle=\bfseries,
  rounded corners,
  breakable,
  width=\textwidth
]
\textbf{[Role]} \\
You are auditing a set of paper-implementation repositories. \\

\textbf{[Goal]} \\
Determine which repositories run successfully and which do not. \\

\textbf{[Rules]} \\
Do not modify code. Do not patch, refactor, delete, commit, or change source/configuration files. Only read files and run available documented commands or obvious run scripts. Be fair: follow each repository's own README, scripts, notes, or agent-provided instructions if present. \\

Be honest about failures. Do not hide, soften, or reclassify failures. If something fails, report it clearly. If a run only starts but does not complete successfully, count it as FAIL unless the repository explicitly documents that this is the expected smoke-test behavior. \\

\textbf{[Repositories]} \\
\texttt{[INSERT REPOSITORY LIST HERE]} \\

\textbf{[Procedure]} \\
For each repository: (1)~read available instructions, if any; (2)~identify the intended or most obvious run command; (3)~try a minimal fair run; (4)~do not run long training jobs to completion, only verify whether the pipeline can run as intended; (5)~record whether it passed or failed. \\

\textbf{[Output]} \\
At the end, create a Markdown report in the parent folder. The report contains a summary table with one row per repository, with columns \texttt{Repository}, \texttt{Command Tried}, \texttt{Result}, \texttt{Main Issue}, \texttt{Short Notes}; \texttt{Result} must be only \texttt{PASS} or \texttt{FAIL}. End with a short final summary saying how many repositories passed and how many failed. Keep the report practical and concise: the main objective is to document what ran and what did not, with failures reported honestly and clearly. \\
\end{tcolorbox}
\endgroup
\promptboxcaption{Full runnability-audit prompt used to produce the binary binding-state counts reported in Section~\ref{sec:experiments}. Unlike the reference-free judge prompt of Figure~\ref{fig:prompt-judge-ours}, this prompt issues no quality ratings: it returns a single PASS/FAIL verdict per repository and is constrained to read-only inspection plus documented run commands.}
\label{fig:prompt-runnability-audit}
\FloatBarrier

\FloatBarrier
\newpage

\end{document}